\pdfoutput=1
\documentclass{article} %
\usepackage{iclr2027_conference,times}
\iclrfinalcopy
\usepackage{graphicx}

\usepackage{amsmath,amsfonts,bm}

\def\eqref#1{equation~\ref{#1}}

\def\1{\bm{1}}

\DeclareMathAlphabet{\mathsfit}{\encodingdefault}{\sfdefault}{m}{sl}
\SetMathAlphabet{\mathsfit}{bold}{\encodingdefault}{\sfdefault}{bx}{n}

\DeclareMathOperator*{\argmax}{arg\,max}

\DeclareMathOperator{\Tr}{Tr}

\usepackage[hidelinks]{hyperref}
\usepackage{url}
\usepackage{mathtools}
\usepackage[inline]{enumitem}
\usepackage{wrapfig}

\usepackage{tikz}
\usetikzlibrary{arrows.meta,decorations.pathreplacing}
\usetikzlibrary{arrows.meta}
\usepackage{fix-cm}

\newcommand{\Rd}{\mathbb{R}^d}

\title{Does LinUCB Need Valid Confidence Sets?}

\author{Ahmed Ben~Yahmed$^{1,2,3 }$\thanks{Correspondence to: a.benyahmed@criteo.com.}  ,  Marc Abeille$^{1,3}$, David Janz$^{4}$ \& Clément Calauzènes$^{1,3}$  \\
$^1$~Criteo AI Lab, Paris, France\\
$^2$~CREST, ENSAE, Institut Polytechnique de Paris, France\\
$^3$~FairPlay joint team, Inria, France \\
$^4$~University of Oxford
}

\usepackage{global_aliases}

\begin{document}
\etocdepthtag.toc{main}

\maketitle
\lhead{Preprint}

\begin{abstract}
We show that replacing LinUCB's usual confidence width $\beta$ by any fixed $\alpha>0$ preserves sublinear regret when the action set is $M$-smooth and $m$-strongly convex. Our high-probability regret bound for this generalisation, $\alpha$-LinUCB, establishes that valid confidence sets are not necessary for the success of LinUCB. We identify the mechanism behind this result: action-set curvature drives $\alpha$-LinUCB to explore every direction, even when optimism is not guaranteed. To analyse this mechanism, we introduce a logarithmically capped spectral potential that, unlike the standard log-determinant potential, caps contributions from well-explored directions. This yields lower bounds on the smallest eigenvalue of the design matrix. At the standard confidence width $\alpha=\beta\asymp\sqrt{d\log T}$, our $T$-round regret bound in $d$ dimensions is $\cO\bigl(\frac{M}{m}d\sqrt{T\log T}\bigr)$, up to iterated-logarithmic factors, improving the horizon dependence of the classical LinUCB guarantee by a factor of $\sqrt{\log T}$.
\end{abstract}

\section{Introduction}\phantomsection\label{sec:introduction}

A learner interacting with a bandit environment chooses actions sequentially to earn rewards while gathering the information needed to improve future decisions. We consider this exploration--exploitation trade-off in the stochastic linear bandit setting. Fix an unknown parameter $\theta^\star \in \Rd$, a closed action set $\cX$ contained in the unit ball of $\Rd$, and an integer horizon $T \geq 1$. At each round $t \in \{1,\dots,T\}$, the learner selects an action $x_t \in \cX$ and receives a reward $y_t = \dotprod{\theta^\star}{x_t} + \eta_t$, where the noise $\eta_t$ is conditionally centred and sub-Gaussian. We consider the $T$-round (pseudo-)regret,
\[
    \cR_T := \sum_{t=1}^{T} \dotprod{\theta^\star}{x^\star - x_t}, \qquad x^\star \in \argmax_{x \in \cX} \dotprod{\theta^\star}{x}.
\]

LinUCB addresses this trade-off using upper confidence bounds on rewards \citep{dani2008stochastic,Li-LINUCB,yassin}. We study $\alpha$-LinUCB (\cref{alg:alpha-ucb}), a generalisation that treats the confidence width as a free parameter. Given fixed $\alpha,\lambda_0 > 0$, at each round $t$, the algorithm selects $x_t$ through
\[
    (x_t,\theta_t) \in \argmax_{(x,\theta) \in \cX \times \cC_t(\alpha)} \dotprod{\theta}{x}, \qquad \cC_t(\alpha) := \left\{ \theta \in \Rd : \norm{\theta - \hat{\theta}_t}_{V_t} \leq \alpha \right\},
\]
where $\hat{\theta}_t = V_t^{-1} \sum_{s=1}^{t-1} x_s y_s$ is the ridge estimate of $\theta^\star$ with design $V_t := \lambda_0 \I + \sum_{s=1}^{t-1} x_s x_s^\top$. Equivalently, the action satisfies
\[
    x_t \in \argmax_{x \in \cX} \left\{ \dotprod{\hat{\theta}_t}{x} + \alpha \norm{x}_{V_t^{-1}} \right\}.
\]
Thus, actions are selected through a combination of the mean and an $\alpha$-weighted exploration bonus, with $\norm{x}_{V_t^{-1}}$ being large for $x$ aligned with directions poorly represented by previous actions.

Standard LinUCB chooses a confidence width $\beta \asymp \sqrt{d\log T}$ so that $\theta^\star \in \bigcap_{t=1}^{T} \cC_t(\beta)$ with high probability. On this event, the selected pair $(x_t,\theta_t)$ satisfies the optimism inequality $\dotprod{\theta_t}{x_t} \geq \dotprod{\theta^\star}{x^\star}$, which underlies its usual regret guarantee \citep{yassin}. For arbitrary $\alpha>0$, $\alpha$-LinUCB retains an exploration bonus, but its ellipsoids need not be valid confidence sets. The optimism inequality is then no longer guaranteed, so the usual regret argument does not extend to arbitrary choices of $\alpha$. This raises the question: \emph{Does LinUCB Need Valid Confidence Sets?}

Under an additional curvature condition on the action set, we answer this question in the negative:
\begin{itemize}
    \item \emph{Regret without valid confidence sets.} When the squared support function of the action set is $M$-smooth and $m$-strongly convex, we prove a high-probability regret bound of\footnote{Throughout, we use the standard asymptotic notation $\cO( \cdot )$ and $\Omega( \cdot )$, up to iterated-logarithm factors.} 
    \[
        \cR_T = \cO\Bigl(\frac{M}{m}\frac{\alpha^2+\beta^2}{\alpha}\sqrt{dT}\Bigr),
    \]
    up to iterated-logarithmic factors (\cref{thm:alpha-UCB-Regret}). The bound is sublinear for every fixed $\alpha>0$, including widths for which the exploration ellipsoids are not valid confidence sets. The geometric condition, stated in \cref{ass:armset-int}, includes Euclidean balls, ellipsoids, and convex bodies with uniformly positive and bounded curvature.

    \item \emph{A geometric exploration mechanism.} We identify the mechanism through which $\alpha$-LinUCB works without optimism: action-set curvature drives $\alpha$-LinUCB to explore every direction, even when optimism is not guaranteed. Our spectral theorem makes this precise by controlling the growth of the design matrix (\cref{thm:spectral_growth}). On the unit ball, it gives a lower bound proportional to $\sqrt{t}$ on $\lambda_{\min}(V_t)$, holding with high probability at every round after a burn-in period; the general result controls the corresponding width in the action set geometry.

    \item \emph{A logarithmically capped potential.} To analyse this exploration mechanism, we introduce a logarithmically capped spectral potential that, unlike the standard log-determinant potential, caps contributions from well-explored directions. The resulting spectral bounds yield regret control without the usual elliptical-potential argument. At the standard width $\alpha=\beta$, this gives $\cO\bigl(\frac{M}{m}d\sqrt{T\log T}\bigr)$ regret, improving the horizon dependence of the classical $\cO(d\sqrt{T}\log T)$ guarantee by a factor of $\sqrt{\log T}$, up to iterated-logarithmic factors.
\end{itemize}
Moreover, we construct a counterexample confirming that assumptions on the action set are necessary: on a two-dimensional polygon, $\alpha$-LinUCB incurs linear regret with constant probability, for every sufficiently small $\alpha>0$ (\cref{prop:counter-egs}). We support our theoretical findings with experiments: on the unit ball, the observed regret and smallest-eigenvalue growth are consistent with the predicted $\sqrt{t}$ rates, even when the ellipsoids exclude the true parameter in most rounds.

\paragraph{Notation.}
We write $[T]:=\{1,\ldots,T\}$, $\I$ for the identity matrix in $\lR^{d\times d}$, and $\lambda_{\min}(V)$ for the smallest eigenvalue of a symmetric matrix $V$. For $u,v\in\lR^d$, let $\dotprod{u}{v}$ denote their inner product, and define $\norm{u}_2:=\sqrt{\dotprod{u}{u}}$ and $\norm{u}_V:=\sqrt{\dotprod{u}{Vu}}$ for $V\succ0$. We write $\mathbb{B}_2^d(r)$ for the $d$-dimensional closed Euclidean ball of radius $r$ centred at the origin.

\section{Related work}
\phantomsection\label{sec:Related_Work}

Our work is closest to that of \citet{abeille_when_why}, who use the same geometric condition to analyse Thompson sampling without forced optimism or posterior inflation. \citet{pmlr-v206-banerjee23b} also connect curvature with exploration, deriving growth of the smallest eigenvalue of the expected design matrix from an assumed $\cO(\sqrt{T})$ regret guarantee. We discuss these and further connections below.

\paragraph{Optimism and confidence sets in linear bandits.}
The classical $\cO(d\sqrt{T}\Log{T})$ regret analysis of LinUCB/OFUL combines self-normalised concentration with the elliptical-potential lemma \citep{Li-LINUCB,yassin}. Each contributes a $\sqrt{\Log{T}}$ factor, leaving a logarithmic gap to the $\Omega(d\sqrt{T})$ minimax lower bound \citep{dani2008stochastic,rusmevichientong2010linearly,lattimore2020bandit}. \citet{LowerBoundTor} proves that the continuous-time counterpart of the self-normalised concentration bound cannot be improved for general adaptive covariates. \citet{flynn2023improved} construct tighter confidence sequences using adaptive martingale mixtures, improving empirical performance while retaining $\cO(d\sqrt{T}\Log{T})$ worst-case regret. Our analysis uses the standard confidence bound but with a logarithmically capped potential.

Classical frequentist analyses of linear Thompson sampling also rely on optimism: posterior inflation ensures a constant probability of an optimistic draw, at the cost of an additional $\sqrt{d}$ factor in the regret bound \citep{agrawal2013thompson,abeille2017linear}. For finite-armed linear bandits, \citet{pmlr-v54-lattimore17a} show that standard optimism-based algorithms can be suboptimal. Subsequent asymptotically optimal algorithms explicitly optimise exploration allocations to account for the information provided by suboptimal actions \citep{hao2020adaptive,tirinzoni2020asymptoticallyoptimalprimaldualincremental}.

\paragraph{Action-set curvature and exploration.} When the action set is smooth and strongly convex, the played arm $x_t = \nabla J_{\cX}( \theta_t )$ moves continuously with the parameter, and perturbations of the estimate translate into exploration. The benefit of curvature is classical in online learning,
where follow-the-leader enjoys fast rates on strongly convex constraint sets for much the same reason \citep{NIPS2016_55a988df}. In stochastic bandits, \citet{pmlr-v206-banerjee23b} quantify this: for action sets with bounded curvature around the optimal arm, any algorithm with $\cO(\sqrt{T} )$ expected regret has a design matrix whose minimum eigenvalue grows as $\Omega( \sqrt{T})$ in expectation; in contrast  well-separated arms \citep{pmlr-v54-lattimore17a} force eventual logarithmic growth. \citet{abeille_when_why} exploit the same geometry, under the assumption we adopt here, to analyse randomised exploration without forced optimism or posterior inflation,
and show that Thompson sampling attains $\cO( d \sqrt{T} \Log{T} )$ regret.

Taking $\alpha = 0$ in our decision rule gives the greedy policy. Greedy methods are studied in contextual bandits, but only known to succeed under diversity of contexts, either assumed \citep{bastani2021mostly} or induced by smoothing \citep{kannan2018smoothed,raghavan2023greedy}. On a fixed action set no such diversity is available and, accordingly, our bound degrades as $\alpha \to 0$.

\paragraph{Inference under adaptive sampling.} Conditions on the growth of the design matrix play a central role in inference from adaptively collected data, beginning with \citet{lai1982least}. Ordinary least squares can have non-Gaussian asymptotic distributions and inflated estimation error under adaptive designs \citep{khamarou_tor_Near_optimal,kAdaptive}, motivating debiasing procedures that recover asymptotic normality under conditions on the data-collection process \citep{deshpande2018accurate,khamarou_tor_Near_optimal,ying2023adaptive,kAdaptive}. A complementary approach establishes stability of the design after deterministic normalisation;  \citet{praharaj2026avoidingpriceadaptivityinference} use this to obtain asymptotically valid Wald intervals for a regularised EXP4 algorithm in linear contextual bandits.

\section{Assumptions}\label{sec:assumptions}

We use the following two assumptions, placed on the noise and the action set, respectively. 

\paragraph{Stochasticity.}
Let $\cF_t$ be the $\sigma$-field generated by the history up to round $t$
together with the next action, so that $x_{t+1}$ is $\cF_t$-measurable;
namely, $\cF_t := \sigma(x_1, y_1, \ldots, x_t, y_t, x_{t+1})$. 

\begin{assumption}[\textbf{Sub-Gaussian Noise}]
\phantomsection\label{assump:subgaussian-noise}
    The noise variables are conditionally $\mu$-sub-Gaussian: for every
    $t \in [T]$,
    \[
        \lE[\eta_t \mid \cF_{t-1}] = 0
        \quad \text{and} \quad
        \lE\left[ \Exp{\upsilon \, \eta_t} \mid \cF_{t-1} \right]
        \leq \Exp{\frac{\mu^2 \upsilon^2}{2}}
        \quad \text{for all } \upsilon \in \lR.
    \]
\end{assumption}

The sub-Gaussian property ensures the concentration of self-normalized
martingales, which we shall leverage to control the cumulative effect of
noise under adaptive design.

\paragraph{The geometry of the action set.}
We consider a generic action set $\cX$ and impose the same assumptions as
in~\citet{abeille_when_why}, which are expressed in terms of the support
function of $\cX$,
\begin{align*}
    J_{\cX}(\theta) := \max_{x \in \cX} \dotprod{\theta}{x} .
\end{align*}
The optimal expected reward is denoted by
$J_{\star} := J_{\cX}(\theta^{\star})$. The support function plays a
central role in the analysis: whenever it is differentiable at
$\theta_t$, the action selected by the algorithm is
$x_t = \nabla J_{\cX}(\theta_t)$, so the instantaneous regret can be
written entirely in terms of $J_{\cX}$,
\begin{align*}
    r_t
    = J_{\cX}(\theta^{\star}) - \dotprod{\theta^{\star}}{x_t}
    = J_{\cX}(\theta^{\star}) - J_{\cX}(\theta_t)
      - \dotprod{\nabla J_{\cX}(\theta_t)}{\theta^{\star} - \theta_t}
    = D_{J_{\cX}}(\theta^{\star}, \theta_t),
\end{align*}
where $D_f( u, v ) := f( u ) - f( v ) - \dotprod{\nabla f( v )}{u - v}$
denotes the Bregman divergence of a differentiable convex function $f$.

\begin{assumption}[\textbf{Smooth, Strongly Convex Action Set}]
\phantomsection\label{ass:armset-int}
    There exist a norm $\norm{\cdot}_{\ast}$ on $\lR^d$ and constants
    $0 < m \leq M$ such that $J_{\cX}^2$ is differentiable on $\lR^d$ and,
    for all $\theta, \theta' \in \lR^d$,
    \begin{equation}\label{eq:ajp-int}
        \frac{m}{2} \norm{\theta - \theta'}_{\ast}^2
        \leq D_{J_{\cX}^2}(\theta, \theta')
        \leq \frac{M}{2} \norm{\theta - \theta'}_{\ast}^2;
    \end{equation}
    moreover, $\cX$ is absorbing, i.e., it contains a neighborhood of the
    origin.
\end{assumption}

\cref{ass:armset-int} admits a simple geometric interpretation: the boundary of $\cX$ must be uniformly curved, with neither flat faces nor corners. Polytopes are thus excluded. 

Strong convexity and smoothness are imposed on the squared support
function $J_{\cX}^2$ rather than on $J_{\cX}$ itself because its gradient,
\(
    \nabla J_{\cX}^2(\theta) = 2 J_{\cX}(\theta) \nabla J_{\cX}(\theta),
\)
remains well behaved as $\theta \to 0$, where it vanishes, whereas
$\nabla J_{\cX}$ admits no limit at the origin. The absorbing property
ensures that the multiplier $J_{\cX}(\theta)$ is positive for every
$\theta \neq 0$. \cref{ass:armset-int}  covers the following geometries:

\begin{itemize}
    \item \emph{Unit ball.} For $\cX = \mathbb{B}_2^d(1)$, the support
    function is $J_{\cX}(\theta) = \norm{\theta}_2$ and
    $D_{J_{\cX}^2}(\theta, \theta') = \norm{\theta - \theta'}_2^2$; hence
    \cref{ass:armset-int} holds with $\norm{\cdot}_{\ast} = \norm{\cdot}_2$
    and $m = M = 2$.
       
     \item \emph{Ellipsoids.} For
    $\cX = \{ x : x^{\top} \bA^{-1} x \leq 1 \} = \bA^{1/2} \mathbb{B}_2^d(1)$
    with a positive definite matrix $\bA$, the support function is
    $J_{\cX}(\theta) = \norm{\theta}_{\bA}$. Under the adapted norm
    $\norm{u}_{\ast} := \norm{\bA^{1/2} u}_2$,
    \(
        D_{J_{\cX}^2}(\theta, \theta')
        = \norm{\theta - \theta'}_{\bA}^2
        = \norm{\theta - \theta'}_{\ast}^2,
    \)
    so that, once again, $m = M = 2$.
    
 \item \emph{$C^2$ bodies.} More generally, the assumption holds for $\cX \subseteq \mathbb{B}_2^d(1)$ containing the origin in its interior with a $C^2$ boundary whose principal radii are bounded away from zero.
\end{itemize}
See~\cref{sec:armset-more} for more details, including the parameters $M,m$ for the $C^2$ body example.

\section{Main Results}
\phantomsection\label{sec:main_results}

We now present our regret upper bound for $\alpha$-LinUCB. Its statement uses the standard concentration width $\beta$ and the concentration event $\cE$ of \citet{yassin}, which we first reproduce.

\begin{restatable}[Confidence event; \citealp{yassin}]{restatableproposition}{propConf}
\phantomsection\label{prop:Yassin_conf}
Let \cref{assump:subgaussian-noise} hold, let $\norm{\theta^{\star}}_2
\leq S^{\star}$ and $\cX \subseteq \mathbb{B}_2^d(1)$, and let $\hat{\theta}_t$ be the ridge
estimate with regularisation $\lambda_0 > 0$. Fix a
horizon $T$ and a confidence level $\delta \in (0, 1)$, and define
\begin{align*}
\beta \;:&=\; \mu \sqrt{2 \Log{\frac{1}{\delta}}
+ d \Log{\Big( 1 + \frac{T}{d \lambda_0} \Big)}}
\;+\; \sqrt{\lambda_0}\, S^{\star} ,
\\
\cE \;:&=\; \Big\{\, \norm{\hat{\theta}_t - \theta^{\star}}_{V_t} \leq \beta
\ \ \text{for all } t \in [T] \,\Big\} .
\end{align*}
Then $\lP( \cE ) \geq 1 - \delta$.
\end{restatable}

The following regret bound holds for all $\alpha > 0$; the confidence width $\beta$ and the event $\cE$ are used in the analysis only.

\begin{restatable}[$\alpha$-LinUCB regret]{restatabletheorem}{thmAlphaUCBRegret}
\phantomsection\label{thm:alpha-UCB-Regret}
Let \cref{assump:subgaussian-noise,ass:armset-int} hold. On the event $\cE$, hence
with probability at least $1 - \delta$,
\[
\cR_T
\;=\; {\cO}\Big( \frac{M}{m}\, \frac{\alpha^2 + \beta^2}{\alpha}\,
\sqrt{dT } \Big) .
\]
\end{restatable}

\cref{thm:alpha-UCB-Regret} shows that $\alpha$-LinUCB has sublinear regret for every $\alpha > 0$,
including values for which $\cC_t( \alpha )$ is not a valid confidence set. The bound depends on the geometry of the action set through the ratio $M / m$, similarly to the bound of \citet{abeille_when_why} for randomised exploration algorithms. 

As $\alpha$ decreases, the set $\cC_t( \alpha )$ shrinks towards $\{ \hat{\theta}_t \}$ and the algorithm
approaches the greedy policy; accordingly, the guarantee degrades. Likewise, the bound deteriorates as $\alpha \to \infty$, when the algorithm fails to exploit known information.  Optimising the bound over $\alpha$ gives $\alpha = \beta$, and one recovers standard LinUCB, but with a regret bound of ${\cO}( ( M / m )\, d \sqrt{T \Log{T}} )$, a $\sqrt{\Log{T}}$ improvement over the $\cO( d \sqrt{T} \Log{T} )$ upper bound of \citet{yassin}.

The proof of \cref{thm:alpha-UCB-Regret} does not rely on the optimism inequality $\dotprod{\theta_t}{x_t} \geq \dotprod{\theta^{\star}}{x^{\star}}$, which is no longer guaranteed when $\theta^{\star} \notin \cC_t( \alpha )$ but instead examines the exploration dynamics of $\alpha$-LinUCB. The following bound shows that, in a  precise sense, the algorithm explores every direction for every $\alpha>0$. It controls the squared $\norm{\cdot}_{\ast}$-width of the unit $V_t$-ellipsoid,
\begin{align*}
    \omega_t := \max_{u \in \mathbb{B}_2^d(1)} \norm{V_t^{-1/2} u}_{\ast}^2
    = \max_{\norm{g}_{\ast}^{\circ} \leq 1} \norm{g}_{V_t^{-1}}^2 ,
\end{align*}
where $\norm{\cdot}_{\ast}^{\circ} := \sup_{\norm{v}_{\ast} \leq 1}
\dotprod{\cdot}{v}$ denotes the dual norm of $\norm{\cdot}_{\ast}$. When $\cX$ is the unit ball and $\norm{\cdot}_{\ast} = \norm{\cdot}_2$, one has $\omega_t = \max_{u \in \mathbb{B}_2^d(1)} \norm{V_t^{-1/2} u}_2^2 = \lambda_{\min}( V_t )^{-1}$, so that upper bounding $\omega_t$ amounts to lower bounding the smallest eigenvalue of the design, i.e. the growth of the least-explored direction.

\begin{restatable}[Spectral Growth]{restatabletheorem}{thmMinEigInt}
\phantomsection\label{thm:spectral_growth}
    Let \cref{assump:subgaussian-noise,ass:armset-int} hold, let $\cE$
    and $\beta$ be as in \cref{prop:Yassin_conf}, and let
    $T_B = \cO(\operatorname{poly}(\log T))$ be the burn-in threshold.
    Then, on the event $\cE$ (hence with probability at least
    $1 - \delta$), for all $T_B \leq t \leq T$,
    \[
        \omega_t
        = \max_{u \in \mathbb{B}_2^d(1)} \norm{V_t^{-1/2} u}_{\ast}^2
        \;\leq\; \frac{1}{c_0 \sqrt{t}},
        \qquad \text{where} \qquad
        c_0 = \Omega\Big( \frac{m \alpha}{J_{\star}\sqrt{d }} \Big).
    \]
\end{restatable}

The bound is \emph{deterministic on $\cE$} and holds \emph{at every
round} after the burn-in, not on average or asymptotically; it is stated
in the geometry of the assumption, with a constant proportional to $m
\alpha / ( J_{\star} \sqrt{d} )$; and it holds for \emph{every} $\alpha
> 0$, so the confidence set need not contain $\theta^{\star}$ for the
algorithm to explore. 

On $\cE$, the spectral growth bound shows immediately the following deviation bounds:
\[
    \norm{\theta_t - \hat{\theta}_t}_{\ast}
    \;\leq\; \alpha \sqrt{\omega_t}
    \;\leq\; \frac{\alpha}{\sqrt{c_0}\, t^{1/4}}
    \qquad \text{and} \qquad
    \norm{\hat{\theta}_t - \theta^{\star}}_{\ast}
    \;\leq\; \beta \sqrt{\omega_t}
    \;\leq\; \frac{\beta}{\sqrt{c_0}\, t^{1/4}} ,
\]
so that the $\norm{\cdot}_{\ast}$-radius of the confidence set and the
estimation error both decay at the rate $t^{-1/4}$, for all
$\alpha > 0$. In the unit-ball case, the bound reads
$\lambda_{\min}(V_t) = \Omega\big( \frac{\alpha}{
\sqrt{d}} \sqrt{t} \big)$. For $\alpha \asymp \beta \asymp
\sqrt{d \log T}$, this improves on the $\Omega\big( \sqrt{t} / (d \log
t) \big)$ lower bound for LinUCB implied by \citet{pmlr-v206-banerjee23b}.

\section{Proof sketch}
\phantomsection\label{sec:proof_sketch}

\begin{figure}[h]
\centering
\begin{tikzpicture}[x=1cm,y=1cm,font=\scriptsize,>=Latex,line cap=round,line join=round]
\definecolor{acc}{RGB}{0,109,119}
\definecolor{acc2}{RGB}{186,73,20}
\definecolor{soft}{RGB}{222,236,238}
\definecolor{grid}{RGB}{185,185,185}
\tikzset{
  panel/.style={draw=grid,rounded corners=2.5pt,fill=white,line width=0.4pt,minimum width=4.45cm,minimum height=4.6cm,anchor=south west},
  ptitle/.style={anchor=north west,font=\scriptsize\bfseries,inner sep=0pt},
  note/.style={font=\scriptsize,inner sep=1pt,align=left},
  tnote/.style={font=\fontsize{6}{7}\selectfont,inner sep=0.7pt,align=left},
  key/.style={font=\scriptsize,inner sep=2pt,fill=white,rounded corners=1.5pt,draw=acc,line width=0.5pt,align=center,minimum width=4.2cm},
  flow/.style={->,line width=0.9pt,color=acc},
  dot/.style={circle,fill=acc2,inner sep=0pt,minimum size=2.6pt},
}
\node[panel] (PA) at (0,0) {};
\node[panel] (PB) at (4.75,0) {};
\node[panel] (PC) at (9.5,0) {};
\draw[flow] (PA.east) -- (PB.west);
\draw[flow] (PB.east) -- (PC.west);

\begin{scope}[shift={(0,0)}]
  \node[ptitle] at (0.12,4.45) {(a) Capped vs elliptical potential};
  \draw[->,line width=0.5pt] (0.35,1.3) -- (4.3,1.3);
  \node[note,anchor=west] at (4.28,1.3) {$\lambda$};
  \draw[->,line width=0.5pt] (0.35,1.3) -- (0.35,4.0);
  \node[note,anchor=south west] at (0.4,3.95) {$\Phi(\lambda)$};
  \node[tnote,anchor=north] at (1.3,1.18) {$\lambda_0$};
  \draw[line width=0.4pt] (1.3,1.25) -- (1.3,1.35);
  \node[tnote,anchor=north,text=gray] at (4.05,1.18) {(log)};
  \draw[acc,densely dotted,line width=0.5pt] (0.35,3.6) -- (4.2,3.6);
  \node[tnote,text=acc,anchor=south west] at (0.42,3.62) {cap $\vartheta\Log{8\Gamma}$};
  \draw[->,acc2,densely dashed,line width=0.7pt] (1.3,1.3) -- (3.49,3.95);
  \node[dot] at (3.2,3.6) {};
  \node[tnote,text=acc2,anchor=west] at (3.28,3.35) {passes the cap,};
  \node[tnote,text=acc2,anchor=west] at (3.28,3.17) {then $\to\infty$};
  \node[tnote,text=acc2,anchor=east] at (2.5,2.97) {$\Phi^{\mathrm{ell}}(\lambda)=\log(\sfrac{\lambda}{\lambda_0})$};
  \draw[acc,line width=0.9pt,domain=0.35:4.15,samples=100]
    plot (\x,{1.3+0.5*( ln(1+10^((\x-0.35)/0.95-1)) - ln(1+10^((\x-0.35)/0.95-1)/99) )});
  \node[tnote,text=acc,anchor=west] at (2.75,2.35) {$\Phi_\vartheta(\lambda)=h(\lambda)$};
  \node[tnote,text=acc,anchor=west] at (2.75,2.17) {saturates};
  \node[key,anchor=south] at (2.225,0.1) {$\Phi_\vartheta\le d\,\vartheta\Log{8\Gamma}$; $\ \Phi_{\mathrm{ell}}$ unbounded};
\end{scope}

\begin{scope}[shift={(4.75,0)}]
  \node[ptitle] at (0.12,4.45) {(b) Step 2: the budget of a phase};
  \draw[->,line width=0.5pt] (0.35,1.3) -- (4.3,1.3);
  \node[note,anchor=west] at (4.28,1.3) {$t$};
  \draw[->,line width=0.5pt] (0.35,1.3) -- (0.35,4.0);
  \node[note,anchor=south west] at (0.4,3.95) {$\Phi_\vartheta(V_t)$};
  \draw[acc,densely dotted,line width=0.5pt] (0.35,3.55) -- (4.2,3.55);
  \node[tnote,text=acc,anchor=south west] at (0.42,3.57) {cap: $d\,\vartheta\Log{8\Gamma}$};
  \foreach \x/\h in {0.75/1.8,1.35/2.15,1.95/2.5,3.1/3.27}{ \fill[acc!75] (\x-0.13,1.3) rectangle (\x+0.13,\h); }
  \node[note,text=acc,anchor=center] at (2.52,2.8) {$\cdots$};
  \fill[acc2!25] (3.42,1.3) rectangle (3.68,3.66);
  \draw[acc2,line width=0.5pt] (3.42,1.3) rectangle (3.68,3.66);
  \node[tnote,text=acc2,anchor=south] at (3.55,3.7) {$>$ cap};
  \node[tnote,text=acc2,anchor=west] at (3.72,3.3) {impossible};
  \foreach \x/\a/\b in {1.05/1.8/2.15,1.65/2.15/2.5}{ \draw[<->,acc2,line width=0.7pt,>={Latex[length=1.8pt]}] (\x,\a) -- (\x,\b); }
  \node[tnote,text=acc2,anchor=east] at (1.0,1.975) {$\alpha^2/\vartheta\lesssim$};
  \node[tnote,text=acc2,anchor=east] at (1.6,2.325) {$\alpha^2/\vartheta\lesssim$};
  \node[tnote,text=acc2,anchor=west] at (0.45,3.3) {each active round adds};
  \node[tnote,text=acc2,anchor=west] at (0.45,3.11) {$\Delta\Phi_\vartheta\gtrsim\alpha^2/\vartheta$};
  \foreach \x/\n in {0.75/{\tau},1.35/{\tau{+}1},1.95/{\tau{+}2},2.52/{\cdots},3.55/{\tau{+}N}}{ \node[tnote,anchor=north] at (\x,1.2) {$\n$}; }
  \draw[decorate,decoration={brace,amplitude=2pt,mirror},line width=0.4pt] (0.62,0.98) -- (3.68,0.98);
  \node[tnote,anchor=north] at (2.12,0.9) {active phase: $\lambda_{t,d}\le2\vartheta$};
  \node[key,anchor=south] at (2.225,0.1) {$\Rightarrow$ fewer than $N$ active rounds};
\end{scope}

\begin{scope}[shift={(9.5,0)}]
  \node[ptitle] at (0.12,4.45) {(c) Step 3: doubling $\Rightarrow\sqrt{t}$};
  \draw[->,line width=0.5pt] (0.3,1.3) -- (4.3,1.3);
  \node[note,anchor=west] at (4.28,1.3) {$t$};
  \draw[->,line width=0.5pt] (0.3,1.3) -- (0.3,3.88);
  \node[note,anchor=south] at (0.3,3.9) {$\lambda_{t,d}$};
  \draw[line width=0.4pt] (0.25,1.9) -- (0.35,1.9);  \node[tnote,anchor=east] at (0.23,1.9) {$\vartheta/2$};
  \draw[line width=0.4pt] (0.25,2.5) -- (0.35,2.5);  \node[tnote,anchor=east] at (0.23,2.5) {$\vartheta$};
  \draw[line width=0.4pt] (0.25,3.7) -- (0.35,3.7);  \node[tnote,anchor=east] at (0.23,3.7) {$2\vartheta$};
  \fill[soft,opacity=0.9] (1.08,2.5) rectangle (3.16,3.7);
  \node[tnote,text=acc,anchor=center] at (2.12,3.3) {active rounds};
  \node[tnote,text=acc,anchor=center] at (2.12,3.1) {$\lambda_{t,d}\le2\vartheta$};
  \draw[acc,line width=1pt] (0.43,1.6) -- (0.56,1.6) -- (0.56,1.9) -- (1.08,1.9) -- (1.08,2.5) -- (3.16,2.5) -- (3.16,3.7) -- (4.2,3.7);
  \node[dot,fill=acc,minimum size=2.4pt] at (3.16,3.7) {};
  \node[tnote,text=acc,anchor=south east] at (3.1,3.76) {$\lambda_{\tau+N,d}\ge2\vartheta$};
  \node[tnote,anchor=north] at (1.08,1.2) {$\tau$};
  \node[tnote,anchor=north] at (3.16,1.2) {$\tau{+}N$};
  \draw[decorate,decoration={brace,amplitude=2pt,mirror},line width=0.4pt] (1.1,0.95) -- (3.14,0.95);
  \node[tnote,anchor=north] at (2.12,0.87) {$N=\lceil K\vartheta^2\rceil$ rounds};
  \draw[line width=0.8pt,domain=0.625:4.2,samples=50] plot (\x,{1.3+0.24*sqrt((\x-0.3)/0.13)});
  \node[note,anchor=south east] at (4.2,2.62) {$c_0\sqrt{t}$};
  \node[key,anchor=south] at (2.225,0.1) {$\Rightarrow\lambda_{t,d}\ge c_0\sqrt{t}$};
\end{scope}
\end{tikzpicture}
\caption{The three steps of the proof of \cref{thm:spectral_growth} in the
unit-ball case.
}
\label{fig:sketch}
\end{figure}
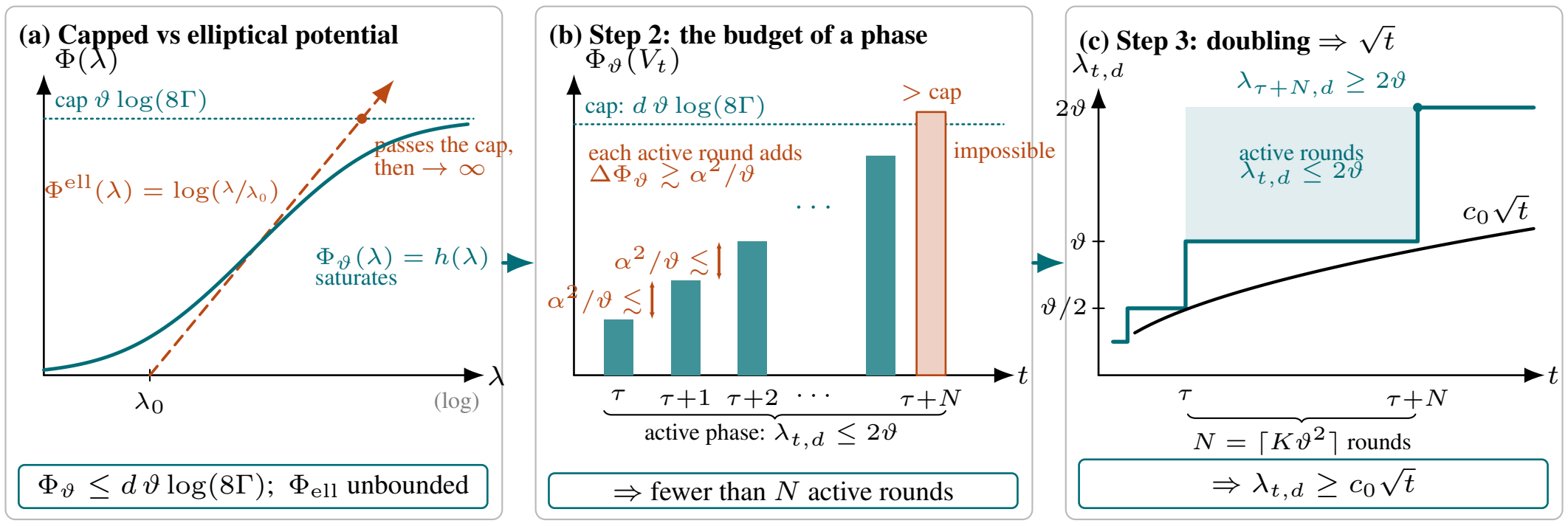

The detailed proofs of \cref{thm:spectral_growth,thm:alpha-UCB-Regret}
are given in \cref{sec:spectral_growth_proof,sec:regret-proof}. The
argument has two parts: an energy argument, which shows that optimizing
over $\cC_t( \alpha )$ forces the design matrix to grow in every
direction, and a regret decomposition, which uses this growth in place
of the optimism inequality of the standard analysis. We now sketch the two proofs, starting with spectral growth.

\subsection{Spectral growth}
\label{sec:sketch_spectral}

For ease of presentation, we sketch the proof in the unit-ball case, i.e.\ $\cX = \mathbb{B}_2^d(1)$. Take $\|\cdot\|_* = \|\cdot\|_2$, so that $\omega_t = \lambda_{t,d}^{-1}$. Assume that $\theta^\star \neq 0$. The proof follows an energy argument, which tracks the amount of energy that $\alpha$-LinUCB injects into the least-explored directions. Let $\lambda_{t,1} \geq \dots \geq \lambda_{t,d}$ denote the eigenvalues of $V_t$, and $v_{t,1}, \dots, v_{t,d}$ the eigenvectors. We define the energy of an action $x_t$ in the $i$th direction, $i \in [d]$, as $x_{t,i}^2 / \lambda_{t,i}$, where $x_{t,i} := \dotprod{x_t}{v_{t,i}}$. The argument has three steps, summarized in~\cref{fig:sketch}.

\emph{Step 1.} We first define a \emph{logarithmically capped potential}, rather than the usual elliptical potential (denoted by $\Phi_{\mathrm{ell}}$ in~\cref{fig:sketch}). Fix a reference scale $\vartheta > 0$ and let $\Gamma \geq 1$ (with a specific value of $\Gamma$ fixed in~\cref{sec:constant_def}).
\begin{definition}[Capped potential]
For the scale $\vartheta$, set $L := 8 \Gamma \vartheta$ and define,
for every $V \succeq 0$,
\begin{equation}
\label{eq:potential-def}
\Phi_{\vartheta}( V ) := \Tr\, h( V ) ,
\qquad
h( \lambda ) := \vartheta \Log{ \frac{L\, ( \lambda + \vartheta )}
{\vartheta\, ( \lambda + L )} } ,
\qquad \lambda \geq 0 .
\end{equation}
\end{definition}
The map $\Phi_{\vartheta}$ sends positive semidefinite matrices to
$\lR$ and is a rescaled trace in which each eigenvalue is
counted through $h$. It can alternatively be expressed as a difference of $\vartheta$- and $L$-regularised log-determinants,
\[
    \Phi_{\vartheta}(V) = \vartheta \big( \log\det(V + \vartheta \I) - \log\det(V + 8\Gamma \vartheta \I ) + d\Log{8\Gamma} \big),
\]
where the additive constant $d\Log{8\Gamma}$ ensures that $\Phi_{\vartheta}(0) = 0$. The growths of the two log-determinant terms along directions of $V$ that significantly exceed $8\Gamma\vartheta$ essentially cancel out, so that the potential is capped; indeed
\[
0 \;\leq\; \Phi_{\vartheta}( V ) \;\leq\; d\, \vartheta \Log{8 \Gamma}
\qquad \text{for every } V \succeq 0 .
\]

\emph{Step 2.} After a burn-in of polylogarithmic length $T_B$ (see~\cref{def:ladder-int}), on the
event $\cE$, the whole set $\cC_t(\alpha)$ lies in the neighbourhood
$\{ \theta : \norm{\theta - \theta^\star} \leq \norm{\theta^\star}/2 \}$
(\cref{prop:burnin-int}). In particular,
$\frac{1}{2}\norm{\theta^\star} \leq \norm{\theta} \leq
\frac{3}{2}\norm{\theta^\star}$ for all $\theta \in \cC_t(\alpha)$, and the
played arms satisfy $\langle x_t, x^\star \rangle \geq 1/2$ for every
$t \geq T_B$. Since $x_t \in \mathbb{B}^d_2(1)$,
\[
 \norm{\theta_t} = \langle \hat\theta_t, x_t \rangle + \alpha \norm{x_t}_{V_t^{-1}} \leq \norm{\hat\theta_t} +\alpha \norm{x_t}_{V_t^{-1}} \implies \alpha \norm{x_t}_{V_t^{-1}} \geq \norm{\theta_t} - \norm{\hat\theta_t},
\]
where the equality follows from the definition of $(x_t, \theta_t)$ together with $x_t = \theta_t / \norm{\theta_t}$. Next, by definition, $\hat\theta_t \pm \alpha v_{t,d}/\sqrt{\lambda_{t,d}} \in \cC_t(\alpha)$, and since $\theta_t$ maximises the norm over $\cC_t(\alpha)$, averaging over the two signs gives
\[
\norm{\theta_t}^2 \geq \frac{1}{2} \norm{\hat\theta_t + \frac{\alpha}{\sqrt{\lambda_{t,d}}} v_{t,d}}^2 + \frac{1}{2}\norm{\hat\theta_t - \frac{\alpha}{\sqrt{\lambda_{t,d}}} v_{t,d}}^2 = \norm{\hat\theta_t}^2 + \frac{\alpha^2}{\lambda_{t,d}}.
\]
Combining the above with the previous display, we have
\[
    \alpha \norm{x_t}_{V_t^{-1}} \geq \norm{\theta_t} - \norm{\hat\theta_t} = \frac{\norm{\theta_t}^2 - \norm{\hat\theta_t}^2}{\norm{\theta_t} + \norm{\hat\theta_t}}  \geq \frac{\alpha^2}{3\norm{\theta^\star}\lambda_{t,d}},
\]
where the last inequality uses that $\theta_t, \hat\theta_t \in \cC_t(\alpha)$, so that $\norm{\theta_t} + \norm{\hat\theta_t} \leq 3\norm{\theta^\star}$ by assumption.

Fix a time-step $\tau$ and let $\vartheta := \lambda_{\tau,d}$, the smallest eigenvalue at round $\tau$, and let $\tau'$ be the last time-step for which $\lambda_{t,d} \leq 2\vartheta$. Call $t \in [\tau, \tau']$ the active time-steps.

We show in~\cref{lem:potential-int} that for every round $t \geq \tau$,
\[
\Phi_{\vartheta}( V_{t+1} ) - \Phi_{\vartheta}( V_t )
\;\geq\; \frac{\vartheta}{4} \sum_{\substack{i \geq 2 \\ \lambda_{t,i} < \Gamma \vartheta}} \frac{x^2_{t,i}}{\lambda_{t,i}} ,
\]
a straightforward algebraic bound. Now, using our inequality on $\alpha \norm{x_t}_{V_t^{-1}}$,

\begin{equation}\label{eq:engine-lb}
    \frac{x^2_{t,1}}{\lambda_{t,1}}
    + \sum_{\substack{i \geq 2 \\ \lambda_{t,i} \geq \Gamma \vartheta}} \frac{x^2_{t,i}}{\lambda_{t,i}}
    + \sum_{\substack{i \geq 2 \\ \lambda_{t,i} < \Gamma \vartheta}} \frac{x^2_{t,i}}{\lambda_{t,i}}
    = \norm{x_t}_{V_t^{-1}}^2
    \;\geq\; \frac{\alpha^2}{9\norm{\theta^\star}^2\lambda_{t,d}^2}
\end{equation}

If we show that the first two terms are small relative to the lower bound, we can conclude that
\begin{equation}\label{eq:potential-growth}
    \Phi_{\vartheta}( V_{t+1} ) - \Phi_{\vartheta}( V_t )  \gtrsim \frac{\vartheta \alpha^2}{\norm{\theta^\star}^2\lambda_{t,d}^2},
\end{equation}
where $\gtrsim$ denotes inequality up to absolute constants; this is
precisely what is established in \cref{lem:band-int,cor:band-int}.
Indeed, let $T_B$ denote the first time-step after the burn-in, which
ensures that $\langle x_t, x^\star \rangle \geq 1/2$ for every
$t \geq T_B$, and assume $t \geq 2T_B$. Then
\[
    \lambda_{t,1} \geq \norm{x^\star}^2_{V_t} \geq \sum_{s=T_B}^{t-1} \langle x_s, x^\star \rangle^2 \geq \frac{t-T_B}{4} \geq \frac{t}{8} \implies \frac{x^2_{t,1}}{\lambda_{t,1}} \leq \frac{1}{\lambda_{t,1}} \leq \frac{8}{t}.
\]
The upper bound $8/t$ is negligible relative to the lower bound of \eqref{eq:engine-lb} as long
as $\lambda_{t,d} = o(\sqrt{t})$; and once that is no longer true, the
conclusion of the theorem already holds at round $t$. For the remaining term, a less immediate
argument (\hyperref[step:band-b]{step~(b)} of the proof of~\cref{lem:band-int}) yields
$\sum_{i \geq 2} x^2_{t,i} \leq C_T/\lambda_{t,d}$, where $C_T > 0$ is
polylogarithmic in $T$, and so
\[
    \sum_{\substack{i \geq 2 \\ \lambda_{t,i}  \geq \Gamma \vartheta}} \frac{x^2_{t,i}}{\lambda_{t,i}} \leq  \frac{2 C_T}{\Gamma \lambda_{t,d}^2}.
\]
The choice of $\Gamma$ is precisely what makes this upper bound small relative to the lower bound of (\ref{eq:engine-lb}).

\emph{Step 3.} Set $N := \lceil K \vartheta^2 \rceil$ with $K \gtrsim
d \Log{8 \Gamma} / \alpha^2$, and suppose that all rounds $\tau, \dots,
\tau + N - 1$ are active. By \eqref{eq:potential-growth} with
$\lambda_{t,d} \leq 2 \vartheta$, each of them increases the potential by
$\gtrsim \alpha^2 / \vartheta$ (absorbing $\norm{\theta^{\star}}$ into
the constant), hence
\begin{align*}
    \Phi_{\vartheta}( V_{\tau + N} ) - \Phi_{\vartheta}( V_\tau )
    \;\gtrsim\; N\, \frac{\alpha^2}{\vartheta}
    \;>\; \frac{d \Log{8 \Gamma}}{\alpha^2}\, \vartheta^2 \cdot
    \frac{\alpha^2}{\vartheta}
    \;=\; d\, \vartheta \Log{8 \Gamma} ,
\end{align*}
which, for a suitable choice of $K$, exceeds the cap
$d \vartheta \Log{8 \Gamma}$ of the potential: a contradiction. Hence the
active phase ends before round $\tau + N$, i.e.\ the smallest eigenvalue
doubles within $N = O( d\, \vartheta^2 \Log{8 \Gamma} / \alpha^2 )$ rounds:
\(
    \lambda_{\tau + N, d} \;\geq\; 2 \lambda_{\tau,d}.
\)
Finally, a backward induction on the first round $t$ at which
$\lambda_{t,d} < c_0 \sqrt{t}$ would fail turns this doubling into
$\lambda_{t,d} \geq c_0 \sqrt{t}$ (see~\cref{prop:induction-int}).

While we chose to illustrate the proof sketch in the unit-ball case,
the appendix presents the proof in the general case, in the geometry of
$\norm{\cdot}_{\ast}$. The key is a change of variables in the spirit of
John's theorem (see~\cref{sec:distortion}): the Euclidean structure
needed for the spectral bookkeeping is chosen \emph{adapted} to
$\norm{\cdot}_{\ast}$, so that the two norms are compared only through
the distortion $D_{\ast} \leq \min\{ \sqrt{d}, \sqrt{M / m} \}$ of the
unit ball of $\norm{\cdot}_{\ast}$, which enters the bounds only inside
logarithms.

\subsection{From spectral growth to regret}
\label{sec:sketch_regret}

The argument has two steps. First, after the polylogarithmic burn-in
phase $T_B$(see~\cref{def:ladder-int}), the parameters $\theta_t$ lie in a neighbourhood of
$\theta^{\star}$ on which $J_{\cX}$ is comparable to $J_{\star}$
(\cref{prop:burnin-int}), and the instantaneous regret is controlled by
the Bregman divergence of $J_{\cX}^2$ (\cref{lem:regret-decomposition}):
\[
r_t \;\leq\; \frac{D_{J_{\cX}^2}( \theta^{\star}, \theta_t )}{J_{\star}} .
\]
This upper bound does not require $\theta^{\star} \in \cC_t( \alpha )$.
Second, the divergence is bounded by the width $\omega_t$, which
\cref{thm:spectral_growth} controls; summing over rounds then bounds
the regret directly, and in particular the elliptical potential lemma,
with the additional $\sqrt{\Log{T}}$ factor it introduces, is never
used to sum the widths.

In detail, the horizon is split at the burn-in time $T_B$. Before
$T_B$, the instantaneous regret $r_t$ is bounded by a constant, owing
to the boundedness of $\theta^{\star}$ and of the action set. From
$T_B$ on, the smoothness half of~\eqref{eq:ajp-int} and the definition
of the confidence sets reduce the divergence to a multiple of
$\omega_t$, and \cref{thm:spectral_growth} bounds the sum of the widths
by $2 \sqrt{T} / c_0$. On the event $\cE$,
\begin{align*}
\cR_T
&\leq \cO( T_B )
+ \sum_{t = T_B}^{T} \frac{D_{J_{\cX}^2}( \theta^{\star}, \theta_t )}{J_{\star}}
&& (\text{\cref{lem:regret-decomposition}}) \\
&\leq \cO( T_B )
+ \frac{M}{2 J_{\star}} \sum_{t = T_B}^{T} \norm{\theta^{\star} - \theta_t}_{\ast}^2
&& (\text{\cref{ass:armset-int}}) \\
&\leq \cO( T_B )
+ \frac{M}{J_{\star}} \sum_{t = T_B}^{T} \Big( \norm{\theta^{\star} - \hat{\theta}_t}_{\ast}^2
+ \norm{\hat{\theta}_t - \theta_t}_{\ast}^2 \Big)
&& \big( ( a + b )^2 \leq 2 a^2 + 2 b^2 \big) \\
&\leq \cO( T_B )
+ \frac{M ( \alpha^2 + \beta^2 )}{J_{\star}} \sum_{t = T_B}^{T} \omega_t
&& \big( \theta^{\star} \in \cC_t( \beta ),\ \theta_t \in \cC_t( \alpha ) \big) \\
&\leq \cO( T_B )
+ \frac{M ( \alpha^2 + \beta^2 )}{J_{\star}\, c_0} \sum_{t = T_B}^{T} \frac{1}{\sqrt{t}}
&& (\text{\cref{thm:spectral_growth}}) \\
&\leq \cO( T_B )
+ \frac{2 M ( \alpha^2 + \beta^2 )}{J_{\star}\, c_0}\, \sqrt{T}
= \cO\Big( \frac{M}{m}\, \frac{\alpha^2 + \beta^2}{\alpha}\,
\sqrt{dT} \Big) ,
\end{align*}
where the last step uses $c_0 = \Omega( m \alpha / ( J_{\star} \sqrt{d} ) )$
and $T_B = \cO( \operatorname{poly}( \log T ) )$.

\section{Curvature matters: a failure case}
\phantomsection\label{sec:counter_egs}

\begin{wrapfigure}{r}{0.4\textwidth}
\vspace{-1.2\baselineskip}
\centering
\resizebox{\linewidth}{!}{%
\begin{tikzpicture}[x=1cm,y=1cm,font=\scriptsize,>=Latex,line cap=round,line join=round]
\definecolor{acc}{RGB}{0,109,119}
\definecolor{acc2}{RGB}{186,73,20}
\definecolor{acc3}{RGB}{84,58,140}
\definecolor{soft}{RGB}{222,236,238}
\tikzset{
  tnote/.style={font=\fontsize{8}{9}\selectfont,inner sep=0.7pt,align=left},
  dot/.style={circle,fill=black,inner sep=0pt,minimum size=3.2pt},
  adot/.style={circle,fill=acc,inner sep=0pt,minimum size=4pt},
  rdot/.style={circle,fill=acc2,inner sep=0pt,minimum size=3.4pt},
}
\fill[soft] (5.8,2.3) -- (3.4,3.5) -- (1.0,2.3) -- (3.4,1.1) -- cycle;
\draw[acc,line width=0.8pt] (5.8,2.3) -- (3.4,3.5) -- (1.0,2.3) -- (3.4,1.1) -- cycle;
\node[tnote,text=acc,anchor=west] at (2,1.85) {$\cX_a$};
\draw[->,line width=1pt] (0.5,2.3) -- (6.4,2.3);
\node[tnote,anchor=west] at (6.42,2.3) {$x_1$};
\draw[->,line width=1pt] (3.4,0.7) -- (3.4,5.1);
\node[tnote,anchor=south] at (3.4,5.12) {$x_2$};
\node[dot] at (3.4,2.3) {};
\node[tnote,anchor=north east] at (3.36,2.26) {$0$};
\node[adot] at (5.8,2.3) {};
\node[tnote,text=acc,anchor=north] at (5.8,2.2) {$e_1$};
\node[adot] at (1.0,2.3) {};
\node[tnote,text=acc,anchor=north] at (1.0,2.2) {$-e_1$};
\node[adot] at (3.4,3.5) {};
\node[tnote,text=acc,anchor=west] at (2,3.56) {$x^\star=ae_2$};
\node[adot] at (3.4,1.1) {};
\node[tnote,text=acc,anchor=east] at (3.3,1.16) {$-ae_2$};
\draw[->,acc2,line width=1.1pt] (3.4,2.3) -- (4.0,4.7);
\node[rdot] at (4.0,4.7) {};
\node[tnote,text=acc2,anchor=west] at (4.08,4.72) {$\theta^\star=(\varepsilon,1)$};
\draw[acc2,densely dashed,line width=0.6pt] (4.0,4.7) -- (4.0,2.3);
\node[rdot] at (4.0,2.3) {};
\draw[acc2,densely dashed,line width=0.6pt] (4.0,4.7) -- (3.4,4.7);
\node[rdot] at (3.4,4.7) {};
\node[tnote,text=acc2,anchor=east] at (3.32,4.7) {$\dotprod{\theta^\star}{e_2}=1$};
\draw[<->,acc3,line width=1pt] (3.25,2.32) -- (3.25,3.48);
\node[tnote,text=acc3,anchor=east] at (3.18,2.9) {$a=\dotprod{\theta^\star}{ae_2}$};
\draw[<->,acc2,line width=1pt] (3.42,2.15) -- (3.98,2.15);
\node[tnote,text=acc2,anchor=north west] at (3.4,2.08) {$\varepsilon=\dotprod{\theta^\star}{e_1}$};
\end{tikzpicture}%
}
\caption{Counter-example.}
\label{fig:counter_egs}
\vspace{-1\baselineskip}
\end{wrapfigure}
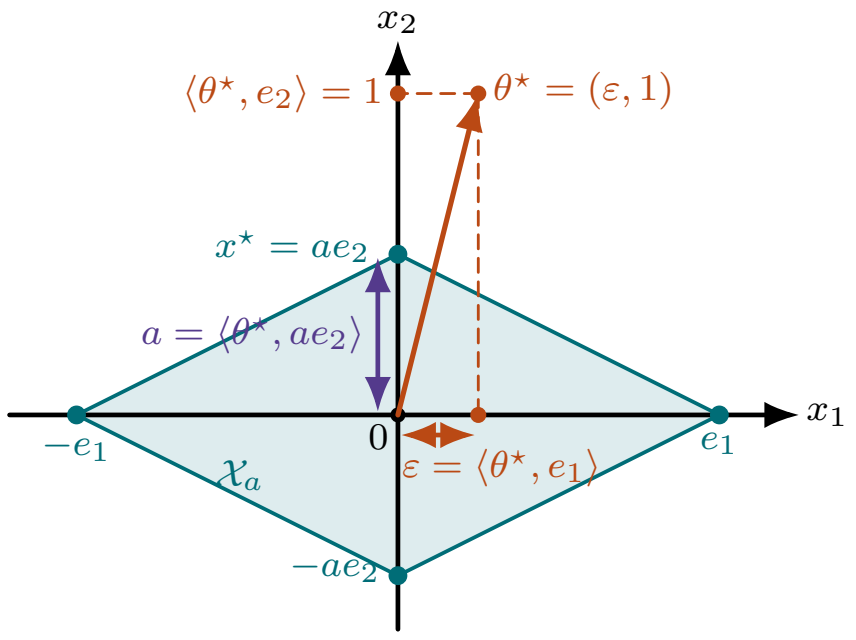

\cref{thm:spectral_growth,thm:alpha-UCB-Regret} hold for \emph{every} radius $\alpha > 0$, and~\cref{sec:proof_sketch} explained what the curvature imposed
by~\cref{ass:armset-int} does for the algorithm: a perturbation of the
optimistic parameter is transmitted to the played arm. On a body with a
corner this transmission stops, and the next proposition shows that the
conclusion of \cref{thm:spectral_growth,thm:alpha-UCB-Regret} then fails
for every small radius: $\alpha$-LinUCB locks on a suboptimal vertex
with probability bounded away from zero, the design matrix does not
grow, and the regret is linear. The instance is the simplest possible (presented in~\cref{fig:counter_egs}):
in $\lR^2$, with canonical basis $( e_1, e_2 )$, for $a \in ( 0, 1 )$,
let
\[
    \cX_a := \mathrm{conv} \{ e_1, -e_1, a e_2, -a e_2 \} ,
    \qquad
    \theta^{\star} := ( \varepsilon, 1 ) \quad \text{with } 0 <
    \varepsilon < a ,
    \qquad
    \eta_t \overset{\text{i.i.d.}}{\sim} \mathcal{N}( 0, \mu^2 ) ,
\]
where $\operatorname{conv}$ denotes the convex hull; that is, $\cX_a$ is
the diamond with vertices $\pm e_1$ and $\pm a e_2$. 

The set $\cX_a$ is a convex body containing the origin in its interior
and contained in $\mathbb{B}_2^2(1)$, whose support function is
$J_{\cX_a}( \theta ) = \max\{ | \theta_1 |, a | \theta_2 | \}$ for
$\theta = ( \theta_1, \theta_2 )$; it violates \cref{ass:armset-int} at
its four corners and four flat faces (\cref{fig:armset-int}(b)). The
optimal arm is $x^{\star} = a e_2$, with $J_{\star} = a$, and the gap of
the two vertices $\pm e_1$ is $\Delta := a - \varepsilon > 0$.

\begin{restatable}[Linear regret on a polytope]{restatableproposition}{PropCounterEgs}
\phantomsection\label{prop:counter-egs}
In the instance above, for every $\lambda_0 > 0$, every horizon $T$ and
every radius
\[
    0 < \alpha \leq \alpha_0 := \frac{( 1 - a^2 ) \sqrt{\lambda_0}}{2 a}\,
    \varepsilon ,
\]
with probability at least
\[
    p_0 := \frac{1}{2} \Big( 1 - \exp\Big( - \frac{( 1 - a^2 ) \lambda_0
    \varepsilon^2}{2 a^2 \mu^2} \Big) \Big) > 0 ,
\]
$\alpha$-LinUCB plays $x_t \in \{ e_1, -e_1 \}$ at every round $t \in [
T ]$. On this event $\lambda_{\min}( V_t ) = \lambda_0$ for all $t$ and
$\cR_T \geq \Delta T$; in particular $\mathbb{E}[ \cR_T ] \geq p_0
\Delta T$.
\end{restatable}

The proof is given in~\cref{sec:counter-exmaple}. For instance, $a = 1/2$,
$\varepsilon = 1/4$, $\lambda_0 = 1$ and Gaussian noise with $\mu = 1/8$
give $p_0 > 0.49$: for every $\alpha \leq 3/16$, half of the runs never
play the optimal arm and suffer $\cR_T \geq T/4$. The curvature imposed
by \cref{ass:armset-int} is precisely what excludes such lock-ins,
however small the radius; we do not claim that it is necessary, only
sufficient for our analysis.

\section{Numerical Simulations}
\label{sec:exp}

\begin{figure*}[h]
    \centering
    \begin{minipage}{1\textwidth}
        \centering

        \begin{subfigure}[t]{0.24\linewidth}
            \centering
            \includegraphics[width=\linewidth]{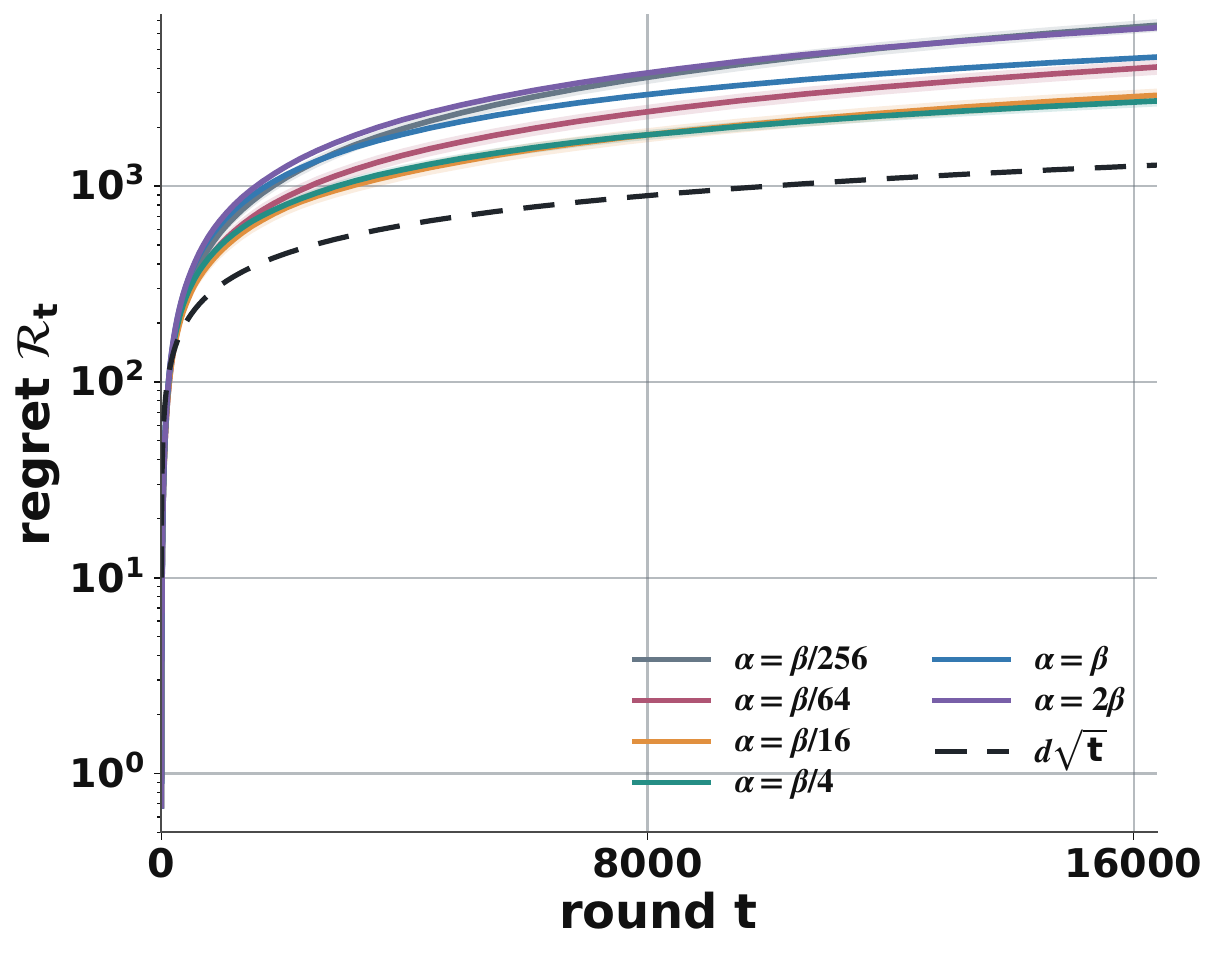}
            \caption{Regret}
            \label{fig:beta20-regret}
        \end{subfigure}\hfill
        \begin{subfigure}[t]{0.24\linewidth}
            \centering
            \includegraphics[width=\linewidth]{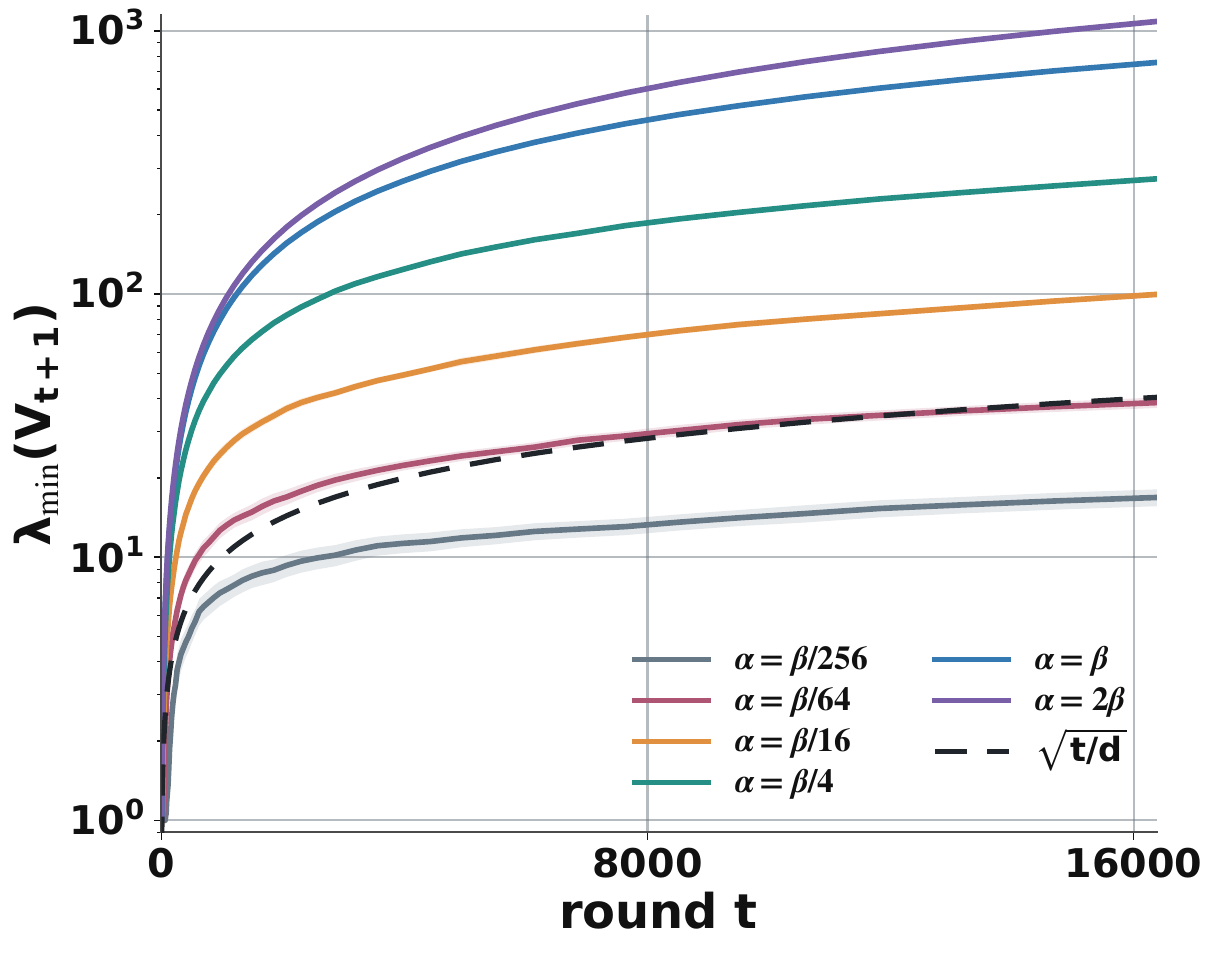}
            \caption{Minimum eigenvalue}
            \label{fig:beta20-eigenvalue}
        \end{subfigure}\hfill
        \begin{subfigure}[t]{0.24\linewidth}
            \centering
            \includegraphics[width=\linewidth]{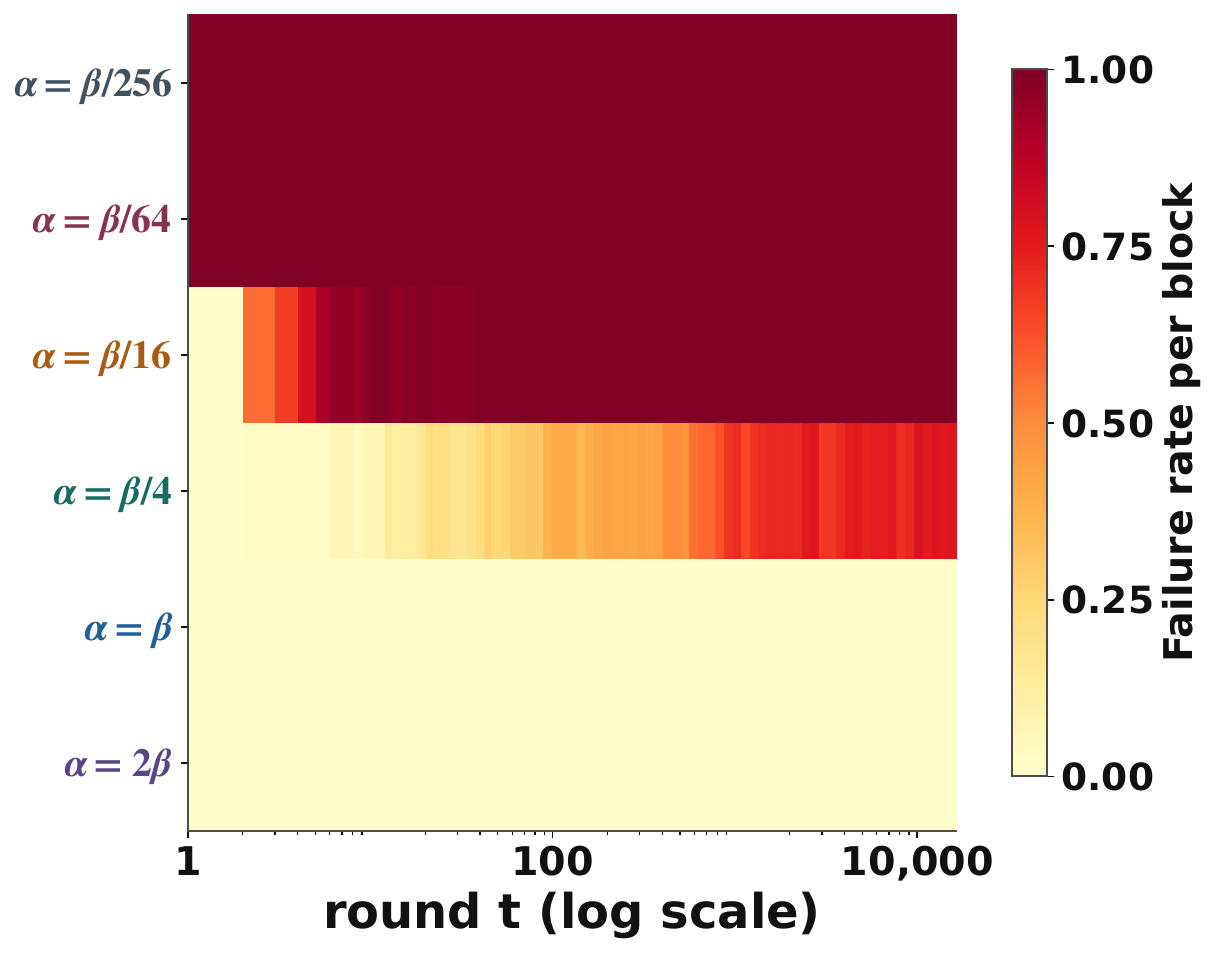}
            \caption{Confidence failures}
            \label{fig:beta20-failures}
        \end{subfigure}\hfill
        \begin{subfigure}[t]{0.24\linewidth}
            \centering
            \includegraphics[width=\linewidth]{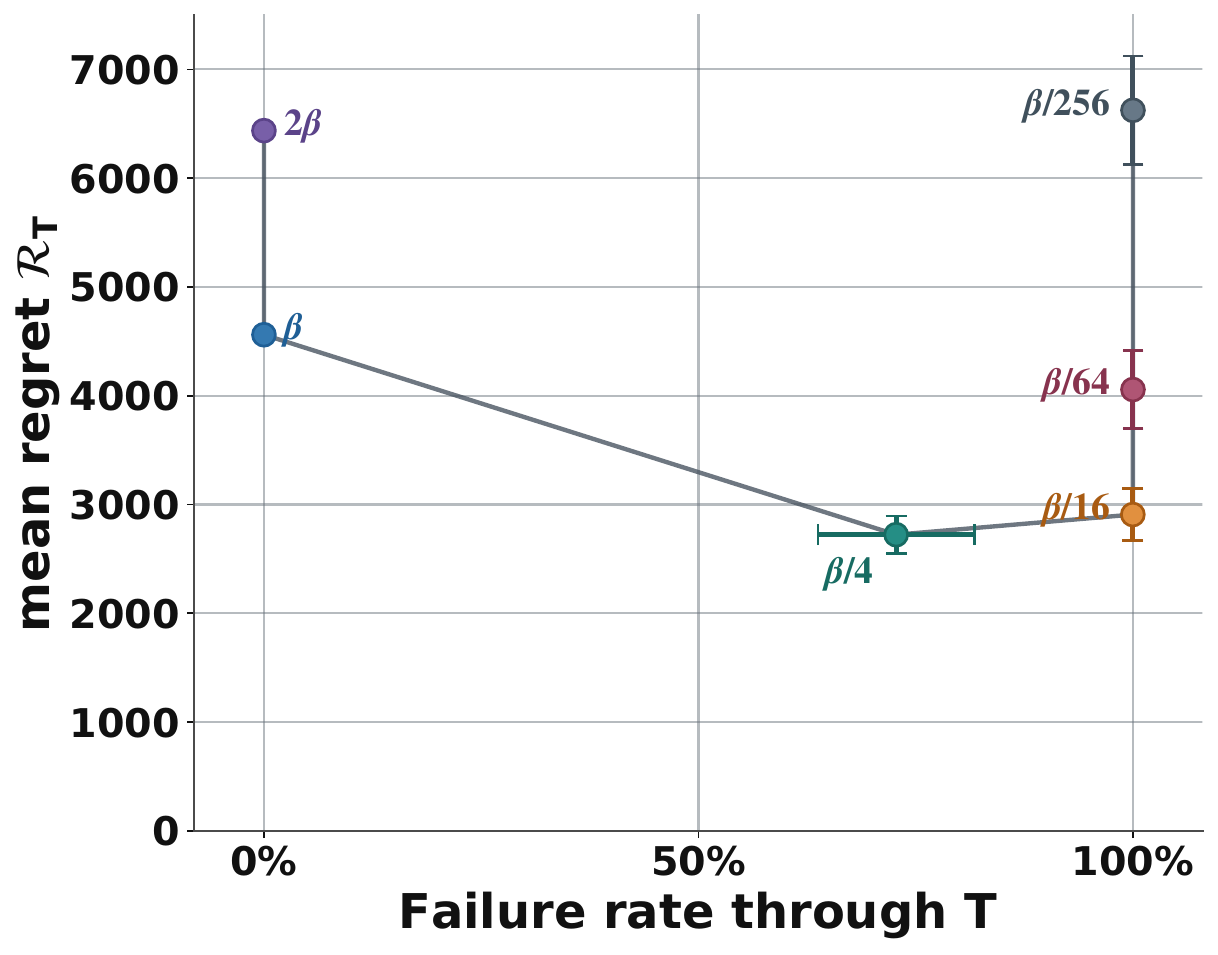}
            \caption{Failure frequency vs.\ regret}
            \label{fig:beta20-failures-regret}
        \end{subfigure}

        \caption{Effect of the confidence radius on the Euclidean unit ball
        for $d=10$, $\beta=20$, and $T=16{,}300$, with
        $\alpha\in\{\beta/256,\beta/64,\beta/16,\beta/4,\beta,2\beta\}$.
        Curves average 48 independent runs; shaded bands are pointwise
        $95\%$ confidence intervals for their means. The dashed guides
        in (a) and (b) show $d\sqrt{t}$ and $\sqrt{t/d}$ without fitted
        constants; they are not finite-time bounds. In (c), a failure means
        $\theta^\star\notin C_t(\alpha)$, and color shows its frequency in
        each time block. In (d), bars show pointwise $95\%$ confidence
        intervals for run means.}
        \label{fig:exp}
    \end{minipage}
\end{figure*}

While the contribution of this paper is mainly theoretical, we
corroborate our results with simulations. We run $\alpha$-LinUCB on the
unit ball for $d = 10$, $\theta^{\star} = \mathbf{1}_d / \sqrt{d}$,
Gaussian noise, $\beta = 20$ and $T = 16{,}300$, with $\alpha \in \{
\beta/256, \beta/64, \beta/16, \beta/4, \beta, 2\beta \}$. For each
$\alpha$ we run $48$ independent replications with the same noise
streams, and report in \cref{fig:exp} the mean of the
pseudo-regret $\cR_t$, of $\lambda_{\min}( V_{t+1} )$ and of the failure
frequency of the confidence set, $\theta^{\star} \notin \cC_t( \alpha )$, with pointwise approximate $95\%$ confidence intervals shaded.

In accordance with \cref{thm:alpha-UCB-Regret}, the regret scales as
$\sqrt{t}$ for every $\alpha$ (panel (a)), and the lowest regret is
attained at $\alpha = \beta/4$, not at the valid radius $\beta$: the
latter minimises the bound, not the regret. In accordance with
\cref{thm:spectral_growth}, $\lambda_{\min}( V_{t+1} )$ grows like
$\sqrt{t}$ for every $\alpha$, with a constant increasing in $\alpha$
(panel (b)). Panel (c) confirms that the sets are invalid for small
radii: for $\alpha \leq \beta/16$ the confidence set fails at almost
every round, whereas it never fails for $\alpha \in \{ \beta, 2\beta \}$.
Finally, panel (d) shows that the trade-off between regret and validity
is best at $\alpha = \beta/4$: its set fails in about three quarters of
the rounds, yet it achieves $\approx 40\%$ less regret than $\alpha =
\beta$.

\section*{Conclusion}

This paper studied the $\alpha$-LinUCB family of algorithms for
stochastic linear bandits, in which the exploration radius $\alpha$ is
treated as a free parameter rather than as a statistical confidence
width. Under a curvature assumption on the action set, we established a
sublinear regret bound that holds for every $\alpha > 0$, irrespective of
whether $\cC_t( \alpha )$ is a valid confidence set. The bound rests on a
fine-grained analysis of the spectral growth of the design matrix, which
shows that $\alpha$-LinUCB explores every direction at a rate that we
quantify explicitly. This analysis serves two purposes. It substitutes
for the optimism inequality, which is no longer guaranteed once
$\theta^{\star} \notin \cC_t( \alpha )$, and thereby extends the guarantee
to arbitrary radii; and it circumvents the elliptical potential lemma,
thereby improving by a factor $\sqrt{\Log{T}}$ the best known regret
bound for standard LinUCB.

A single factor $\sqrt{\Log{T}}$ nonetheless separates our bound from the
$\Omega( d \sqrt{T} )$ lower bound, and it is now inherited from the
concentration inequality for the adaptive design alone. While
\citet{LowerBoundTor} rules out the removal of this factor in the general
adaptive case, it may still be possible to dispense with it within the family of
algorithms considered here, now that the dynamics of the algorithm, and
the rate at which each direction of the design matrix grows, are better
understood. This question is both interesting and far from
straightforward, and we leave it for future work.

\newpage

\section*{AI use statement}
In this work, we used generative AI tools to edit the manuscript for readability and to check spelling, grammar, and the consistency of notation and citations. We have not used generative AI tools to develop theoretical models or conceptual frameworks, formulate mathematical claims, provide ingredients for or write proofs, propose or refine hypotheses, design or provide feedback on the methodology or experiments, implement methods, or interpret results; the remaining tasks with required disclosure (generating synthetic data, translation, dataset cleaning, and qualitative data analysis) are not applicable to this work. We have reviewed all AI-assisted work: every suggested edit was checked by the authors against the original text and the mathematical content. We take responsibility for the final content of this work, including text, claims, or artifacts produced with the aid of generative AI.

\section*{Reproducibility Statement}
This work is primarily theoretical. All assumptions required to derive our results are precisely stated and discussed in the main text (\cref{assump:subgaussian-noise,ass:armset-int}). Proof sketches are provided in the main text, and complete proofs are included in the appendices. The full protocol of the simulations (instance, parameters, number of replications) is specified in~\cref{sec:exp}.

\bibliography{Library}

\begin{thebibliography}{30}
\providecommand{\natexlab}[1]{#1}
\providecommand{\url}[1]{\texttt{#1}}
\expandafter\ifx\csname urlstyle\endcsname\relax
  \providecommand{\doi}[1]{doi: #1}\else
  \providecommand{\doi}{doi: \begingroup \urlstyle{rm}\Url}\fi

\bibitem[Abbasi-Yadkori et~al.(2011)Abbasi-Yadkori, P{\'a}l, and
  Szepesv{\'a}ri]{yassin}
Yasin Abbasi-Yadkori, D{\'a}vid P{\'a}l, and Csaba Szepesv{\'a}ri.
\newblock Improved algorithms for linear stochastic bandits.
\newblock In \emph{Advances in Neural Information Processing Systems}, 2011.

\bibitem[Abeille \& Lazaric(2017)Abeille and Lazaric]{abeille2017linear}
Marc Abeille and Alessandro Lazaric.
\newblock Linear {Thompson} sampling revisited.
\newblock In \emph{Proceedings of the 20th International Conference on
  Artificial Intelligence and Statistics}, 2017.

\bibitem[Abeille et~al.(2025)Abeille, Janz, and Pike-Burke]{abeille_when_why}
Marc Abeille, David Janz, and Ciara Pike-Burke.
\newblock When and why randomised exploration works (in linear bandits).
\newblock In \emph{Proceedings of the 36th International Conference on
  Algorithmic Learning Theory}, 2025.

\bibitem[Agrawal \& Goyal(2013)Agrawal and Goyal]{agrawal2013thompson}
Shipra Agrawal and Navin Goyal.
\newblock {Thompson} sampling for contextual bandits with linear payoffs.
\newblock In \emph{Proceedings of the 30th International Conference on Machine
  Learning}, 2013.

\bibitem[Alexandrov(1939)]{Alexandrov1939}
A.~D. Alexandrov.
\newblock Almost everywhere existence of the second differential of a convex
  function and some properties of convex surfaces connected with it.
\newblock \emph{Leningrad State University Annals [Uchenye Zapiski],
  Mathematical Series}, 6:\penalty0 3--35, 1939.
\newblock In Russian.

\bibitem[Banerjee et~al.(2023)Banerjee, Ghosh, Ray~Chowdhury, and
  Gopalan]{pmlr-v206-banerjee23b}
Debangshu Banerjee, Avishek Ghosh, Sayak Ray~Chowdhury, and Aditya Gopalan.
\newblock Exploration in linear bandits with rich action sets and its
  implications for inference.
\newblock In \emph{Proceedings of the 26th International Conference on
  Artificial Intelligence and Statistics}, 2023.

\bibitem[Bastani et~al.(2021)Bastani, Bayati, and Khosravi]{bastani2021mostly}
Hamsa Bastani, Mohsen Bayati, and Khashayar Khosravi.
\newblock Mostly exploration-free algorithms for contextual bandits.
\newblock \emph{Management Science}, 67\penalty0 (3):\penalty0 1329--1349,
  2021.

\bibitem[Beck(2017)]{Beck2017}
Amir Beck.
\newblock \emph{First-Order Methods in Optimization}.
\newblock Society for Industrial and Applied Mathematics, 2017.

\bibitem[Dani et~al.(2008)Dani, Hayes, and Kakade]{dani2008stochastic}
Varsha Dani, Thomas~P. Hayes, and Sham~M. Kakade.
\newblock Stochastic linear optimization under bandit feedback.
\newblock In \emph{Proceedings of the 21st Annual Conference on Learning
  Theory}, 2008.

\bibitem[Deshpande et~al.(2018)Deshpande, Mackey, Syrgkanis, and
  Taddy]{deshpande2018accurate}
Yash Deshpande, Lester Mackey, Vasilis Syrgkanis, and Matt Taddy.
\newblock Accurate inference for adaptive linear models.
\newblock In \emph{Proceedings of the 35th International Conference on Machine
  Learning}, 2018.

\bibitem[Evans \& Gariepy(2015)Evans and Gariepy]{EvansGariepy2015}
Lawrence~C. Evans and Ronald~F. Gariepy.
\newblock \emph{Measure Theory and Fine Properties of Functions}.
\newblock CRC Press, Boca Raton, FL, revised edition, 2015.

\bibitem[Flynn et~al.(2023)Flynn, Reeb, Kandemir, and
  Peters]{flynn2023improved}
Hamish Flynn, David Reeb, Melih Kandemir, and Jan~R. Peters.
\newblock Improved algorithms for stochastic linear bandits using tail bounds
  for martingale mixtures.
\newblock In \emph{Advances in Neural Information Processing Systems}, 2023.

\bibitem[Hao et~al.(2020)Hao, Lattimore, and Szepesv{\'a}ri]{hao2020adaptive}
Botao Hao, Tor Lattimore, and Csaba Szepesv{\'a}ri.
\newblock Adaptive exploration in linear contextual bandit.
\newblock In \emph{Proceedings of the 23rd International Conference on
  Artificial Intelligence and Statistics}, 2020.

\bibitem[Huang et~al.(2016)Huang, Lattimore, Gy{\"o}rgy, and
  Szepesv{\'a}ri]{NIPS2016_55a988df}
Ruitong Huang, Tor Lattimore, Andr{\'a}s Gy{\"o}rgy, and Csaba Szepesv{\'a}ri.
\newblock Following the leader and fast rates in linear prediction: Curved
  constraint sets and other regularities.
\newblock In \emph{Advances in Neural Information Processing Systems}, 2016.

\bibitem[Kannan et~al.(2018)Kannan, Morgenstern, Roth, Waggoner, and
  Wu]{kannan2018smoothed}
Sampath Kannan, Jamie Morgenstern, Aaron Roth, Bo~Waggoner, and Zhiwei~Steven
  Wu.
\newblock A smoothed analysis of the greedy algorithm for the linear contextual
  bandit problem.
\newblock In \emph{Advances in Neural Information Processing Systems}, 2018.

\bibitem[Khamaru et~al.(2025)Khamaru, Deshpande, Lattimore, Mackey, and
  Wainwright]{khamarou_tor_Near_optimal}
Koulik Khamaru, Yash Deshpande, Tor Lattimore, Lester Mackey, and Martin~J.
  Wainwright.
\newblock Near-optimal inference in adaptive linear regression.
\newblock \emph{The Annals of Statistics}, 53\penalty0 (6):\penalty0
  2329--2355, 2025.

\bibitem[Lai \& Wei(1982)Lai and Wei]{lai1982least}
Tze~Leung Lai and Ching~Zong Wei.
\newblock Least squares estimates in stochastic regression models with
  applications to identification and control of dynamic systems.
\newblock \emph{The Annals of Statistics}, 10\penalty0 (1):\penalty0 154--166,
  1982.

\bibitem[Lattimore(2023)]{LowerBoundTor}
Tor Lattimore.
\newblock A lower bound for linear and kernel regression with adaptive
  covariates.
\newblock In \emph{Proceedings of the 36th Conference on Learning Theory},
  2023.

\bibitem[Lattimore \& Szepesv{\'a}ri(2017)Lattimore and
  Szepesv{\'a}ri]{pmlr-v54-lattimore17a}
Tor Lattimore and Csaba Szepesv{\'a}ri.
\newblock The end of optimism? an asymptotic analysis of finite-armed linear
  bandits.
\newblock In \emph{Proceedings of the 20th International Conference on
  Artificial Intelligence and Statistics}, 2017.

\bibitem[Lattimore \& Szepesv{\'a}ri(2020)Lattimore and
  Szepesv{\'a}ri]{lattimore2020bandit}
Tor Lattimore and Csaba Szepesv{\'a}ri.
\newblock \emph{Bandit Algorithms}.
\newblock Cambridge University Press, 2020.

\bibitem[Li et~al.(2010)Li, Chu, Langford, and Schapire]{Li-LINUCB}
Lihong Li, Wei Chu, John Langford, and Robert~E. Schapire.
\newblock A contextual-bandit approach to personalized news article
  recommendation.
\newblock In \emph{Proceedings of the 19th International Conference on World
  Wide Web}, 2010.

\bibitem[Lin et~al.(2023)Lin, Ying, Ghosh, Khamaru, and Zhang]{kAdaptive}
Licong Lin, Mufang Ying, Suvrojit Ghosh, Koulik Khamaru, and Cun-Hui Zhang.
\newblock Statistical limits of adaptive linear models: Low-dimensional
  estimation and inference.
\newblock In \emph{Advances in Neural Information Processing Systems}, 2023.

\bibitem[Praharaj \& Khamaru(2026)Praharaj and
  Khamaru]{praharaj2026avoidingpriceadaptivityinference}
Samya Praharaj and Koulik Khamaru.
\newblock Avoiding the price of adaptivity: Inference in linear contextual
  bandits via stability.
\newblock arXiv:2512.20368v2, 2026.

\bibitem[Raghavan et~al.(2023)Raghavan, Slivkins, Vaughan, and
  Wu]{raghavan2023greedy}
Manish Raghavan, Aleksandrs Slivkins, Jennifer~Wortman Vaughan, and
  Zhiwei~Steven Wu.
\newblock Greedy algorithm almost dominates in smoothed contextual bandits.
\newblock \emph{SIAM Journal on Computing}, 52\penalty0 (2):\penalty0 487--524,
  2023.

\bibitem[Rockafellar \& Wets(1998)Rockafellar and Wets]{RockafellarWets1998}
R.~Tyrrell Rockafellar and Roger J.-B. Wets.
\newblock \emph{Variational Analysis}.
\newblock Springer, 1998.

\bibitem[Rusmevichientong \& Tsitsiklis(2010)Rusmevichientong and
  Tsitsiklis]{rusmevichientong2010linearly}
Paat Rusmevichientong and John~N. Tsitsiklis.
\newblock Linearly parameterized bandits.
\newblock \emph{Mathematics of Operations Research}, 35\penalty0 (2):\penalty0
  395--411, 2010.

\bibitem[Schneider(2014)]{schneider2014}
Rolf Schneider.
\newblock \emph{Convex Bodies: The {Brunn--Minkowski} Theory}.
\newblock Cambridge University Press, 2 edition, 2014.

\bibitem[Tirinzoni et~al.(2020)Tirinzoni, Pirotta, Restelli, and
  Lazaric]{tirinzoni2020asymptoticallyoptimalprimaldualincremental}
Andrea Tirinzoni, Matteo Pirotta, Marcello Restelli, and Alessandro Lazaric.
\newblock An asymptotically optimal primal-dual incremental algorithm for
  contextual linear bandits.
\newblock In \emph{Advances in Neural Information Processing Systems}, 2020.

\bibitem[Ville(1939)]{Ville1939}
Jean Ville.
\newblock \emph{{\'E}tude critique de la notion de collectif}.
\newblock PhD thesis, Universit{\'e} de Paris, 1939.

\bibitem[Ying et~al.(2023)Ying, Khamaru, and Zhang]{ying2023adaptive}
Mufang Ying, Koulik Khamaru, and Cun-Hui Zhang.
\newblock Adaptive linear estimating equations.
\newblock In \emph{Advances in Neural Information Processing Systems}, 2023.

\end{thebibliography}
\bibliographystyle{iclr2027_conference}

\newpage
\appendix

\thmboxesfalse   %

\etocdepthtag.toc{appendix}

\section*{Appendix Contents}
\etocsettagdepth{main}{none}
\etocsettagdepth{appendix}{subsection}
\etocsettocstyle{}{}
\tableofcontents

\newpage

\section{Self-normalised bound for adaptive designs}
\phantomsection\label{sec:Useful_facts}

\propConf* 

\Cref{prop:Yassin_conf} is Theorem~2 of \citet{yassin}, applied with
noise parameter $\mu$, $\norm{x_s}_2 \leq 1$, regularisation $\lambda_0$
and $\norm{\theta^{\star}}_2 \leq S^{\star}$; the radius there is
$\mu \sqrt{2 \Log{( 1 / \delta )} + \Log{( \Det{V_t} / \Det{\lambda_0
\I} )}} + \sqrt{\lambda_0} S^{\star}$, and $\Det{V_t} \leq ( \lambda_0 +
t / d )^d$ (since $\Tr V_t \leq d \lambda_0 + t$) gives the
horizon-dependent radius above. We comment on the role of this result
in our analysis.

\paragraph{Adaptive design.}
The covariates $x_1, x_2, \dots$ are not fixed in advance: $x_t$ is
chosen by the algorithm as a function of the past rewards, so that
$( x_s )_{s \leq t}$ and $( \eta_s )_{s \leq t}$ are dependent and the
classical confidence sets of fixed-design least squares, of radius
$\asymp \mu \sqrt{d}$ with high probability, are not available. The
bound of \citet{yassin} handles the sequential design through a
self-normalised martingale inequality (the method of mixtures): the
quantity $\norm{\sum_{s < t} x_s \eta_s}_{V_t^{-1}}$ is controlled
simultaneously for all $t$, whatever the dependence of $x_s$ on the
past, at the price of the $\log \det V_t \leq d \Log{( 1 + t / ( d
\lambda_0 ) )}$ term, i.e.\ of the factor $\sqrt{\Log{T}}$ in $\beta
\asymp \sqrt{d \Log{T}}$. This is the \emph{price of adaptivity}: it
cannot be removed in general, since \citet{LowerBoundTor} shows that
the bound is unimprovable for adversarially adaptive designs, and the
same phenomenon governs estimation and inference from adaptively
collected data \citep{khamarou_tor_Near_optimal,kAdaptive}. Sharper
radii are known only under additional structure on the design, such as
the stability condition of \citet{lai1982least}, i.e.\ convergence of
the normalised covariance to a deterministic limit; we do not use such
refinements, and the radius $\beta$ of \cref{prop:Yassin_conf} is the
one that enters \cref{thm:spectral_growth,thm:alpha-UCB-Regret}.

\paragraph{How the event is used.}
The event $\cE$ is the only probabilistic input of the paper: every
statement of \cref{sec:spectral_growth_proof,sec:regret-proof} is
deterministic on $\cE$, and the algorithm itself never uses $\beta$.
Its role is purely analytical, to locate $\theta^{\star}$ within
$\norm{\cdot}_{V_t}$-distance $\beta$ of the estimate, which the width
$\omega_t$ converts into the $\norm{\cdot}_{\ast}$-distance $\beta\,
\omega_t^{1/2}$ (\cref{prop:width-int}); the set $\cC_t( \alpha )$ used
by $\alpha$-LinUCB may be smaller or larger than $\cC_t( \beta )$. We
take $\beta$ horizon-dependent for simplicity; the anytime radius
$\beta_t$ with $\Log{( \Det{V_t} / \Det{\lambda_0 \I} )}$ in place of
$d \Log{( 1 + T / ( d \lambda_0 ) )}$ can be used instead, at the cost
of carrying a nondecreasing $\beta_t$ through the burn-in constants.

\section{Further properties of the action set}
\label{sec:armset-more}

\paragraph{Consequences of \cref{ass:armset-int}.}
Setting $\theta' = 0$ in~\eqref{eq:ajp-int} yields
$\frac{m}{2} \norm{v}_{\ast}^2 \leq J_{\cX}(v)^2 \leq
\frac{M}{2} \norm{v}_{\ast}^2$. Consequently, for
$\theta^{\star} \neq 0$, the scalars
\begin{equation}
\label{eq:scalars-upperbound}
\begin{gathered}
    \ell_{\ast} := \sup_{x \in \cX} \norm{x}_{\ast}^{\circ},
    \;\;
    J_{\star} := J_{\cX}(\theta^{\star}),
    \;\;
    \varrho_{\ast} := J_{\star} / \norm{\theta^{\star}}_{\ast} \\
    \text{satisfy} \;\;
    J_{\star} > 0,
    \;\;
    \sqrt{m/2} \leq \varrho_{\ast}
    \leq \norm{x^{\star}}_{\ast}^{\circ}
    \leq \ell_{\ast} \leq \sqrt{M/2},
\end{gathered}
\end{equation}
where $x^{\star} := \nabla J_{\cX}(\theta^{\star})$. In particular, $\cX$
is compact and $J_{\cX}$ is $\ell_{\ast}$-Lipschitz with respect to
$\norm{\cdot}_{\ast}$. These quantities enter the analysis only; the
algorithm does not require them.

\paragraph{Unit ball and ellipsoids.}
For $\cX = \mathbb{B}_2^d(1)$, one has $\ell_{\ast} = \varrho_{\ast} = 1$
and $J_{\star} = \norm{\theta^{\star}}_2$. For
$\cX = \bA^{1/2} \mathbb{B}_2^d(1)$, under the adapted norm
$\norm{u}_{\ast} := \norm{\bA^{1/2} u}_2$, one likewise has
$\ell_{\ast} = \varrho_{\ast} = 1$.

\paragraph{$C^2$ bodies.}
A convex body $\cX \subseteq \mathbb{B}_2^d(1)$ containing the origin in
its interior is a \emph{$C^2$ body} if its boundary is a $C^2$
hypersurface whose principal radii of curvature lie in $[ r_-, r_+ ]$
for some $0 < r_- \leq r_+ < \infty$ (class $C^2_+$
in~\citet[Sect.~2.5]{schneider2014}). Its support function is $C^2$ on
$\lR^d \setminus \{ 0 \}$. For a unit vector $u$, the Hessian
$\nabla^2 J_{\cX}( u )$ vanishes on $u$ (Euler's relation) and coincides
on $u^{\perp}$ with the reverse Weingarten map of $\partial \cX$ at the
point $x = \nabla J_{\cX}( u )$ with outer normal $u$; its eigenvalues
are the principal radii of curvature of $\partial \cX$ at $x$. Since
$\nabla^2 J_{\cX}^2 = 2 J_{\cX} \nabla^2 J_{\cX}
+ 2 \nabla J_{\cX} \nabla J_{\cX}^{\top}$ with both terms homogeneous of
degree zero, writing $u = \theta / \norm{\theta}_2$ and $v = a u + w$
with $w \perp u$ gives
\begin{equation}
\label{eq:hessJ2-int}
    v^{\top} \nabla^2 J_{\cX}^2( \theta )\, v
    = 2\, J_{\cX}( u )\, w^{\top} \nabla^2 J_{\cX}( u )\, w
    + 2\, \dotprod{x}{v}^2 ,
    \qquad
    r_- \norm{w}_2^2 \leq w^{\top} \nabla^2 J_{\cX}( u )\, w \leq r_+
    \norm{w}_2^2 .
\end{equation}
The first (tangential) term captures the curvature of $\partial \cX$;
the second (radial) term captures the position of the origin, since
$\dotprod{x}{u} = J_{\cX}( u )$ is the distance from the origin to the
supporting hyperplane with normal $u$. Let
\[
    \rho_{\min} := \min_{\norm{u}_2 = 1} J_{\cX}( u ) \in ( 0, 1 ]
\]
denote the distance from the origin to $\partial \cX$, positive since
$\cX$ is absorbing and at most one since
$\cX \subseteq \mathbb{B}_2^d(1)$. With
$\norm{\cdot}_{\ast} = \norm{\cdot}_2$, which is self-dual,
$\ell_{\ast} \leq 1$ and $\varrho_{\ast} \geq \rho_{\min}$.

\emph{Upper bound.} Using $J_{\cX}( u ) \leq 1$ and
$\dotprod{x}{v}^2 \leq \norm{v}_2^2$,
\[
    v^{\top} \nabla^2 J_{\cX}^2( \theta )\, v
    \;\leq\; 2 r_+ \norm{w}_2^2 + 2 \norm{v}_2^2
    \;\leq\; 2 ( r_+ + 1 ) \norm{v}_2^2 =: M \norm{v}_2^2 .
\]

\emph{Lower bound.} Using $J_{\cX}( u ) \geq \rho_{\min}$ and
$| \dotprod{x}{v} | \geq \rho_{\min} | a | - \norm{w}_2$,
\[
    v^{\top} \nabla^2 J_{\cX}^2( \theta )\, v
    \;\geq\; 2 \rho_{\min} r_- \norm{w}_2^2
    + 2 \big( \rho_{\min} | a | - \norm{w}_2 \big)_{+}^2 .
\]
If $\norm{w}_2 \geq \rho_{\min} | a | / 2$, then
$\norm{v}_2^2 \leq 5 \norm{w}_2^2 / \rho_{\min}^2$ and the first term is
at least $\frac{2}{5} \rho_{\min}^3 r_- \norm{v}_2^2$. Otherwise,
$\norm{v}_2^2 \leq \frac{5}{4} a^2$ and the second term is at least
$\frac{2}{5} \rho_{\min}^2 \norm{v}_2^2$. Hence
\[
    v^{\top} \nabla^2 J_{\cX}^2( \theta )\, v
    \;\geq\; \frac{2 \rho_{\min}^2}{5} \min\{ 1, \rho_{\min} r_- \}
    \norm{v}_2^2
    \;\geq\; \frac{\rho_{\min}^2}{9} \min\Big\{ 1, \frac{\rho_{\min}
    r_-}{3} \Big\} \norm{v}_2^2 =: m \norm{v}_2^2 ,
\]. The two-sided bound
of \cref{ass:armset-int} follows by integrating twice along the segment
from $\theta'$ to $\theta$; the segment may pass through the origin,
since $J_{\cX}^2$ is $C^1$ on $\lR^d$ (its gradient
$2 J_{\cX} \nabla J_{\cX}$ is bounded by $2 \norm{\theta}_2$ and extends
continuously by $0$ at the origin).

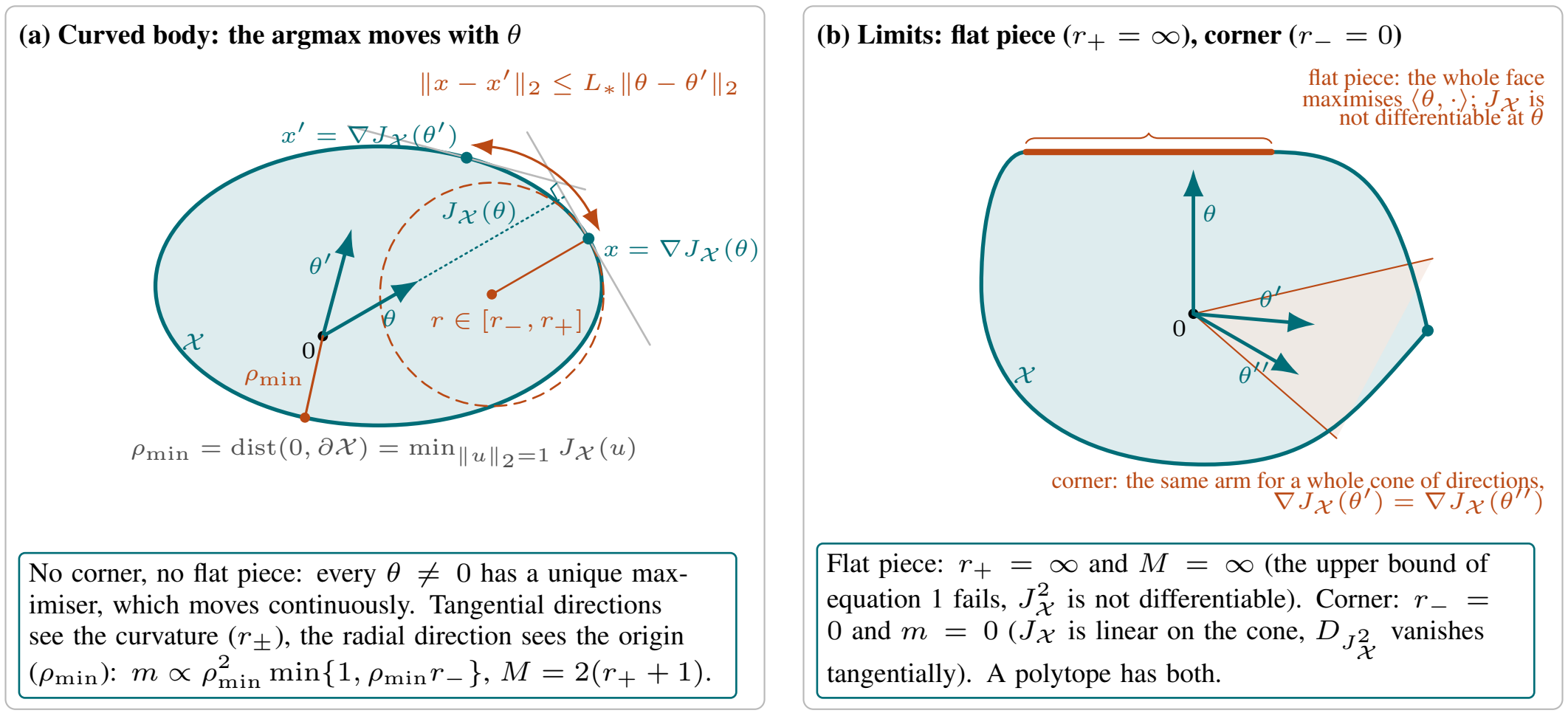
\begin{figure}[h]
\centering
\begin{tikzpicture}[x=1cm,y=1cm,font=\scriptsize,>=Latex,line cap=round,line join=round]
\definecolor{acc}{RGB}{0,109,119}
\definecolor{acc2}{RGB}{186,73,20}
\definecolor{soft}{RGB}{222,236,238}
\definecolor{softb}{RGB}{246,232,224}
\definecolor{grid}{RGB}{185,185,185}
\tikzset{
  panel/.style={draw=grid,rounded corners=2.5pt,fill=white,line width=0.4pt,minimum width=6.8cm,minimum height=6.3cm,anchor=south west},
  ptitle/.style={anchor=north west,font=\scriptsize\bfseries,inner sep=0pt},
  tnote/.style={font=\fontsize{6}{7}\selectfont,inner sep=0.7pt,align=left},
  key/.style={font=\scriptsize,inner sep=2.5pt,fill=white,rounded corners=1.5pt,draw=acc,line width=0.5pt,align=left,text width=6.25cm},
  dot/.style={circle,fill=black,inner sep=0pt,minimum size=2.6pt},
  adot/.style={circle,fill=acc,inner sep=0pt,minimum size=3pt},
  rdot/.style={circle,fill=acc2,inner sep=0pt,minimum size=2.6pt},
  dir/.style={->,acc,line width=0.9pt},
}
\node[panel] (PA) at (0,0) {};
\node[panel] (PB) at (7.15,0) {};

\begin{scope}[shift={(0,0)}]
  \node[ptitle] at (0.12,6.15) {(a) Curved body: the argmax moves with $\theta$};
  \fill[soft] (3.35,3.8) ellipse [x radius=2.0,y radius=1.25];
  \draw[acc,line width=1pt] (3.35,3.8) ellipse [x radius=2.0,y radius=1.25];
  \node[tnote,text=acc,anchor=west] at (1.55,3.3) {$\cX$};
  \draw[acc2,densely dashed,line width=0.5pt] (4.364,3.724) circle [radius=1.001];
  \draw[acc2,line width=0.5pt] (4.364,3.724) -- (5.231,4.224);
  \node[rdot] at (4.364,3.724) {};
  \node[tnote,text=acc2,anchor=north] at (4.5,3.62) {$r\in[r_-,r_+]$};
  \draw[grid,line width=0.5pt] (4.681,5.177) -- (5.781,3.272);
  \draw[grid,line width=0.5pt] (3.076,5.234) -- (5.201,4.664);
  \node[dot] at (2.85,3.35) {};
  \node[tnote,anchor=north east] at (2.82,3.32) {$0$};
  \draw[dir] (2.85,3.35) -- (3.716,3.85);
  \node[tnote,text=acc,anchor=north west] at (3.35,3.62) {$\theta$};
  \draw[dir] (2.85,3.35) -- (3.109,4.316);
  \node[tnote,text=acc,anchor=east] at (2.98,4.0) {$\theta'$};
  \draw[acc,densely dotted,line width=0.5pt] (3.716,3.85) -- (5.015,4.6);
  \draw[acc,line width=0.4pt] (4.945,4.72) -- (4.892,4.671) -- (4.945,4.58);
  \node[tnote,text=acc,anchor=south] at (4.25,4.32) {$J_{\cX}(\theta)$};
  \node[adot] at (5.231,4.224) {};
  \node[tnote,text=acc,anchor=west] at (5.33,4.12) {$x=\nabla J_{\cX}(\theta)$};
  \node[adot] at (4.138,4.949) {};
  \node[tnote,text=acc,anchor=south east] at (4.1,5.0) {$x'=\nabla J_{\cX}(\theta')$};
  \draw[<->,acc2,line width=0.6pt] (5.335,4.284) to[bend right=28] (4.169,5.065);
  \node[tnote,text=acc2,anchor=east] at (6.62,5.62) {$\norm{x-x'}_2\le L_{\ast}\norm{\theta-\theta'}_2$};
  \draw[acc2,line width=0.6pt] (2.85,3.35) -- (2.69,2.62);
  \node[rdot] at (2.69,2.62) {};
  \node[tnote,text=acc2,anchor=east] at (2.72,2.98) {$\rho_{\min}$};
  \node[tnote,text=gray!60!black,anchor=north] at (3.4,2.48) {$\rho_{\min}=\mathrm{dist}(0,\partial\cX)=\min_{\norm{u}_2=1}J_{\cX}(u)$};
  \node[key,anchor=south west] at (0.12,0.1)
    {No corner, no flat piece: every $\theta\ne0$ has a unique maximiser, which moves continuously. Tangential directions see the curvature ($r_\pm$), the radial direction sees the origin ($\rho_{\min}$): $m\propto\rho_{\min}^2\min\{1,\rho_{\min}r_-\}$, $M=2(r_++1)$.};
\end{scope}

\begin{scope}[shift={(7.15,0)}]
  \node[ptitle] at (0.12,6.15) {(b) Limits: flat piece ($r_+=\infty$), corner ($r_-=0$)};
  \fill[soft] (2.0,5.000) -- (4.2,5.000) .. controls (5.0,5.000) and (5.3,4.700) .. (5.6,3.400)
     .. controls (4.9,2.600) and (4.6,2.200) .. (3.6,2.200) .. controls (2.4,2.200) and (1.6,2.800) .. (1.6,3.800)
     .. controls (1.6,4.600) and (1.8,5.000) .. (2.0,5.000) -- cycle;
  \fill[softb,opacity=0.7] (3.5,3.550) -- (4.783,2.435) -- (5.644,4.045) -- cycle;
  \draw[acc2,line width=0.4pt] (3.5,3.550) -- (4.783,2.435);
  \draw[acc2,line width=0.4pt] (3.5,3.550) -- (5.644,4.045);
  \draw[acc,line width=1pt] (2.0,5.000) -- (4.2,5.000) .. controls (5.0,5.000) and (5.3,4.700) .. (5.6,3.400)
     .. controls (4.9,2.600) and (4.6,2.200) .. (3.6,2.200) .. controls (2.4,2.200) and (1.6,2.800) .. (1.6,3.800)
     .. controls (1.6,4.600) and (1.8,5.000) .. (2.0,5.000) -- cycle;
  \node[tnote,text=acc,anchor=west] at (1.85,3.000) {$\cX$};
  \draw[acc2,line width=1.6pt] (2.0,5.000) -- (4.2,5.000);
  \draw[decorate,decoration={brace,amplitude=2.5pt},acc2,line width=0.4pt] (2.0,5.080) -- (4.2,5.080);
  \node[tnote,text=acc2,anchor=north east] at (6.68,5.78) {flat piece: the whole face};
  \node[tnote,text=acc2,anchor=north east] at (6.68,5.60) {maximises $\dotprod{\theta}{\cdot}$; $J_{\cX}$ is};
  \node[tnote,text=acc2,anchor=north east] at (6.68,5.42) {not differentiable at $\theta$};

  \node[adot] at (5.6,3.400) {};
  \node[tnote,text=acc2,anchor=east] at (6.68,2.05) {corner: the same arm for a whole cone of directions,};
  \node[tnote,text=acc2,anchor=east] at (6.68,1.87) {$\nabla J_{\cX}(\theta')=\nabla J_{\cX}(\theta'')$};
  \node[dot] at (3.5,3.550) {};
  \node[tnote,anchor=north east] at (3.47,3.520) {$0$};
  \draw[dir] (3.5,3.550) -- (3.5,4.850);
  \node[tnote,text=acc,anchor=west] at (3.55,4.450) {$\theta$};
  \draw[dir] (3.5,3.550) -- (4.596,3.454);
  \node[tnote,text=acc,anchor=south] at (4.2,3.580) {$\theta'$};
  \draw[dir] (3.5,3.550) -- (4.453,3.000);
  \node[tnote,text=acc,anchor=north] at (4.05,3.150) {$\theta''$};
  \node[key,anchor=south west] at (0.12,0.1)
    {Flat piece: $r_+=\infty$ and $M=\infty$ (the upper bound of \eqref{eq:ajp-int} fails, $J_{\cX}^2$ is not differentiable). Corner: $r_-=0$ and $m=0$ ($J_{\cX}$ is linear on the cone, $D_{J_{\cX}^2}$ vanishes tangentially). A polytope has both.};
\end{scope}
\end{tikzpicture}
\caption{\Cref{ass:armset-int} illustrated (parameter and action spaces
superimposed). (a)~A $C^2$ body: each direction $\theta$ selects a unique
arm $x = \nabla J_{\cX}( \theta )$, the point of $\partial \cX$ with outer
normal $\theta$, which moves Lipschitz-continuously with $\theta$. The
radius of curvature $r \in [ r_-, r_+ ]$ at $x$ governs the tangential
term of \eqref{eq:hessJ2-int}, and $\rho_{\min}$ its radial term.
(b)~The two degenerate limits: a flat face ($r_+ = \infty$, $M = \infty$)
and a corner ($r_- = 0$, $m = 0$).}
\label{fig:armset-int}
\end{figure}

\paragraph{Role of curvature and excluded geometries.}
In tangential directions, the Hessian of $J_{\cX}^2$ equals
$2 J_{\cX}( u )$ times the reverse Weingarten map: flat regions
($r_+$ large) inflate $M$, and sharp regions ($r_-$ small) deflate $m$.
In the radial direction, it equals $2 J_{\cX}( u )^2 \geq 2 \rho_{\min}^2$,
which deflates $m$ as the origin approaches the boundary, irrespective of
curvature (\cref{fig:armset-int}). The two limits are the excluded cases.
A flat face is a direction $\theta$ with a whole face of maximisers:
$J_{\cX}$ is not differentiable at $\theta$, the arm jumps across the
face under arbitrarily small perturbations of $\theta$, and $M = \infty$.
A corner $x_k$ is the common maximiser of a whole cone of directions: on
this cone, $J_{\cX}( \theta ) = \dotprod{\theta}{x_k}$ is linear,
$\nabla^2 J_{\cX}^2 = 2 x_k x_k^{\top}$ has rank one,
$D_{J_{\cX}^2}( \theta, \theta' ) = \dotprod{\theta - \theta'}{x_k}^2$
vanishes for $\theta - \theta' \perp x_k$, and $m = 0$. Polytopes and
finite arm sets exhibit both. On a $C^2$ body, by contrast, every
$\theta \neq 0$ has a unique maximiser, which moves
Lipschitz-continuously with $\theta$ (\cref{prop:lip-int}); this allows a
perturbation of the optimistic parameter to reach the arm, and is what
the energy argument of the proof exploits. The constants are not sharp:
for the unit ball, $r_{\pm} = \rho_{\min} = 1$ and \eqref{eq:hessJ2-int}
equals $2 \norm{v}_2^2$, so that $m = M = 2$, whereas the bounds above
give $m = 2/5$ and $M = 4$.

\subsection{Failure case}
\phantomsection\label{sec:counter-exmaple}

\PropCounterEgs*

\begin{proof}[\cref{prop:counter-egs}]
\emph{Step 1: the algorithm plays vertices.} For $x \in \cX_a$,
$\max_{\theta \in \cC_t( \alpha )} \dotprod{x}{\theta} = \dotprod{x}
{\hat{\theta}_t} + \alpha \norm{x}_{V_t^{-1}} =: g_t( x )$, so $x_t \in
\argmax_{\cX_a} g_t$. The function $g_t$ is convex, and strictly convex
along every segment not contained in a line through the origin. Its maximum over
the polytope is therefore attained at vertices only: a convex function
with a maximum at the midpoint of a segment is constant on that segment,
and $g_t$ is constant on no segment, so no maximiser lies in
the interior of $\cX_a$ or in the relative interior of an edge. Since the four vertices lie on
the coordinate axes, $V_t = \mathrm{diag}( \lambda_0 + n_{t,1},
\lambda_0 + a^2 n_{t,2} )$, where $n_{t,i}$ is the number of rounds $s <
t$ with $x_s$ on the $i$-th axis, $\hat{\theta}_{t,1} = ( \lambda_0 +
n_{t,1} )^{-1} \sum_{s < t,\, x_s = \pm e_1} \dotprod{x_s}{e_1}\, y_s$,
and the four indices are
\[
    g_t( \pm e_1 ) = \pm \hat{\theta}_{t,1} + \frac{\alpha}{\sqrt{\lambda_0
    + n_{t,1}}} ,
    \qquad
    g_t( \pm a e_2 ) = \pm a\, \hat{\theta}_{t,2} + \frac{a \alpha}
    {\sqrt{\lambda_0 + a^2 n_{t,2}}} .
\]

\emph{Step 2: a sufficient condition for locking on the first axis.}
Let $\tilde{\eta}_s := \dotprod{x_s}{e_1}\, \eta_s$ if $x_s \in \{ \pm
e_1 \}$ and $\tilde{\eta}_s := \eta_s$ otherwise, and $W_n := \sum_{s
\leq n} \tilde{\eta}_s$. As $\dotprod{x_s}{e_1} \in \{ \pm 1 \}$ is
$\cF_{s-1}$-measurable and $\eta_s \sim \mathcal{N}( 0, \mu^2 )$ is
independent of $\cF_{s-1}$, the conditional law of $\tilde{\eta}_s$
given $\cF_{s-1}$ is $\mathcal{N}( 0, \mu^2 )$: $( \tilde{\eta}_s )_s$ is
i.i.d.\ $\mathcal{N}( 0, \mu^2 )$. Suppose that $x_s \in \{ \pm e_1 \}$
for all $s < t$ and put $n := t - 1$. Then $n_{t,1} = n$, $n_{t,2} = 0$,
$\hat{\theta}_{t,2} = 0$ and, since $\dotprod{x_s}{e_1}\, y_s =
\varepsilon + \tilde{\eta}_s$,
\[
    \hat{\theta}_{t,1} = \frac{n \varepsilon + W_n}{\lambda_0 + n} ,
\]
so that $x_t \in \{ \pm e_1 \}$ as soon as
\begin{equation}
\label{eq:lock-int}
    | \hat{\theta}_{t,1} | + \frac{\alpha}{\sqrt{\lambda_0 + n}}
    \;>\; \frac{a \alpha}{\sqrt{\lambda_0}} ,
\end{equation}
the left-hand side being the larger of $g_t( \pm e_1 )$ and the
right-hand side the common value of $g_t( \pm a e_2 )$. For $n < n_0 :=
\lambda_0 ( a^{-2} - 1 )$, \eqref{eq:lock-int} holds whatever the
rewards, because $\alpha / \sqrt{\lambda_0 + n} > a \alpha /
\sqrt{\lambda_0}$: the bonus of the long vertices alone exceeds the
frozen bonus of the short ones. For $n \geq N_0 := \lceil n_0 \rceil$,
\eqref{eq:lock-int} holds as soon as $\hat{\theta}_{t,1} > a \alpha /
\sqrt{\lambda_0}$, i.e.
\[
    W_n + \delta n > c_{\alpha} ,
    \qquad
    \delta := \varepsilon - \frac{a \alpha}{\sqrt{\lambda_0}} ,
    \qquad
    c_{\alpha} := a \alpha \sqrt{\lambda_0} .
\]
Since $x_1 \in \{ \pm e_1 \}$ (at $t = 1$, $g_1( \pm e_1 ) = \alpha /
\sqrt{\lambda_0} > a \alpha / \sqrt{\lambda_0} = g_1( \pm a e_2 )$), an
induction on $t$ gives
\[
    \big\{ W_n + \delta n > c_{\alpha} \ \text{ for all } n \geq N_0
    \big\}
    \;\subseteq\;
    \big\{ x_t \in \{ \pm e_1 \} \ \text{ for all } t \geq 1 \big\} .
\]

\emph{Step 3: the probability of the left-hand event.} For $\alpha \leq
\alpha_0$ we have $a \alpha / \sqrt{\lambda_0} \leq ( 1 - a^2 )
\varepsilon / 2$, hence $\delta \geq \varepsilon / 2$ and, using $( 1 -
a^2 ) ( \lambda_0 + N_0 ) \leq N_0$ (which is $N_0 \geq n_0$),
\[
    h := \delta N_0 - c_{\alpha}
    = \varepsilon N_0 - \frac{a \alpha}{\sqrt{\lambda_0}} ( \lambda_0 +
    N_0 )
    \;\geq\; \varepsilon N_0 - \frac{( 1 - a^2 ) \varepsilon}{2} (
    \lambda_0 + N_0 )
    \;\geq\; \frac{\varepsilon N_0}{2} .
\]
Now $\{ W_n + \delta n > c_{\alpha}\ \forall n \geq N_0 \} \supseteq \{
W_{N_0} \geq 0 \} \cap \{ W_{N_0 + k} - W_{N_0} + \delta k > - h\
\forall k \geq 1 \}$: on the intersection, $W_{N_0} + \delta N_0 \geq
c_{\alpha} + h > c_{\alpha}$ and $W_{N_0 + k} + \delta ( N_0 + k ) > (
c_{\alpha} + h ) - h$. The first event is $\cF_{N_0}$-measurable and has
probability $1 / 2$, as $W_{N_0} \sim \mathcal{N}( 0, \mu^2 N_0 )$. For
the second, let $\lambda := 2 \delta / \mu^2$; conditionally on
$\cF_{N_0}$, $M_k := \exp\big( - \lambda ( W_{N_0 + k} - W_{N_0} ) -
\lambda^2 \mu^2 k / 2 \big)$ is a nonnegative martingale with $M_0 = 1$,
and $W_{N_0 + k} - W_{N_0} + \delta k \leq - h$ implies $\log M_k \geq
\lambda h + k ( \lambda \delta - \lambda^2 \mu^2 / 2 ) = \lambda h$.
Ville's maximal inequality \citep{Ville1939} therefore gives
\[
    \lP\big( \exists k \geq 1 : W_{N_0 + k} - W_{N_0} + \delta k \leq -
    h \,\big|\, \cF_{N_0} \big)
    \;\leq\; e^{- \lambda h} = e^{- 2 \delta h / \mu^2}
    \;\leq\; e^{- \varepsilon^2 N_0 / ( 2 \mu^2 )}
    \;\leq\; \exp\Big( - \frac{( 1 - a^2 ) \lambda_0 \varepsilon^2}{2
    a^2 \mu^2} \Big) ,
\]
using $N_0 \geq n_0 = ( 1 - a^2 ) \lambda_0 / a^2$. Multiplying by $\lP(
W_{N_0} \geq 0 ) = 1 / 2$ gives the probability $p_0$.

\emph{Step 4: consequences.} On the locking event, $x_s x_s^{\top} = e_1
e_1^{\top}$ for every $s$, so $V_t = \mathrm{diag}( \lambda_0 + t - 1,
\lambda_0 )$ and $\lambda_{\min}( V_t ) = \lambda_0$; and
$\dotprod{\theta^{\star}}{x^{\star} - x_t} = a \mp \varepsilon \geq
\Delta$ at every round, so $\cR_T \geq \Delta T$.
\end{proof}

\section{Pseudo-code of \texorpdfstring{$\alpha$}{alpha}-LinUCB}
\phantomsection\label{sec:pseudo-code}

\Cref{alg:alpha-ucb} summarises the algorithm analysed in this paper; we
briefly comment on its steps, complementing the description given in
\cref{sec:introduction}.

\begingroup
\LinesNumbered
\SetAlgoVlined
\SetKwComment{Comment}{$\triangleright$\ }{}

\begin{algorithm2e}[h]
    \SetAlgorithmName{Algorithm}{}{}
    \caption{$\alpha$-LinUCB}
    \label{alg:alpha-ucb}
    \DontPrintSemicolon

    \textbf{Input:} Action set $\cX$, horizon $T$, regularization
    parameter $\lambda_0>0$, and exploration radius $\alpha>0$\;

    \textbf{Initialize:} $V_1\gets\lambda_0\I$\;

    \For{$t=1,\ldots,T$}{
        Compute the ridge estimate $\hat{\theta}_t$
        \label{step:Estimation}
        \Comment*[r]{\small Estimation}

        Select an action--parameter:
        \(
            (x_t,\theta_t)
            \in
            \argmax\limits_{(x,\theta)\in\cX\times\cC_t(\alpha)}
            \dotprod{x}{\theta}
        \)
        \label{step:Optimal_arms}
        \Comment*[r]{\small Decision rule}

        Play $x_t$ and observe the reward $y_t$
        \Comment*[r]{\small Feedback}

        Update $V_{t+1}\gets V_t+x_t x_t^\top$
        \Comment*[r]{\small Design-matrix update}
    }
\end{algorithm2e}
\endgroup

\paragraph{Estimation (line~\ref{step:Estimation}).}
The ridge estimate is $\hat{\theta}_t = V_t^{-1} \sum_{s=1}^{t-1} x_s
y_s$ with design matrix $V_t = \lambda_0 \I + \sum_{s=1}^{t-1} x_s
x_s^{\top}$; both admit rank-one updates, so each round costs
$\cO( d^2 )$ beyond the decision rule.

\paragraph{Decision rule (line~\ref{step:Optimal_arms}).}
The pair $( x_t, \theta_t )$ maximises $\dotprod{x}{\theta}$ jointly
over $\cX \times \cC_t( \alpha )$, where $\cC_t( \alpha ) = \{ \theta :
\norm{\theta - \hat{\theta}_t}_{V_t} \leq \alpha \}$. Optimising first
over $\theta$ at fixed $x$ gives the equivalent bonus form
\[
    x_t \in \argmax_{x \in \cX}
    \Big\{ \dotprod{\hat{\theta}_t}{x} + \alpha \norm{x}_{V_t^{-1}}
    \Big\} ,
\]
the familiar upper-confidence-bound rule, with $\norm{x}_{V_t^{-1}}$
largest along the directions least represented in the design;
conversely, once $x_t$ is chosen, $\theta_t = \hat{\theta}_t + \alpha
V_t^{-1} x_t / \norm{x_t}_{V_t^{-1}}$ attains the inner maximum.

\paragraph{The radius $\alpha$.}
The only departure from standard LinUCB is that the radius $\alpha$ is
an arbitrary input, held fixed over the run, rather than the
concentration width $\beta$ of \cref{prop:Yassin_conf}: for $\alpha =
\beta$ the algorithm is exactly the optimistic algorithm of
\citet{yassin}, while for $\alpha < \beta$ the set $\cC_t( \alpha )$
need not contain $\theta^{\star}$ at any round, and as $\alpha \to 0$
the rule approaches the greedy policy. Neither $\beta$
nor any property of the noise beyond the observations themselves is
used by the algorithm; they appear in the analysis only.

\section{Spectral growth}
\label{sec:spectral_growth_proof}

Let $\beta_{+} := \alpha + \beta$, where $\alpha > 0$ is the radius of
the set $\cC_t( \alpha )$ used by $\alpha$-LinUCB, no restriction on 
$\alpha$ is needed, and $\beta$ is the
radius of the valid confidence set $\cC_t( \beta )$ of
\cref{prop:Yassin_conf}; the confidence event is
\[
\cE := \Big\{ \theta^{\star} \in \bigcap_{t \in [T]} \cC_t( \beta ) \Big\}
= \big\{ \norm{\hat{\theta}_t - \theta^{\star}}_{V_t} \leq \beta
\ \text{ for all } t \in [T] \big\} ,
\qquad
\lP( \cE ) \geq 1 - \delta .
\]

\subsection{Preliminary results}

\paragraph{Localisation.}
We introduce the set $\cN_{\ast}$ of parameters lying within a controlled
$\norm{\cdot}_{\ast}$-distance of $\theta^{\star}$; on this set, both
$J_{\cX}$ and $\norm{\cdot}_{\ast}$ remain comparable to their values at
$\theta^{\star}$:
\begin{equation}
\label{eq:N-int}
\begin{gathered}
\cN_{\ast} := \Big\{ \theta \in \lR^d :
\norm{\theta - \theta^{\star}}_{\ast} \leq \frac{J_{\star}}{2 \ell_{\ast}} \Big\} ,
\\[2pt]
\tfrac12 J_{\star} \leq J_{\cX}( \theta ) \leq \tfrac32 J_{\star}
\quad \text{and} \quad
\tfrac12 \norm{\theta^{\star}}_{\ast} \leq \norm{\theta}_{\ast}
\leq \tfrac32 \norm{\theta^{\star}}_{\ast}
\qquad ( \theta \in \cN_{\ast} ) ,
\end{gathered}
\end{equation}
where the two-sided bounds follow from the $\ell_{\ast}$-Lipschitz
property of $J_{\cX}$ and from $J_{\star} \leq \ell_{\ast}
\norm{\theta^{\star}}_{\ast}$ (\cref{prop:global-int}\ref{it:gl-sandwich}
below); in particular $\cN_{\ast}$ is a convex set not containing the
origin.

\paragraph{Elementary propositions.}
The proofs rest on the following elementary propositions about the
support function and about the width $\omega_t = \max_{u \in
\mathbb{B}_2^d(1)} \norm{V_t^{-1/2} u}_{\ast}^2$;
none of them involves a Euclidean structure beyond the one implicit in
the formula for $\omega_t$, which \cref{prop:width-int}\ref{it:wint-dual}
removes.

\begin{restatable}[Consequences of the assumption]{restatableproposition}{propGlobalInt}
\phantomsection\label{prop:global-int}%
\label{prop:support-int}\label{prop:bregman-int}\label{prop:bregman-equiv-int}%
Under \cref{ass:armset-int}, with $\cX$ closed:
\begin{enumerate}[label=(\roman*), itemsep=1pt]
\item \label{it:gl-support}
$J_{\cX}$ is convex and positively homogeneous of degree one, $J_{\cX}^2$
is convex on $\lR^d$, $J_{\cX}( \theta ) > 0$ for every $\theta \neq 0$,
and $J_{\cX} = ( J_{\cX}^2 )^{1/2}$ is differentiable on $\lR^d \setminus
\{ 0 \}$.
\item \label{it:gl-sandwich}
$\frac{m}{2} \norm{v}_{\ast}^2 \leq J_{\cX}( v )^2 \leq \frac{M}{2}
\norm{v}_{\ast}^2$ for all $v \in \lR^d$. Consequently $\ell_{\ast} \leq
\sqrt{M / 2}$, $\cX$ is compact, $\abs{J_{\cX}( \theta ) - J_{\cX}(
\theta' )} \leq \ell_{\ast} \norm{\theta - \theta'}_{\ast}$ for all
$\theta, \theta'$ (in particular $J_{\cX}( \theta ) \leq \ell_{\ast}
\norm{\theta}_{\ast}$), and $J_{\star} \geq \sqrt{m / 2}\,
\norm{\theta^{\star}}_{\ast} > 0$.
\item \label{it:gl-argmax}
For every $\theta \neq 0$, $x \mapsto \dotprod{\theta}{x}$ has a unique
maximiser on $\cX$, namely $\nabla J_{\cX}( \theta )$; moreover
$\norm{\nabla J_{\cX}( \theta )}_{\ast}^{\circ} \leq \ell_{\ast}$ and
$\nabla J_{\cX}^2( \theta ) = 2 J_{\cX}( \theta ) \nabla J_{\cX}( \theta
)$. In particular $x^{\star} = \nabla J_{\cX}( \theta^{\star} )$ is the
unique optimal arm and \eqref{eq:scalars-upperbound} holds.
\item \label{it:gl-smooth}
the gradient of $J_{\cX}^2$ is $M$-Lipschitz from $( \lR^d,
\norm{\cdot}_{\ast} )$ to $( \lR^d, \norm{\cdot}_{\ast}^{\circ} )$:
\begin{equation}
\label{eq:sm-int}
\norm{\nabla J_{\cX}^2( \theta ) - \nabla J_{\cX}^2( \theta' )}_{\ast}^{\circ}
\;\leq\; M \norm{\theta - \theta'}_{\ast}
\qquad ( \theta, \theta' \in \lR^d ) .
\end{equation}
\item \label{it:gl-round}
At every round $t$ (recall $V_t \succeq \lambda_0 \I \succ 0$), the optimistic parameter
satisfies $\theta_t \neq 0$ and the played arm is $x_t = \nabla J_{\cX}(
\theta_t )$.
\end{enumerate}
\end{restatable}

\begin{proof}[\cref{prop:global-int}]
\ref{it:gl-support} $J_{\cX}$ is a supremum of linear functions, hence
convex and positively homogeneous, with $J_{\cX}( 0 ) = 0$. Since $\cX$
contains a neighbourhood of the origin and all norms on $\lR^d$ are
equivalent, there is $\varepsilon > 0$ with $\{ x : \norm{x}_{\ast}^{\circ}
\leq \varepsilon \} \subseteq \cX$, so $J_{\cX}( \theta ) \geq
\sup_{\norm{x}_{\ast}^{\circ} \leq \varepsilon} \dotprod{\theta}{x} =
\varepsilon \norm{\theta}_{\ast} > 0$ for $\theta \neq 0$. As $J_{\cX}
\geq 0$ and $s \mapsto s^2$ is convex and nondecreasing on $[ 0, \infty
)$, $J_{\cX}^2$ is convex; as $J_{\cX}^2$ is differentiable and positive
off the origin, so is $J_{\cX} = ( J_{\cX}^2 )^{1/2}$.

\ref{it:gl-sandwich} $J_{\cX}^2( 0 ) = 0$ and $0$ minimises the
differentiable function $J_{\cX}^2 \geq 0$, so $\nabla J_{\cX}^2( 0 ) =
0$; hence $D_{J_{\cX}^2}( v, 0 ) = J_{\cX}( v )^2$, and \eqref{eq:ajp-int}
at $\theta' = 0$ is the sandwich. For $x \in \cX$, $\norm{x}_{\ast}^{\circ}
= \sup_{\norm{v}_{\ast} \leq 1} \dotprod{x}{v} \leq \sup_{\norm{v}_{\ast}
\leq 1} J_{\cX}( v ) \leq \sqrt{M / 2}$, so $\ell_{\ast} \leq \sqrt{M /
2}$; thus $\cX$ is bounded and, being closed, compact. If $x \in \cX$
maximises $\dotprod{\theta}{\cdot}$ (a maximiser exists by compactness),
then $J_{\cX}( \theta ) - J_{\cX}( \theta' ) \leq \dotprod{\theta}{x} -
\dotprod{\theta'}{x} \leq \norm{x}_{\ast}^{\circ} \norm{\theta -
\theta'}_{\ast} \leq \ell_{\ast} \norm{\theta - \theta'}_{\ast}$, because
$J_{\cX}( \theta' ) \geq \dotprod{\theta'}{x}$; exchange the roles of
$\theta, \theta'$, and take $\theta' = 0$ for the bound $J_{\cX}( \theta )
\leq \ell_{\ast} \norm{\theta}_{\ast}$. Finally, the sandwich at $v =
\theta^{\star} \neq 0$ gives $J_{\star} \geq \sqrt{m / 2}\,
\norm{\theta^{\star}}_{\ast} > 0$.

\ref{it:gl-argmax} A maximiser $x \in \cX$ exists by compactness. For
every $\theta'$, $J_{\cX}( \theta' ) \geq \dotprod{\theta'}{x} = J_{\cX}(
\theta ) + \dotprod{\theta' - \theta}{x}$, i.e.\ $x$ is a subgradient of
the convex function $J_{\cX}$ at $\theta$; as $J_{\cX}$ is
differentiable at $\theta \neq 0$ by~\ref{it:gl-support}, its
subdifferential there is $\{ \nabla J_{\cX}( \theta ) \}$ ~\citep[Thm.~25.1]{RockafellarWets1998}, so every maximiser equals $\nabla
J_{\cX}( \theta ) \in \cX$, whence $\norm{\nabla J_{\cX}( \theta
)}_{\ast}^{\circ} \leq \ell_{\ast}$; the chain rule gives $\nabla
J_{\cX}^2 = 2 J_{\cX} \nabla J_{\cX}$ off the origin. At $\theta =
\theta^{\star}$, $J_{\star} = \dotprod{\theta^{\star}}{x^{\star}} \leq
\norm{x^{\star}}_{\ast}^{\circ} \norm{\theta^{\star}}_{\ast}$ gives
$\varrho_{\ast} \leq \norm{x^{\star}}_{\ast}^{\circ} \leq \ell_{\ast}$,
which together with \ref{it:gl-sandwich} is~\eqref{eq:scalars-upperbound}.

\ref{it:gl-smooth} For a convex function that is differentiable on
$\lR^d$, the upper bound in~\eqref{eq:ajp-int} is equivalent to
$M$-Lipschitz continuity of the gradient from $( \lR^d, \norm{\cdot}_{\ast}
)$ to $( \lR^d, \norm{\cdot}_{\ast}^{\circ} )$~\citep[Thm.~5.8]{Beck2017}; apply this to $J_{\cX}^2$,
which is convex by~\ref{it:gl-support}.

\ref{it:gl-round} Since $V_t \succ 0$, the confidence set $\cC_t( \alpha )$ is
compact with nonempty interior, so it contains some $\theta \neq 0$,
and $J_{\cX}( \theta ) > 0$ by~\ref{it:gl-support}. By definition of the
optimistic pair,
\[
\dotprod{\theta_t}{x_t}
= \max_{\theta \in \cC_t( \alpha )} \max_{x \in \cX} \dotprod{\theta}{x}
= \max_{\theta \in \cC_t( \alpha )} J_{\cX}( \theta )
\;>\; 0 = J_{\cX}( 0 ) ,
\]
so $\theta_t \neq 0$; and $x_t$ maximises $\dotprod{\theta_t}{\cdot}$ on
$\cX$, so $x_t = \nabla J_{\cX}( \theta_t )$ by~\ref{it:gl-argmax}.
\end{proof}

\begin{restatable}[The width in the $\norm{\cdot}_{\ast}$ geometry]{restatableproposition}{propWidthInt}
\phantomsection\label{prop:width-int}
For every $t$, let $\omega_t = \max_{u \in \mathbb{B}_2^d(1)}
\norm{V_t^{-1/2} u}_{\ast}^2$ be the width. Then:
\begin{enumerate}[label=(\roman*), itemsep=1pt]
\item \label{it:wint-dual}
$\omega_t = \max\{ \norm{v}_{\ast}^2 : \norm{v}_{V_t} \leq 1 \}
= \max_{\norm{g}_{\ast}^{\circ} \leq 1} \norm{g}_{V_t^{-1}}^2$, all
maxima being attained;
\item \label{it:wint-radius}
for every $r > 0$, $\ \max\{ \norm{v}_{\ast} : \norm{v}_{V_t} \leq r \}
= r\, \omega_t^{1/2}$; in particular
\[
\max_{\theta \in \cC_t( \alpha )} \norm{\theta - \hat{\theta}_t}_{\ast}
= \alpha\, \omega_t^{1/2}
\qquad \text{and, on } \cE, \qquad
\norm{\hat{\theta}_t - \theta^{\star}}_{\ast} \leq \beta\, \omega_t^{1/2} ;
\]
\item \label{it:wint-energy}
$\norm{g}_{V_t^{-1}} \leq \norm{g}_{\ast}^{\circ}\, \omega_t^{1/2}$ for
every $g \in \lR^d$; in particular $x^{\top} V_t^{-1} x \leq
\ell_{\ast}^2\, \omega_t$ for every $x \in \cX$;
\item \label{it:wint-sandwich}
if $\norm{\cdot}_{\mathrm{E}}$ is a Euclidean norm on $\lR^d$ with
$\kappa' \norm{u}_{\mathrm{E}} \leq \norm{u}_{\ast} \leq \kappa
\norm{u}_{\mathrm{E}}$, and $\lambda_{\min}( V_t )$ denotes the
smallest eigenvalue of $V_t$ in $\norm{\cdot}_{\mathrm{E}}$-orthonormal
coordinates, then $\kappa'^2 / \lambda_{\min}( V_t ) \leq \omega_t \leq
\kappa^2 / \lambda_{\min}( V_t )$;
\item \label{it:wint-monotone}
$t \mapsto \omega_t$ is nonincreasing.
\end{enumerate}
\end{restatable}

\begin{proof}[\cref{prop:width-int}]
\ref{it:wint-dual} Substituting $v = V_t^{-1/2} u$ in the definition
of $\omega_t$ gives $\omega_t = \max\{ \norm{v}_{\ast}^2 : \norm{v}_{V_t} \leq 1 \}$.
Since $\norm{v}_{\ast} = \sup_{\norm{g}_{\ast}^{\circ} \leq 1}
\dotprod{g}{v}$ (duality of norms) and $\sup_{\norm{v}_{V_t} \leq 1}
\dotprod{g}{v} = \norm{g}_{V_t^{-1}}$ (duality of the quadratic norms
$\norm{\cdot}_{V_t}$ and $\norm{\cdot}_{V_t^{-1}}$), exchanging the
two suprema gives $\max\{ \norm{v}_{\ast} : \norm{v}_{V_t} \leq 1 \}
= \max_{\norm{g}_{\ast}^{\circ} \leq 1} \norm{g}_{V_t^{-1}}$; square.
All are suprema of continuous functions over compact sets.
\ref{it:wint-radius} $\{ v : \norm{v}_{V_t} \leq r \} = r \{ v :
\norm{v}_{V_t} \leq 1 \}$ and $\norm{\cdot}_{\ast}$ is homogeneous, so
the maximum is $r\, \omega_t^{1/2}$ by~\ref{it:wint-dual}; apply this to
$v = \theta - \hat{\theta}_t$ with $r = \alpha$, and to $v =
\hat{\theta}_t - \theta^{\star}$ with $r = \beta$ on $\cE$.
\ref{it:wint-energy} $\norm{g}_{V_t^{-1}} = \sup_{\norm{v}_{V_t}
\leq 1} \dotprod{g}{v} \leq \sup_{\norm{v}_{V_t} \leq 1}
\norm{g}_{\ast}^{\circ} \norm{v}_{\ast} = \norm{g}_{\ast}^{\circ}\,
\omega_t^{1/2}$, and $\norm{x}_{\ast}^{\circ} \leq \ell_{\ast}$ for $x \in
\cX$.
\ref{it:wint-sandwich} In $\norm{\cdot}_{\mathrm{E}}$-orthonormal
coordinates, $\omega_t = \sup_{\norm{u}_{\mathrm{E}} \leq 1}
\norm{V_t^{-1/2} u}_{\ast}^2$, where
\[
\kappa'^2 \norm{V_t^{-1/2} u}_{\mathrm{E}}^2
\;\leq\; \norm{V_t^{-1/2} u}_{\ast}^2
\;\leq\; \kappa^2 \norm{V_t^{-1/2} u}_{\mathrm{E}}^2
\qquad \text{and} \qquad
\sup_{\norm{u}_{\mathrm{E}} \leq 1} \norm{V_t^{-1/2} u}_{\mathrm{E}}^2
= \frac{1}{\lambda_{\min}( V_t )} .
\]
\ref{it:wint-monotone} For each fixed $g$, $g^{\top} V_t^{-1} g$ is
nonincreasing in $t$ because $V_{t+1}^{-1} \preceq V_t^{-1}$; take
the maximum over $\norm{g}_{\ast}^{\circ} \leq 1$ in~\ref{it:wint-dual}.
\end{proof}

\begin{restatable}[Lipschitz argmax]{restatableproposition}{propLipInt}
\phantomsection\label{prop:lip-int}
Under \cref{ass:armset-int}, for all $\theta, \theta' \in \cN_{\ast}$,
\begin{equation}
\label{eq:lip-int}
\norm{\nabla J_{\cX}( \theta ) - \nabla J_{\cX}( \theta' )}_{\ast}^{\circ}
\;\leq\; L_{\ast}\, \norm{\theta - \theta'}_{\ast} ,
\qquad
L_{\ast} = \frac{M + 2 \ell_{\ast}^2}{J_{\star}} .
\end{equation}
Lipschitz continuity of the argmax map is the condition under which
\citet{rusmevichientong2010linearly} call an arm set strongly
convex; here it is derived, in the geometry of the assumption, from
the smoothness of $J_{\cX}^2$ and the localisation $J_{\cX} \geq
J_{\star} / 2$ on $\cN_{\ast}$.
\end{restatable}

\begin{proof}[\cref{prop:lip-int}]
Since $0 \notin \cN_{\ast}$, $\nabla J_{\cX} = \nabla J_{\cX}^2 / ( 2
J_{\cX} )$ on $\cN_{\ast}$ (\cref{prop:global-int}\ref{it:gl-argmax}), so
\begin{align*}
\norm{\nabla J_{\cX}( \theta ) - \nabla J_{\cX}( \theta' )}_{\ast}^{\circ}
&\leq \frac{\norm{\nabla J_{\cX}^2( \theta ) - \nabla J_{\cX}^2( \theta' )}_{\ast}^{\circ}}
{2 J_{\cX}( \theta )}
+ \frac{\norm{\nabla J_{\cX}^2( \theta' )}_{\ast}^{\circ}\,
\abs{J_{\cX}( \theta' ) - J_{\cX}( \theta )}}
{2 J_{\cX}( \theta ) J_{\cX}( \theta' )} \\
&\leq \frac{M \norm{\theta - \theta'}_{\ast}}{J_{\star}}
+ \frac{2 \ell_{\ast}^2 \norm{\theta - \theta'}_{\ast}}{J_{\star}} ,
\end{align*}
where we used~\eqref{eq:sm-int}, $J_{\cX}( \theta ) \geq J_{\star} / 2$
from~\eqref{eq:N-int}, $\norm{\nabla J_{\cX}^2( \theta' )}_{\ast}^{\circ}
= 2 J_{\cX}( \theta' ) \norm{\nabla J_{\cX}( \theta' )}_{\ast}^{\circ}
\leq 2 \ell_{\ast} J_{\cX}( \theta' )$ (\cref{prop:global-int}\ref{it:gl-argmax}),
and the $\ell_{\ast}$-Lipschitz property of $J_{\cX}$.
\end{proof}

\subsection{Geometric distortion and adapted coordinates}
\phantomsection\label{sec:distortion}

\paragraph{Adapted Euclidean structure.}
The proofs perform spectral bookkeeping (eigenvalues and eigenvectors
of $V_t$, a top direction, a band of small eigenvalues, a
determinant), which requires a Euclidean structure. Since
$\alpha$-LinUCB is equivariant under linear reparametrisations and
$\omega_t$ is intrinsic, this structure is a proof device, and it can be
chosen adapted to $\norm{\cdot}_{\ast}$. Call $D_{\ast} \geq 1$ an
\emph{admissible distortion} if there is $\bP \succ 0$ with
\begin{equation}
\label{eq:adapted-position}
\norm{u}_{\ast} \;\leq\; \norm{\bP u}_2 \;\leq\; D_{\ast} \norm{u}_{\ast}
\qquad ( u \in \lR^d ) ,
\end{equation}
that is, the Euclidean unit ball $E$ of $u \mapsto \norm{\bP u}_2$
satisfies $D_{\ast}^{-1} B_{\ast} \subseteq E \subseteq B_{\ast}$; the
smallest admissible value is the Banach--Mazur distance of $B_{\ast}$
to the Euclidean ball. By John's theorem $D_{\ast} = \sqrt{d}$ is
admissible, and under \cref{ass:armset-int} so is $D_{\ast} = \sqrt{M /
m}$ (\cref{lem:hessian-int}, from the Hessian of $J_{\cX}^2$ at a
point of twice differentiability); when $B_{\ast}$ is an ellipsoid,
$D_{\ast} = 1$. We fix once and for all an admissible pair $( D_{\ast},
\bP )$ with
\begin{equation}
\label{eq:Dstar}
D_{\ast} \;\leq\; \min\big\{ \sqrt{d},\ \sqrt{M / m} \big\} ,
\end{equation}
and use it only inside the proofs (\cref{lem:adapted-int}). The
distortion $D_{\ast}$ enters the results through the burn-in threshold,
through $T_\star$, and through $\Log{8 \Gamma}$ (where $4
\Log{D_{\ast}} \leq 2 \Log{d}$); it never multiplies the rate constant.

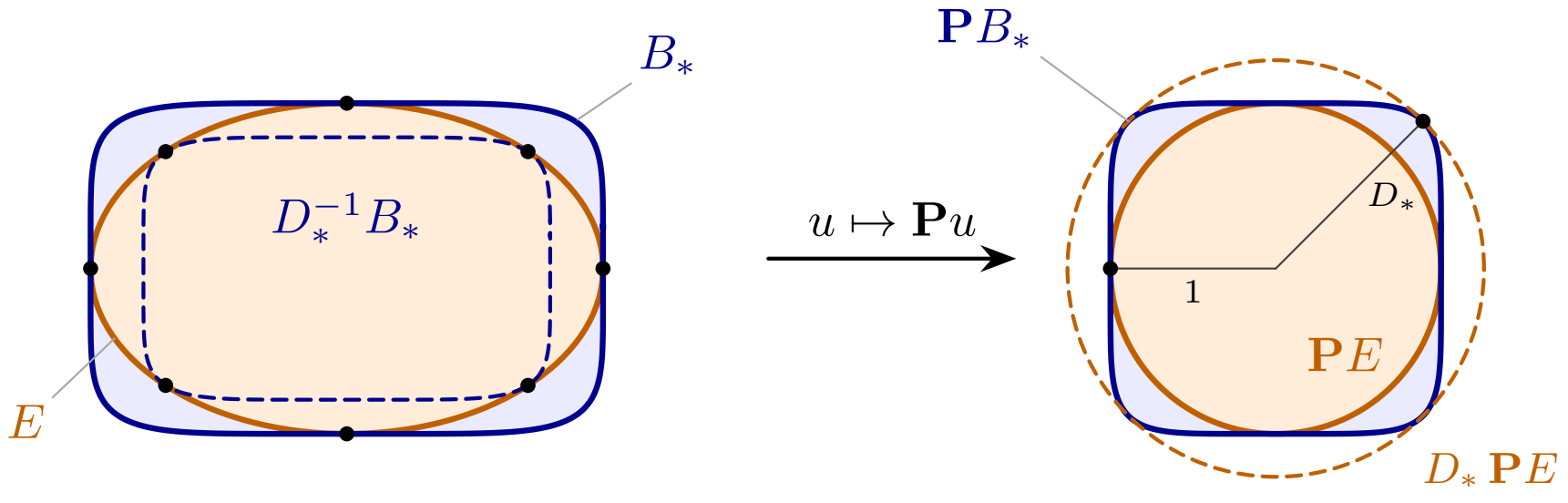
\begin{figure}[h]
\centering
\begin{tikzpicture}[scale=1.20, line cap=round, line join=round,
    >={Stealth[length=2.6mm]},
    bstar/.style={draw=blue!55!black, very thick},
    eball/.style={draw=orange!75!black, very thick},
    scaled/.style={densely dashed, thick},
    tang/.style={fill=black, draw=none}]
\def\figA{1.55}   %
\def\figB{1.00}   %
\def\figD{1.2599} %

\begin{scope}
  \fill[blue!8]
    plot[domain=0.75:360.75, samples=181, smooth cycle]
    ({\figA*sign(cos(\x))*abs(cos(\x))^(1/3)},
     {\figB*sign(sin(\x))*abs(sin(\x))^(1/3)});
  \fill[orange!15] (0,0) ellipse ({\figA} and {\figB});
  \draw[eball]    (0,0) ellipse ({\figA} and {\figB});
  \draw[bstar]
    plot[domain=0.75:360.75, samples=181, smooth cycle]
    ({\figA*sign(cos(\x))*abs(cos(\x))^(1/3)},
     {\figB*sign(sin(\x))*abs(sin(\x))^(1/3)});
  \draw[scaled, blue!55!black]
    plot[domain=0.75:360.75, samples=181, smooth cycle]
    ({\figA/\figD*sign(cos(\x))*abs(cos(\x))^(1/3)},
     {\figB/\figD*sign(sin(\x))*abs(sin(\x))^(1/3)});
  \foreach \px/\py in {\figA/0, -\figA/0, 0/\figB, 0/-\figB}
    \fill[tang] (\px,\py) circle (1.3pt);
  \foreach \sx in {1,-1} \foreach \sy in {1,-1}
    \fill[tang] ({\sx*\figA*0.70711},{\sy*\figB*0.70711}) circle (1.3pt);
  \node[blue!55!black]   at (0, 0.30) {$D_{\ast}^{-1} B_{\ast}$};
  \draw[gray!70, thin] (1.395, 0.90) -- (1.72, 1.12)
    node[above right=-2pt, blue!55!black] {$B_{\ast}$};
  \draw[gray!70, thin] ({\figA*cos(205)}, {\figB*sin(205)}) -- (-1.78, -0.78)
    node[below left=-2pt, orange!75!black] {$E$};
  \node[align=center, font=\small] at (0, -1.78)
    {(a) original coordinates\\[1pt]
     $D_{\ast}^{-1} B_{\ast} \subseteq E \subseteq B_{\ast}$};
\end{scope}

\draw[->, thick] (2.55, 0.06) -- (4.05, 0.06)
  node[midway, above=1pt] {$u \mapsto \bP u$}
  node[midway, below=1pt, font=\scriptsize, text=gray!50!black]
    {};

\begin{scope}[xshift=5.62cm]
  \fill[blue!8]
    plot[domain=0.75:360.75, samples=181, smooth cycle]
    ({sign(cos(\x))*abs(cos(\x))^(1/3)},
     {sign(sin(\x))*abs(sin(\x))^(1/3)});
  \fill[orange!15] (0,0) circle (1);
  \draw[eball]     (0,0) circle (1);
  \draw[bstar]
    plot[domain=0.75:360.75, samples=181, smooth cycle]
    ({sign(cos(\x))*abs(cos(\x))^(1/3)},
     {sign(sin(\x))*abs(sin(\x))^(1/3)});
  \draw[scaled, orange!75!black] (0,0) circle (\figD);
  \draw[thin, gray!50!black] (0,0) -- (-1,0)
    node[midway, below=-1pt, font=\scriptsize, text=black] {$1$};
  \draw[thin, gray!50!black] (0,0) -- (0.8909, 0.8909)
    node[pos=0.60, below right=-3pt, font=\scriptsize, text=black] {$D_{\ast}$};
  \fill[tang] (-1,0) circle (1.3pt);
  \fill[tang] (0.8909, 0.8909) circle (1.3pt);
  \node[orange!75!black] at (0.42, -0.52) {$\bP E$};
  \draw[gray!70, thin] (-0.8909, 0.8909) -- (-1.42, 1.28)
    node[above left=-2pt, blue!55!black] {$\bP B_{\ast}$};
  \node[orange!75!black, font=\small] at (1.30, -1.22) {$D_{\ast}\, \bP E$};
  \node[align=center, font=\small] at (0, -1.78)
    {(b) adapted coordinates\\[1pt]
     $\bP E \subseteq \bP B_{\ast} \subseteq D_{\ast}\, \bP E$};
\end{scope}
\end{tikzpicture}
\caption{The adapted Euclidean structure of \cref{sec:distortion}.
\emph{(a)}~The unit ball $B_{\ast}$ of $\norm{\cdot}_{\ast}$ (blue), the
Euclidean unit ball $E$ of $u \mapsto \norm{\bP u}_2$ (orange) inscribed in
it, and the rescaled copy $D_{\ast}^{-1} B_{\ast}$ (dashed); dots mark
tangencies.
\emph{(b)}~The reparametrisation $u \mapsto \bP u$ makes $E$ round and
sandwiches $B_{\ast}$ between two Euclidean balls whose radii differ
exactly by the factor $D_{\ast}$; this is \eqref{eq:adapted-position},
where all spectral bookkeeping takes place.
}
\label{fig:distortion}
\end{figure}

\begin{restatable}[Ellipsoidal distortion of $B_{\ast}$ under the assumption]{restatablelemma}{lemHessianInt}
\phantomsection\label{lem:hessian-int}
Under \cref{ass:armset-int}, there is a symmetric matrix $\bH
\succ 0$ with
\[
m \norm{u}_{\ast}^2 \;\leq\; u^{\top} \bH u \;\leq\; M \norm{u}_{\ast}^2
\qquad ( u \in \lR^d ) .
\]
Consequently $\bP := ( \bH / m )^{1/2}$ satisfies
\eqref{eq:adapted-position} with $D_{\ast} = \sqrt{M / m}$: the
assumption itself forbids $B_{\ast}$ from being far from an ellipsoid,
and John's $\sqrt{d}$ can always be replaced by $\min\{ \sqrt{d},
\sqrt{M / m} \}$.
\end{restatable}

\begin{proof}[\cref{lem:hessian-int}]
$J_{\cX}^2$ is convex and finite on $\lR^d$
(\cref{prop:global-int}\ref{it:gl-support}). By Alexandrov's theorem
\citep{Alexandrov1939} (see also \citealp[Thm.~13.51]{RockafellarWets1998},
or \citealp[Thm.~6.9]{EvansGariepy2015}), it is twice differentiable at
almost every $\theta_0 \in \lR^d$: there is a symmetric $\bH$ with
\[
D_{J_{\cX}^2}( \theta_0 + v, \theta_0 )
= J_{\cX}^2( \theta_0 + v ) - J_{\cX}^2( \theta_0 )
- \dotprod{\nabla J_{\cX}^2( \theta_0 )}{v}
= \tfrac12 v^{\top} \bH v + o( \norm{v}_{\ast}^2 ) .
\]
Fix such a $\theta_0$. For $u \in \lR^d$ and $\varepsilon > 0$,
\eqref{eq:ajp-int} with $\theta = \theta_0 + \varepsilon u$ and $\theta'
= \theta_0$ gives
\[
\frac{m}{2}\, \varepsilon^2 \norm{u}_{\ast}^2
\;\leq\; \frac{\varepsilon^2}{2}\, u^{\top} \bH u + o( \varepsilon^2 )
\;\leq\; \frac{M}{2}\, \varepsilon^2 \norm{u}_{\ast}^2 .
\]
Dividing by $\varepsilon^2 / 2$ and letting $\varepsilon \to 0$ yields
$m \norm{u}_{\ast}^2 \leq u^{\top} \bH u \leq M \norm{u}_{\ast}^2$;
in particular $\bH \succ 0$. With $\bP = ( \bH / m )^{1/2}$,
$\norm{\bP u}_2^2 = u^{\top} \bH u / m \in [ \norm{u}_{\ast}^2,
( M / m ) \norm{u}_{\ast}^2 ]$, which is~\eqref{eq:adapted-position}
with $D_{\ast} = \sqrt{M / m}$.
\end{proof}

\begin{restatable}[Adapted coordinates]{restatablelemma}{lemAdaptedInt}
\phantomsection\label{lem:adapted-int}
Let $( D_{\ast}, \bP )$ be an admissible pair~\eqref{eq:adapted-position}
and define the linear change of variables
\[
\tilde{x} := \frac{\bP^{-1} x}{\ell_{\ast}} ,
\qquad
\tilde{\theta} := \ell_{\ast}\, \bP\, \theta ,
\qquad \text{so that} \qquad
\dotprod{\tilde{\theta}}{\tilde{x}} = \dotprod{\theta}{x} ,
\]
and let $\tilde{\cX} := \{ \tilde{x} : x \in \cX \}$ and
$\tilde{\theta}^{\star} := \ell_{\ast} \bP \theta^{\star}$. The norms
$\norm{\cdot}_{\ast}$ and $\norm{\cdot}_{\ast}^{\circ}$ are not changed:
they are simply read through the change of variables, i.e.\ a point
described by $\tilde{\theta}$ has $\norm{\theta}_{\ast} = \norm{\bP^{-1}
\tilde{\theta}}_{\ast} / \ell_{\ast}$ and an arm described by $\tilde{x}$
has $\norm{x}_{\ast}^{\circ} = \norm{\ell_{\ast} \bP \tilde{x}}_{\ast}^{\circ}$.
\begin{enumerate}[label=(\roman*), itemsep=1pt]
\item \label{it:ad-arms}
$\norm{x}_{\ast}^{\circ} / ( D_{\ast} \ell_{\ast} ) \leq \norm{\tilde{x}}_2
\leq \norm{x}_{\ast}^{\circ} / \ell_{\ast}$ for every $x$; in particular
$\tilde{\cX} \subseteq \mathbb{B}_2^d(1)$ and $\norm{\tilde{x}^{\star}}_2
\geq \varrho_{\ast} / ( D_{\ast} \ell_{\ast} )$.
\item \label{it:ad-params}
$\norm{\tilde{\theta}}_2 / ( D_{\ast} \ell_{\ast} ) \leq \norm{\theta}_{\ast}
\leq \norm{\tilde{\theta}}_2 / \ell_{\ast}$ for every $\theta$.
\item \label{it:ad-equivariance}
Every intrinsic quantity is unchanged: $J_{\tilde{\cX}}( \tilde{\theta} ) =
J_{\cX}( \theta )$, the gradient $\nabla J_{\tilde{\cX}}( \tilde{\theta} )
= \bP^{-1} \nabla J_{\cX}( \theta ) / \ell_{\ast}$ is the image of $\nabla
J_{\cX}( \theta )$ (gradients transform like arms), Bregman divergences
of $J^2$, distances $\norm{\theta - \theta'}_{\ast}$ and $\norm{x -
x'}_{\ast}^{\circ}$, the scalars $\ell_{\ast}$, $\varrho_{\ast}$, $J_{\star}$
and the ball $\cN_{\ast}$ (mapped onto $\ell_{\ast} \bP \cN_{\ast}$) are
the same; so \cref{ass:armset-int} holds for the problem $( \tilde{\cX},
\tilde{\theta}^{\star} )$ with the same $m, M$. Moreover $\alpha$-LinUCB
is equivariant: with
\[
\tilde{V}_t := \ell_{\ast}^{-2} \bP^{-1} V_t \bP^{-1}
= \tilde{\bR} + \sum_{s < t} \tilde{x}_s \tilde{x}_s^{\top} ,
\qquad
\tilde{\bR} := \lambda_0\, \ell_{\ast}^{-2} \bP^{-2} ,
\]
the transported design matrix (the ridge penalty $\lambda_0
\norm{\theta}_2^2 = \tilde{\theta}^{\top} \tilde{\bR}\, \tilde{\theta}$ is
no longer isotropic), and $\hat{\theta}_t = V_t^{-1} \sum_{s < t} x_s
y_s$ the ridge estimate,
\[
\tilde{V}_t^{-1} \sum_{s < t} \tilde{x}_s y_s = \ell_{\ast} \bP\, \hat{\theta}_t ,
\qquad
\norm{\tilde{\theta} - \tilde{\theta}'}_{\tilde{V}_t} = \norm{\theta - \theta'}_{V_t} ,
\qquad
\norm{\tilde{g}}_{\tilde{V}_t^{-1}} = \norm{g}_{V_t^{-1}},
\]
so the confidence ellipsoid, the optimistic pair and the event $\cE$
are the images of the original ones, $( \tilde{x}_t, \tilde{\theta}_t
)_t$ is a trajectory of $\alpha$-LinUCB on $( \tilde{\cX},
\tilde{\theta}^{\star} )$ with the same rewards, and the width is unchanged: $\tilde{\omega}_t = \omega_t$.
\item \label{it:ad-floor}
In the new coordinates the smallest eigenvalue satisfies
$\tilde{\lambda}_{t,d} \geq \omega_t^{-1} / ( D_{\ast}^2 \ell_{\ast}^2 )$, and
$x^{\top} V_t^{-1} x = \tilde{x}^{\top} \tilde{V}_t^{-1} \tilde{x}
\leq \min\{ 1 / \tilde{\lambda}_{t,d},\ \ell_{\ast}^2\, \omega_t \}$ for every
$x \in \cX$; in particular, at every round with $\omega_t^{-1} \geq 2 D_{\ast}^2
\ell_{\ast}^2$, one has $\tilde{\lambda}_{t,d} \geq 2$ and $x^{\top}
V_t^{-1} x \leq 1 / 2$.
\end{enumerate}
\end{restatable}

\begin{proof}[\cref{lem:adapted-int}]
\ref{it:ad-arms} $\norm{\bP^{-1} x}_2 = \sup_{\norm{v}_2 \leq 1}
\dotprod{x}{\bP^{-1} v} = \sup_{\norm{\bP u}_2 \leq 1} \dotprod{x}{u}$.
By~\eqref{eq:adapted-position} the constraint set contains $\{ u :
\norm{u}_{\ast} \leq 1 / D_{\ast} \}$ and is contained in $\{ u :
\norm{u}_{\ast} \leq 1 \}$, so $\norm{x}_{\ast}^{\circ} / D_{\ast} \leq
\norm{\bP^{-1} x}_2 \leq \norm{x}_{\ast}^{\circ}$; divide by $\ell_{\ast}$
and use $\norm{x^{\star}}_{\ast}^{\circ} \geq \varrho_{\ast}$.
\ref{it:ad-params} is \eqref{eq:adapted-position} multiplied by
$\ell_{\ast}$, since $\norm{\tilde{\theta}}_2 = \ell_{\ast} \norm{\bP \theta}_2$.
\ref{it:ad-equivariance} The pairing is invariant, so $J_{\tilde{\cX}}(
\tilde{\theta} ) = \sup_{x \in \cX} \dotprod{\theta}{x} = J_{\cX}(
\theta )$; the chain rule gives $\nabla_{\tilde{\theta}} J_{\tilde{\cX}}
= \ell_{\ast}^{-1} \bP^{-1} \nabla_{\theta} J_{\cX}$, i.e.\ gradients
transform like arms, hence $\dotprod{\nabla J_{\tilde{\cX}}^2(
\tilde{\theta}' )}{\tilde{\theta} - \tilde{\theta}'} = \dotprod{\nabla
J_{\cX}^2( \theta' )}{\theta - \theta'}$ and Bregman divergences are
unchanged. The norms $\norm{\cdot}_{\ast}$, $\norm{\cdot}_{\ast}^{\circ}$
are by definition evaluated on the original points, so distances,
$\ell_{\ast} = \sup_{\cX} \norm{\cdot}_{\ast}^{\circ}$, $\varrho_{\ast} =
J_{\star} / \norm{\theta^{\star}}_{\ast}$, $J_{\star}$ and $\cN_{\ast}$
are unchanged, and \eqref{eq:ajp-int} and~\eqref{eq:sm-int} hold verbatim
for the transported problem. For the algorithm: $\tilde{V}_t =
\ell_{\ast}^{-2} \bP^{-1} ( \lambda_0 \I + \sum_{s < t} x_s x_s^{\top} )
\bP^{-1} = \tilde{\bR} + \sum_{s < t} \tilde{x}_s \tilde{x}_s^{\top}$, so
$\tilde{V}_t^{-1} = \ell_{\ast}^2 \bP V_t^{-1} \bP$ and $\tilde{V}_t^{-1}
\sum_s \tilde{x}_s y_s = \ell_{\ast} \bP V_t^{-1} \sum_s x_s y_s =
\ell_{\ast} \bP \hat{\theta}_t$; the two norm identities follow by
direct substitution. Hence $\tilde{\cC}_t := \{ \tilde{\theta} : \theta
\in \cC_t( \alpha ) \}$, the maximisers of the invariant pairing over $\tilde{\cX}
\times \tilde{\cC}_t$ are the images of those over $\cX \times \cC_t( \alpha )$
(for each $\theta \neq 0$ the maximising arm is unique,
\cref{prop:global-int}\ref{it:gl-argmax}, and the analysis below never
uses a particular selection of $\theta_t$ among maximising parameters),
and the event $\cE$ is
the same event. Finally
$\tilde{\omega}_t = \max\{ \norm{\theta}_{\ast}^2 : \norm{\tilde{\theta}}_{\tilde{V}_t}
\leq 1 \} = \max\{ \norm{\theta}_{\ast}^2 : \norm{\theta}_{V_t} \leq 1 \}
= \omega_t$ by \cref{prop:width-int}\ref{it:wint-dual} and the second
identity.
\ref{it:ad-floor} By \ref{it:ad-params} and
\cref{prop:width-int}\ref{it:wint-sandwich} with $\kappa' = 1 / ( D_{\ast}
\ell_{\ast} )$, $\tilde{\lambda}_{t,d} \geq \omega_t^{-1} / ( D_{\ast}^2
\ell_{\ast}^2 )$. The energy bound is
\cref{prop:width-int}\ref{it:wint-energy} together with
$\norm{\tilde{x}}_2 \leq 1$.
\end{proof}

\paragraph{Convention.}
Throughout the proofs we keep the tildes of \cref{lem:adapted-int}:
$\tilde{x}$, $\tilde{\theta}$, $\tilde{V}_t$ are the adapted coordinates
of $x$, $\theta$, $V_t$, and every Euclidean object --- the norm
$\norm{\cdot}_2$, eigenvalues, eigenvectors, traces, determinants ---
is computed on tilded quantities. Untilded symbols denote the original
objects, and the norms $\norm{\cdot}_{\ast}$, $\norm{\cdot}_{\ast}^{\circ}$
are always applied to original objects; a bound such as
$\norm{\tilde{x}}_2 \leq \norm{x}_{\ast}^{\circ} / \ell_{\ast}$ therefore
compares the Euclidean length of the adapted coordinate vector of a
point with the intrinsic dual norm of that point. In this notation,
\cref{lem:adapted-int}\ref{it:ad-arms}, \ref{it:ad-params} and
\ref{it:ad-floor} read: $\tilde{\cX} \subseteq \mathbb{B}_2^d(1)$,
$\norm{\tilde{x}_s}_2 \leq 1$, $\Tr \tilde{V}_t \leq t - 1 + r_{\lambda_0}$,
where
\begin{equation}
\label{eq:rlambda}
r_{\lambda_0} := \Tr \tilde{\bR} = \lambda_0\, \ell_{\ast}^{-2} \Tr \bP^{-2}
\;\leq\; \frac{\lambda_0\, d}{\ell_{\ast}^2\, \kappa'^2} ,
\qquad
\kappa' := \min_{\norm{u}_2 = 1} \norm{u}_{\ast} ,
\end{equation}
is the trace of the transported regulariser (by~\eqref{eq:adapted-position},
$\lambda_{\min}( \bP ) \geq \kappa'$; $r_{\lambda_0} = \lambda_0 d$ for the
unit ball and $r_{\lambda_0} = \lambda_0 \Tr \bA^{-1}$ for the ellipsoid
$\bA^{1/2} \mathbb{B}_2^d(1)$ with its adapted norm), and
\begin{equation}
\label{eq:adapted-facts}
\begin{gathered}
\frac{\norm{\tilde{\theta}}_2}{D_{\ast} \ell_{\ast}} \leq \norm{\theta}_{\ast} \leq \frac{\norm{\tilde{\theta}}_2}{\ell_{\ast}} ,
\qquad
\frac{\norm{x}_{\ast}^{\circ}}{D_{\ast} \ell_{\ast}} \leq \norm{\tilde{x}}_2 \leq \frac{\norm{x}_{\ast}^{\circ}}{\ell_{\ast}} ,
\qquad
\norm{\tilde{x}^{\star}}_2 \geq \frac{\varrho_{\ast}}{D_{\ast} \ell_{\ast}} ,
\\[2pt]
\tilde{\lambda}_{t,d} \geq \frac{\omega_t^{-1}}{D_{\ast}^2 \ell_{\ast}^2} ,
\qquad \text{hence } \tilde{\lambda}_{t,d} \geq 2 \ \text{ at every round with } \omega_t^{-1} \geq 2 D_{\ast}^2 \ell_{\ast}^2 ,
\end{gathered}
\end{equation}
where $\tilde{\lambda}_{t,1} \geq \cdots \geq \tilde{\lambda}_{t,d} > 0$ are
the eigenvalues of $\tilde{V}_t$ with orthonormal eigenvectors
$\tilde{v}_{t,1}, \dots, \tilde{v}_{t,d}$; each eigenvalue is nondecreasing
in $t$ (because $\tilde{V}_{t+1} \succeq \tilde{V}_t$), Weyl's inequality
gives $\tilde{\lambda}_{t+1,i} \leq \tilde{\lambda}_{t,i} + 1$, and for a
vector $u$ of the adapted coordinates we write $u_i :=
\dotprod{u}{\tilde{v}_{t,i}}$ when the round is clear from context; in
particular $\tilde{x}_{t,i} := \dotprod{\tilde{x}_t}{\tilde{v}_{t,i}}$. We
also write $\tilde{e}^{\star} := \tilde{x}^{\star} / \norm{\tilde{x}^{\star}}_2$
for the unit vector of the optimal arm, $\tilde{P}^{\perp} := \I -
\tilde{e}^{\star} ( \tilde{e}^{\star} )^{\top}$ and $\tilde{\phi}_t :=
\angle( \tilde{v}_{t,1}, \tilde{e}^{\star} )$, the sign of $\tilde{v}_{t,1}$
being chosen so that $\cos \tilde{\phi}_t \geq 0$. The scalars $\omega_t$, $\norm{\theta - \theta'}_{\ast}$, $\norm{x - x'}_{\ast}^{\circ}$,
$\norm{\theta - \theta'}_{V_t} = \norm{\tilde{\theta} - \tilde{\theta}'}_{\tilde{V}_t}$,
$x^{\top} V_t^{-1} x = \tilde{x}^{\top} \tilde{V}_t^{-1} \tilde{x}$,
$J_{\cX}$, $m$, $M$, $\ell_{\ast}$, $\varrho_{\ast}$, $J_{\star}$,
$\alpha$, $\beta$ are the same in both descriptions
(\cref{lem:adapted-int}\ref{it:ad-equivariance}), so every statement about
them proved below is a statement about the original problem.
\Cref{prop:Yassin_conf} is applied in the original coordinates; the
event $\cE$ is invariant under the change of variables
(\cref{lem:adapted-int}\ref{it:ad-equivariance}), and only $\lP( \cE )
\geq 1 - \delta$ is used.

\subsection{Definition of the constants}
\phantomsection\label{sec:constant_def}

\begin{definition}[Constants]
\label{def:constants-int}
All constants of this section are fixed here once, in dependency
order. Two geometric inputs come first,
\begin{equation}
\label{eq:geometric-constants-int}
c_{\mathrm{eng}} := \frac{m}{6\, J_{\star}}
\qquad \text{and} \qquad
L_{\ast} := \frac{M + 2 \ell_{\ast}^2}{J_{\star}} ,
\end{equation}
the exploration-engine constant and the Lipschitz constant of the
argmax map $\theta \mapsto \nabla J_{\cX}( \theta )$ from $( \cN_{\ast},
\norm{\cdot}_{\ast} )$ to $( \cX, \norm{\cdot}_{\ast}^{\circ} )$
(\cref{prop:lip-int}); then
\begin{equation}
\label{eq:constants-int}
\begin{gathered}
c_1 := \frac{c_{\mathrm{eng}}\, \alpha}{2 \sqrt{d}},
\qquad
C_2 := \frac{3\, L_{\ast}^2\, \beta_{+}^2}{\ell_{\ast}^2\, c_1},
\qquad
\Gamma := \frac{5 \times 10^{4}\, d\, D_{\ast}^4\, L_{\ast}^4\, \beta_{+}^4\, \Log{T}}
{c_{\mathrm{eng}}^4\, \alpha^4},
\\[2pt]
K := \frac{40\, d\, \Log{8 \Gamma}}{c_{\mathrm{eng}}^2\, \alpha^2},
\qquad
c_0 := \min\Big( \frac{1}{4 \sqrt{K}},\
\frac{c_1}{\sqrt{\Log{9\, T_\star}}} \Big),
\end{gathered}
\end{equation}
where $T_\star$ is the $\operatorname{poly} (\log T)$ explicit burn-in threshold of \cref{def:ladder-int}
below.
\end{definition}

Two facts about $\Gamma$ are used repeatedly: since
\begin{equation}
\label{eq:LB-ratio-int}
\frac{L_{\ast} \beta_{+}}{c_{\mathrm{eng}} \alpha}
= \frac{6 ( M + 2 \ell_{\ast}^2 ) \beta_{+}}{m \alpha}
\;\geq\; 6
\end{equation}
(as $M \geq m$ and $\beta_{+} \geq \alpha$), $d \geq 2$,
$D_{\ast} \geq 1$ and $\Log{T} \geq 1$,
\begin{equation}
\label{eq:Gamma-facts-int}
\Gamma \;\geq\; \frac{80\, D_{\ast}^2\, L_{\ast}^2\, \beta_{+}^2}{c_{\mathrm{eng}}^2\, \alpha^2}
\qquad \text{and} \qquad
\Gamma \;\geq\; \frac{1280\, D_{\ast}^4\, \ell_{\ast}^4\, C_2^2\, \Log{T}}{c_{\mathrm{eng}}^2\, \alpha^2}
= \frac{46080\, d\, D_{\ast}^4\, L_{\ast}^4\, \beta_{+}^4\, \Log{T}}
{c_{\mathrm{eng}}^4\, \alpha^4} ,
\end{equation}
where $C_2^2 = 9 L_{\ast}^4 \beta_{+}^4 / ( \ell_{\ast}^4 c_1^2 ) = 36\,
d\, L_{\ast}^4 \beta_{+}^4 / ( \ell_{\ast}^4 c_{\mathrm{eng}}^2 \alpha^2 )$;
in particular $\Gamma \geq 1$. Note that $J_{\star}$ cancels in
$\Gamma$, that $\Gamma$ depends on the geometry only
through the invariant ratio~\eqref{eq:LB-ratio-int} and $D_{\ast}$, and
that by~\eqref{eq:Dstar} $4 \Log{D_{\ast}} \leq 2 \Log{d}$, so
$\Log{8 \Gamma} = O\big( \Log{e d} + \Log{e + ( M + 2 \ell_{\ast}^2 )
\beta_{+} / ( m \alpha )} + \Log{e \Log{e T}} \big)$; hence
\[
c_0 \;\asymp\; \frac{m\, \alpha}
{J_{\star} \sqrt{d}\, \sqrt{\Log{8 \Gamma} + \Log{9 T_\star}}} .
\]

\subsection{Exploration engine, burn-in and crude bound}

The first step of the analysis is a \emph{crude} bound on the width,
$\omega_t^{-1} \gtrsim \sqrt{t / \Log{t}}$ (\cref{lem:crude-int}),
obtained from an exploration inequality forced by optimism (the
\emph{engine}, \cref{lem:engine-int}) and an elliptic-potential
argument. The engine holds at every round, deterministically, with a
constant that involves $J_{\cX}( \hat{\theta}_t )$; on $\cE$ it forces
the width to decay from the first round on, which shows that the
threshold $\Lambda_0^{\ast}$ of the burn-in is reached at an explicit
round $t_0$ (\cref{prop:burnin-int}\ref{it:bint-width}). From $t_0$ on,
the estimate, the optimistic parameter and the whole set $\cC_t( \alpha )$ lie in $\cN_{\ast}$, where the argmax map is
$L_{\ast}$-Lipschitz (\cref{prop:lip-int}), the played arms are well
conditioned in the adapted coordinates
(\cref{prop:burnin-int}\ref{it:bint-hat}--\ref{it:bint-eucl}), and the
engine takes its final form with the constant $c_{\mathrm{eng}}$.

\begin{restatable}[Exploration engine]{restatablelemma}{lemEngineInt}
\phantomsection\label{lem:engine-int}
Recall $c_{\mathrm{eng}} = m / ( 6 J_{\star} )$ and $\beta_{+} = \alpha + \beta$.
\begin{enumerate}[label=(\roman*), itemsep=1pt]
\item \label{it:int-identity}
At every round $t \geq 1$, deterministically, $x_t \neq 0$, the
optimistic parameter has the closed form
\begin{equation}
\label{eq:stationarity-int}
\theta_t \;=\; \hat{\theta}_t + \alpha\, \frac{V_t^{-1} x_t}{\norm{x_t}_{V_t^{-1}}} ,
\end{equation}
and
\begin{equation}
\label{eq:engine-general-int}
\norm{x_t}_{V_t^{-1}}
\;\geq\; \frac{m\, \alpha\, \omega_t}
{4\, J_{\cX}( \hat{\theta}_t ) + 2\, \ell_{\ast}\, \alpha\, \omega_t^{1/2}} .
\end{equation}
\item \label{it:int-early}
On $\cE$, at every round $t \in [T]$,
\begin{equation}
\label{eq:engine-early-int}
x_t^{\top} V_t^{-1} x_t
\;\geq\; \frac{m^2 \alpha^2}{64}\,
\min\Big\{ \frac{\omega_t^2}{J_{\star}^2},\ \frac{\omega_t}{\ell_{\ast}^2 \beta_{+}^2} \Big\} .
\end{equation}
\item \label{it:int-engine}
On $\cE$, at every round $t \in [T]$ with $\ell_{\ast} \beta_{+}\,
\omega_t^{1/2} \leq J_{\star} / 2$ (the \emph{exploration engine}),
\begin{equation}
\label{eq:engine-int}
\norm{x_t}_{V_t^{-1}} \;\geq\; c_{\mathrm{eng}}\, \alpha\, \omega_t ,
\qquad \text{hence} \qquad
x_t^{\top} V_t^{-1} x_t \;\geq\; c_{\mathrm{eng}}^2\, \alpha^2\, \omega_t^2 .
\end{equation}
\end{enumerate}
The statements and their proof involve neither the burn-in nor a
Euclidean structure.
\end{restatable}

\begin{proof}[\cref{lem:engine-int}]
\ref{it:int-identity} \emph{Reduction to the confidence set.} By
definition of the optimistic pair and \cref{prop:global-int}\ref{it:gl-round},
\[
\dotprod{\theta_t}{x_t}
= \max_{\theta \in \cC_t( \alpha )} \max_{x \in \cX} \dotprod{\theta}{x}
= \max_{\theta \in \cC_t( \alpha )} J_{\cX}( \theta ) ,
\qquad
x_t = \nabla J_{\cX}( \theta_t ) ,
\qquad
\dotprod{\theta_t}{x_t} = J_{\cX}( \theta_t ) > 0 ,
\]
so $\theta_t \in \argmax_{\theta \in \cC_t( \alpha )} J_{\cX}( \theta )$
and $x_t \neq 0$.

\emph{Stationarity identity.} For $x \neq 0$ and $\theta \in \cC_t( \alpha )$,
the Cauchy--Schwarz inequality in the $V_t$-geometry gives
$\dotprod{\theta - \hat{\theta}_t}{x} \leq \norm{x}_{V_t^{-1}}
\norm{\theta - \hat{\theta}_t}_{V_t} \leq \alpha\, \norm{x}_{V_t^{-1}}$,
with equality if and only if $\theta = \hat{\theta}_t + \alpha\,
V_t^{-1} x / \norm{x}_{V_t^{-1}}$. Since $\theta_t$ maximises
$\dotprod{\cdot}{x_t}$ over $\cC_t( \alpha )$, it is this point:
\begin{equation}
\label{eq:stationarity-proof-int}
\theta_t - \hat{\theta}_t = \frac{\alpha}{\norm{x_t}_{V_t^{-1}}}
V_t^{-1} x_t ,
\qquad
\norm{\theta_t - \hat{\theta}_t}_{V_t} = \alpha ,
\end{equation}
which is the closed form in~\ref{it:int-identity}. It also gives the
\emph{optimism inequality}
\begin{equation}
\label{eq:optimism-int}
J_{\cX}( \theta_t ) = \dotprod{\hat{\theta}_t}{x_t}
+ \dotprod{\theta_t - \hat{\theta}_t}{x_t}
\;\leq\; J_{\cX}( \hat{\theta}_t ) + \alpha\, \norm{x_t}_{V_t^{-1}} .
\end{equation}

\emph{The multiplier, via strong convexity of $J_{\cX}^2$.} Let
$v^{\circ}$ attain the maximum defining $\omega_t$ in
\cref{prop:width-int}\ref{it:wint-dual}, so $\norm{v^{\circ}}_{V_t} \leq 1$
and $\norm{v^{\circ}}_{\ast} = \omega_t^{1/2}$, and set $\zeta := \alpha\, v^{\circ}$, so that
\[
\norm{\zeta}_{V_t} \leq \alpha ,
\qquad
\norm{\zeta}_{\ast} = \alpha\, \omega_t^{1/2} :
\]
the two points $\hat{\theta}_t \pm \zeta$ lie in $\cC_t( \alpha )$,
in the least-explored direction of the $\norm{\cdot}_{\ast}$ geometry.
Since $\theta_t$ maximises $J_{\cX}$ over $\cC_t( \alpha )$,
\eqref{eq:optimism-int} gives
\begin{equation}
\label{eq:G-int}
\alpha\, \norm{x_t}_{V_t^{-1}}
\;\geq\; G := \max\big\{ J_{\cX}( \hat{\theta}_t + \zeta ),\,
J_{\cX}( \hat{\theta}_t - \zeta ) \big\} - J_{\cX}( \hat{\theta}_t )
\;\geq\; 0 ,
\end{equation}
the last inequality by convexity ($G \geq \tfrac12 [ J_{\cX}(
\hat{\theta}_t + \zeta ) + J_{\cX}( \hat{\theta}_t - \zeta ) ]
- J_{\cX}( \hat{\theta}_t )$). Now apply the lower bound
in~\eqref{eq:ajp-int} at $\theta' = \hat{\theta}_t$ with $\theta =
\hat{\theta}_t \pm \zeta$ and add: the linear terms $\pm
\dotprod{\zeta}{\nabla J_{\cX}^2( \hat{\theta}_t )}$ cancel, and
\begin{equation}
\label{eq:E-sum-int}
J_{\cX}^2( \hat{\theta}_t + \zeta ) + J_{\cX}^2( \hat{\theta}_t - \zeta )
- 2 J_{\cX}^2( \hat{\theta}_t )
= D_{J_{\cX}^2}( \hat{\theta}_t + \zeta, \hat{\theta}_t )
+ D_{J_{\cX}^2}( \hat{\theta}_t - \zeta, \hat{\theta}_t )
\;\geq\; m \norm{\zeta}_{\ast}^2
= m\, \alpha^2\, \omega_t .
\end{equation}
On the other hand, for each sign,
\[
J_{\cX}^2( \hat{\theta}_t \pm \zeta ) - J_{\cX}^2( \hat{\theta}_t )
= \big[ J_{\cX}( \hat{\theta}_t \pm \zeta ) - J_{\cX}( \hat{\theta}_t ) \big]
\big[ J_{\cX}( \hat{\theta}_t \pm \zeta ) + J_{\cX}( \hat{\theta}_t ) \big]
\;\leq\; G\, \big( 2 J_{\cX}( \hat{\theta}_t ) + G \big) ,
\]
because the product is nonpositive when the first factor is, and
otherwise the first factor is at most $G$ and the second at most $2
J_{\cX}( \hat{\theta}_t ) + G$. Summing over the two signs and comparing
with~\eqref{eq:E-sum-int}, $2 G\, ( 2 J_{\cX}( \hat{\theta}_t ) + G ) \geq
m \alpha^2\, \omega_t$. Finally $G \leq \alpha\,
\norm{x_t}_{V_t^{-1}} \leq \ell_{\ast} \alpha\, \omega_t^{1/2}$
by~\eqref{eq:G-int} and \cref{prop:width-int}\ref{it:wint-energy}, so
\begin{equation}
\label{eq:x-energy-int}
\norm{x_t}_{V_t^{-1}}
\;\geq\; \frac{G}{\alpha}
\;\geq\; \frac{m\, \alpha\, \omega_t}{4 J_{\cX}( \hat{\theta}_t ) + 2 G}
\;\geq\; \frac{m\, \alpha\, \omega_t}
{4 J_{\cX}( \hat{\theta}_t ) + 2 \ell_{\ast} \alpha\, \omega_t^{1/2}} ,
\end{equation}
which is~\eqref{eq:engine-general-int}. The perturbation is chosen in
the $\norm{\cdot}_{\ast}$ geometry and strong convexity is used in that
same geometry. (The radial direction, along which $J_{\cX}$ is linear,
needs no separate treatment: it is controlled by the rank-one term $2
\nabla J_{\cX} \nabla J_{\cX}^{\top}$ of $\nabla^2 J_{\cX}^2$, which is
what strong convexity of the \emph{square} buys.)

\ref{it:int-early} On $\cE$, $\norm{\hat{\theta}_t - \theta^{\star}}_{\ast}
\leq \beta\, \omega_t^{1/2}$ (\cref{prop:width-int}\ref{it:wint-radius}),
so the $\ell_{\ast}$-Lipschitz property of $J_{\cX}$ gives $J_{\cX}(
\hat{\theta}_t ) \leq J_{\star} + \ell_{\ast} \beta\, \omega_t^{1/2}$, and
the denominator in~\eqref{eq:engine-general-int} is at most
\[
4 J_{\star} + 4 \ell_{\ast} \beta_{+}\, \omega_t^{1/2}
\;\leq\; 8 \max\{ J_{\star},\ \ell_{\ast} \beta_{+}\, \omega_t^{1/2} \} .
\]
Hence $\norm{x_t}_{V_t^{-1}} \geq \frac{m \alpha}{8} \min\{
\omega_t / J_{\star},\ \omega_t^{1/2} / ( \ell_{\ast} \beta_{+} ) \}$, and
\eqref{eq:engine-early-int} follows by squaring.

\ref{it:int-engine} If $\ell_{\ast} \beta_{+} \omega_t^{1/2} \leq J_{\star}
/ 2$, the same denominator is at most $6 J_{\star}$, and
\eqref{eq:engine-general-int} gives $\norm{x_t}_{V_t^{-1}} \geq m \alpha\,
\omega_t / ( 6 J_{\star} ) = c_{\mathrm{eng}}\, \alpha\, \omega_t$.
\end{proof}

\begin{restatable}[Burn-in]{restatableproposition}{propBurninInt}
\phantomsection\label{prop:burnin-int}%
\label{def:burnin-int}\label{lem:burnin-int}\label{lem:burnin-occurs-int}%
Recall $\beta_{+} = \alpha + \beta$ and $L_{\ast} = ( M + 2
\ell_{\ast}^2 ) / J_{\star}$, and define the \emph{burn-in threshold}
and the \emph{burn-in round}
\begin{equation}
\label{eq:burnin-int}
\begin{gathered}
\Lambda_0^{\ast} := \max\Big\{ 2 D_{\ast}^2 \ell_{\ast}^2,\
\frac{16\, \ell_{\ast}^2 \beta_{+}^2}{J_{\star}^2},\
\frac{4\, D_{\ast}^2 L_{\ast}^2 \beta_{+}^2}{\varrho_{\ast}^2} \Big\} ,
\\[2pt]
t_0 := \min\Big\{ t \geq 1 :
\frac{t}{\Log{ 1 + t / ( d \lambda_0 ) }}
> \max\Big\{ 2 d,\ \frac{128\, d\, J_{\star}^2\, \Lambda_0^{\ast\,2}}{m^2 \alpha^2} \Big\}
\Big\} .
\end{gathered}
\end{equation}
The round $t_0$ is finite and explicit, since $t \mapsto t / \Log{ 1 + t
/ ( d \lambda_0 ) }$ increases to infinity. On $\cE$, for every $t_0
\leq t \leq T$:
\begin{enumerate}[label=(\roman*), itemsep=1pt]
\item \label{it:bint-width}
$\omega_t^{-1} \geq \Lambda_0^{\ast}$;
\item \label{it:bint-hat}
$\norm{\hat{\theta}_t - \theta^{\star}}_{\ast} \leq \beta\, \omega_t^{1/2}
\leq J_{\star} / ( 4 \ell_{\ast} )$; in particular $\hat{\theta}_t \in
\cN_{\ast}$;
\item \label{it:bint-theta}
$\norm{\theta_t - \hat{\theta}_t}_{\ast} \leq \alpha\, \omega_t^{1/2} \leq
J_{\star} / ( 4 \ell_{\ast} )$ and $\norm{\theta_t - \theta^{\star}}_{\ast}
\leq \beta_{+}\, \omega_t^{1/2} \leq J_{\star} / ( 4 \ell_{\ast} )$; hence
$\theta_t \in \cN_{\ast}$, and $\hat{\theta}_t + \zeta \in \cN_{\ast}$
for every $\zeta$ with $\norm{\zeta}_{V_t} \leq \alpha$ (the whole
confidence ellipsoid lies in $\cN_{\ast}$);
\item \label{it:bint-arm}
$x_t = \nabla J_{\cX}( \theta_t )$, $\ \norm{x_t - x^{\star}}_{\ast}^{\circ}
\leq L_{\ast} \beta_{+}\, \omega_t^{1/2}$, and $\tfrac12 J_{\star} \leq
J_{\cX}( \theta_t ) \leq \tfrac32 J_{\star}$;
\item \label{it:bint-eucl}
in the adapted coordinates, $\norm{\tilde{x}_t - \tilde{x}^{\star}}_2 \leq L_{\ast}
\beta_{+}\, \omega_t^{1/2} / \ell_{\ast} \leq \norm{\tilde{x}^{\star}}_2 / 2$;
therefore
\begin{equation}
\label{eq:arm-sandwich-int}
\tfrac12 \norm{\tilde{x}^{\star}}_2 \leq \norm{\tilde{x}_t}_2 \leq \tfrac32 \norm{\tilde{x}^{\star}}_2
\qquad \text{and} \qquad
\dotprod{\tilde{e}^{\star}}{\tilde{x}_t} \geq \tfrac12 \norm{\tilde{x}^{\star}}_2 .
\end{equation}
\end{enumerate}
Nothing is assumed about the trajectory: \ref{it:bint-width} is derived
from the engine~\eqref{eq:engine-early-int} and an elliptic-potential
argument run from the first round, and
\ref{it:bint-hat}--\ref{it:bint-eucl} follow from \ref{it:bint-width}.
\end{restatable}

\begin{proof}[\cref{prop:burnin-int}]
\ref{it:bint-width} Fix $T' \in [T]$. In the original coordinates $V_1 = \lambda_0 \I$ and
$\norm{x_t}_2 \leq 1$ (as $\cX \subseteq \mathbb{B}_2^d(1)$), so by the
matrix determinant lemma, telescoping, and the arithmetic--geometric
mean inequality ($\Tr V_{T'+1} \leq d \lambda_0 + T'$),
\[
\sum_{t = 1}^{T'} \Log{ 1 + x_t^{\top} V_t^{-1} x_t }
= \Log{ \frac{\Det{ V_{T'+1} }}{\Det{ V_1 }} }
\;\leq\; d\, \Log{ 1 + \frac{T'}{d \lambda_0} } .
\]
Since $\min\{ 1, y \} \leq 2 \Log{ 1 + y }$ for $y \geq 0$, the left
side is at least $\frac12 \sum_{t \leq T'} \min\{ 1, x_t^{\top} V_t^{-1}
x_t \}$. On $\cE$, insert~\eqref{eq:engine-early-int}: its right-hand
side is $g( \omega_t )$ with $g( \omega ) := \frac{m^2 \alpha^2}{64}
\min\{ \omega^2 / J_{\star}^2,\ \omega / ( \ell_{\ast}^2 \beta_{+}^2 ) \}$
nondecreasing in $\omega$, while $\omega_t \geq \omega_{T'}$ for $t \leq
T'$ (\cref{prop:width-int}\ref{it:wint-monotone}); hence
\[
T' \min\{ 1, g( \omega_{T'} ) \}
\;\leq\; \sum_{t = 1}^{T'} \min\{ 1, x_t^{\top} V_t^{-1} x_t \}
\;\leq\; 2 d\, \Log{ 1 + \frac{T'}{d \lambda_0} } .
\]
Let $T' \geq t_0$. Since $t \mapsto t / \Log{ 1 + t / ( d \lambda_0 ) }$
is increasing, the defining inequality of $t_0$ holds at $T'$: $T' > 2 d
\Log{ 1 + T' / ( d \lambda_0 ) }$, so the minimum is $g( \omega_{T'} )
< 1$, and
\[
g( \omega_{T'} )
\;\leq\; \frac{2 d\, \Log{ 1 + T' / ( d \lambda_0 ) }}{T'}
\;<\; \frac{m^2 \alpha^2}{64\, J_{\star}^2 \Lambda_0^{\ast 2}} .
\]
If $\omega_{T'} > 1 / \Lambda_0^{\ast}$, then by monotonicity of $g$ and
the second entry of $\Lambda_0^{\ast}$ ($\ell_{\ast}^2 \beta_{+}^2 \leq
J_{\star}^2 \Lambda_0^{\ast}$, so that at $\omega = 1 / \Lambda_0^{\ast}$
the first term of the minimum is the smaller one),
\[
g( \omega_{T'} ) \;\geq\; g( 1 / \Lambda_0^{\ast} )
= \frac{m^2 \alpha^2}{64}\,
\min\Big\{ \frac{1}{J_{\star}^2 \Lambda_0^{\ast 2}},\
\frac{1}{\ell_{\ast}^2 \beta_{+}^2 \Lambda_0^{\ast}} \Big\}
= \frac{m^2 \alpha^2}{64\, J_{\star}^2 \Lambda_0^{\ast 2}} ,
\]
a contradiction. Hence $\omega_{T'} \leq 1 / \Lambda_0^{\ast}$.

For \ref{it:bint-hat}--\ref{it:bint-eucl}, fix $t_0 \leq t \leq T$;
throughout, $\omega_t^{-1} \geq \Lambda_0^{\ast}$ by \ref{it:bint-width},
and $\alpha, \beta \leq \beta_{+}$.
\ref{it:bint-hat} By \cref{prop:width-int}\ref{it:wint-radius}, on
$\cE$, $\norm{\hat{\theta}_t - \theta^{\star}}_{\ast} \leq \beta\,
\omega_t^{1/2} \leq \beta_{+} / \sqrt{\Lambda_0^{\ast}} \leq
J_{\star} / ( 4 \ell_{\ast} )$ by the second entry of $\Lambda_0^{\ast}$.
\ref{it:bint-theta} $\theta_t \in \cC_t( \alpha )$ gives $\norm{\theta_t -
\hat{\theta}_t}_{\ast} \leq \alpha\, \omega_t^{1/2} \leq J_{\star} / ( 4 \ell_{\ast} )$
in the same way; the triangle inequality gives $\norm{\theta_t -
\theta^{\star}}_{\ast} \leq ( \alpha + \beta )\, \omega_t^{1/2} =
\beta_{+}\, \omega_t^{1/2} \leq J_{\star} / ( 4 \ell_{\ast} )$; and if $\norm{\zeta}_{V_t}
\leq \alpha$ then $\norm{\zeta}_{\ast} \leq \alpha\, \omega_t^{1/2} \leq J_{\star} /
( 4 \ell_{\ast} )$, so $\norm{\hat{\theta}_t + \zeta - \theta^{\star}}_{\ast}
\leq J_{\star} / ( 2 \ell_{\ast} )$.
\ref{it:bint-arm} $x_t = \nabla J_{\cX}( \theta_t )$ is
\cref{prop:global-int}\ref{it:gl-round}. By~\eqref{eq:lip-int} with
$\theta_t, \theta^{\star} \in \cN_{\ast}$ and by \ref{it:bint-theta},
$\norm{x_t - x^{\star}}_{\ast}^{\circ} \leq L_{\ast} \norm{\theta_t -
\theta^{\star}}_{\ast} \leq L_{\ast} \beta_{+}\, \omega_t^{1/2}$; the
bounds on $J_{\cX}( \theta_t )$ are~\eqref{eq:N-int}.
\ref{it:bint-eucl} By~\eqref{eq:adapted-facts}, $\norm{\tilde{x}_t - \tilde{x}^{\star}}_2
\leq \norm{x_t - x^{\star}}_{\ast}^{\circ} / \ell_{\ast} \leq L_{\ast}
\beta_{+}\, \omega_t^{1/2} / \ell_{\ast} \leq L_{\ast} \beta_{+} /
( \ell_{\ast} \sqrt{\Lambda_0^{\ast}} ) \leq \varrho_{\ast} / ( 2
D_{\ast} \ell_{\ast} ) \leq \norm{\tilde{x}^{\star}}_2 / 2$, using the third
entry of $\Lambda_0^{\ast}$ and $\norm{\tilde{x}^{\star}}_2 \geq \varrho_{\ast}
/ ( D_{\ast} \ell_{\ast} )$. The bounds~\eqref{eq:arm-sandwich-int}
follow from the triangle inequality and from $\dotprod{\tilde{e}^{\star}}{\tilde{x}_t}
= \norm{\tilde{x}^{\star}}_2 + \dotprod{\tilde{e}^{\star}}{\tilde{x}_t - \tilde{x}^{\star}} \geq
\norm{\tilde{x}^{\star}}_2 - \norm{\tilde{x}_t - \tilde{x}^{\star}}_2$.
\end{proof}

\paragraph{Size of the burn-in.}
By \eqref{eq:scalars-upperbound}
and~\eqref{eq:Dstar} ($\ell_{\ast}^2 \leq M / 2$, $\varrho_{\ast}^{-2} \leq
2 / m$, $D_{\ast}^2 \leq M / m$, $L_{\ast} \leq 2 M / J_{\star}$),
\[
\Lambda_0^{\ast} \;\leq\; \max\Big\{ \frac{M^2}{m},\
\frac{8 M \beta_{+}^2}{J_{\star}^2},\
\frac{32 M^3 \beta_{+}^2}{m^2 J_{\star}^2} \Big\} ;
\]
for the unit ball and for every ellipsoid with its adapted norm,
$\Lambda_0^{\ast} = \max\{ 2,\, 64 \beta_{+}^2 / J_{\star}^2 \}$, whatever
the anisotropy. Consequently $t_0 = \widetilde{O}\big( d
+ d\, J_{\star}^2 \Lambda_0^{\ast 2} / ( m^2 \alpha^2 ) \big)$, i.e.
\[
t_0 \;=\; \widetilde{O}\Big( d + \frac{d\, M^4 J_{\star}^2}{m^4 \alpha^2}
+ \frac{d\, M^6 \beta_{+}^4}{m^6 J_{\star}^2 \alpha^2} \Big) ,
\]
with a logarithmic dependence on $1 / ( d \lambda_0 )$.

\begin{restatable}[Crude bound]{restatablelemma}{lemCrudeInt}
\phantomsection\label{lem:crude-int}
Recall $c_1 = c_{\mathrm{eng}} \alpha / ( 2 \sqrt{d} ) = m \alpha / ( 12 J_{\star} \sqrt{d} )$,
and let $T_1 := \max( 2 t_0,\, 3 d )$. On $\cE$, the
engine~\eqref{eq:engine-int} holds at every round $t_0 \leq t \leq T$,
and
\begin{equation}
\label{eq:crude-int}
\omega_t^{-1} \;\geq\; c_1 \sqrt{\frac{t}{\Log{t}}}
\qquad \text{for all } T_1 \leq t \leq T .
\end{equation}
\end{restatable}

\begin{proof}[\cref{lem:crude-int}]
For $t \geq t_0$, $\ell_{\ast} \beta_{+}\, \omega_t^{1/2} \leq \ell_{\ast}
\beta_{+} / \sqrt{\Lambda_0^{\ast}} \leq J_{\star} / 4$ by
\cref{prop:burnin-int}\ref{it:bint-width} and the second entry of $\Lambda_0^{\ast}$,
so \cref{lem:engine-int}\ref{it:int-engine} applies at every $t \geq
t_0$. For the crude bound, an elliptic-potential argument converts
\eqref{eq:engine-int} into~\eqref{eq:crude-int}. We use the
adapted coordinates. By the matrix determinant lemma, $\Det{ \tilde{V}_{t+1}
} = \Det{ \tilde{V}_t } \big( 1 + x_t^{\top} V_t^{-1} x_t \big)$, so
telescoping over $t = t_0, \dots, T'$ for any $t_0 \leq T' \leq T$,
\[
\sum_{t = t_0}^{T'} \Log{ 1 + x_t^{\top} V_t^{-1} x_t }
= \Log{ \frac{\Det{ \tilde{V}_{T'+1} }}{\Det{ \tilde{V}_{t_0} }} }
\;\leq\; d\, \Log{T'} ,
\]
since $\tilde{V}_{T'+1} = \tilde{V}_{t_0} + \sum_{t = t_0}^{T'}
\tilde{x}_t \tilde{x}_t^{\top} \preceq \tilde{V}_{t_0} + ( T' - t_0 + 1 )
\I$ (as $\norm{\tilde{x}_s}_2 \leq 1$) and $\tilde{\lambda}_{t_0,d} \geq 2$
by~\eqref{eq:adapted-facts} and \cref{prop:burnin-int}\ref{it:bint-width}, so that
$\Det{ \tilde{V}_{T'+1} } / \Det{ \tilde{V}_{t_0} } = \Det{ \I +
\tilde{V}_{t_0}^{-1/2} ( \tilde{V}_{T'+1} - \tilde{V}_{t_0} )
\tilde{V}_{t_0}^{-1/2} } \leq ( 1 + T' / 2 )^d \leq T'^{\,d}$ for $T'
\geq 2$. For $t \geq t_0$, $x_t^{\top} V_t^{-1} x_t
\leq 1 / 2$ (\cref{lem:adapted-int}\ref{it:ad-floor}, as $\omega_t^{-1} \geq
\Lambda_0^{\ast} \geq 2 D_{\ast}^2 \ell_{\ast}^2$), so the elementary inequality
$\log( 1 + y ) \geq y / 2$ on $[ 0, 1 ]$ applies; inserting the
engine~\eqref{eq:engine-int} (\cref{lem:engine-int}\ref{it:int-engine},
whose hypothesis $\ell_{\ast} \beta_{+} \omega_t^{1/2} \leq J_{\star} / 4$
holds for $t \geq t_0$ by the second entry of $\Lambda_0^{\ast}$) and the
monotonicity of $\omega_{\cdot}$,
\[
( T' - t_0 + 1 )\, \frac{c_{\mathrm{eng}}^2 \alpha^2\, \omega_{T'}^2}{2}
\;\leq\; \frac12 \sum_{t = t_0}^{T'} x_t^{\top} V_t^{-1} x_t
\;\leq\; d\, \Log{T'} .
\]
For $T' \geq T_1 = \max( 2 t_0, 3 d )$ we have $T' - t_0 + 1 \geq
T' / 2$, whence
$\omega_{T'}^{-2} \geq c_{\mathrm{eng}}^2 \alpha^2\, T' / ( 4 d \Log{T'} )
= c_1^2\, T' / \Log{T'}$; renaming $T'$ as $t$ gives~\eqref{eq:crude-int}.
\end{proof}

\subsection*{The burn-in ladder}
 
All burn-in thresholds are collected in one explicit chain
$t_0 (in~\cref{prop:burnin-int}) \to T_1 := \max( 2 t_0,\, 3 d ) \to T_\star \to T_B$.
 
\begin{definition}[Burn-in ladder]
\phantomsection\label{def:ladder-int}
With $T_1 = \max( 2 t_0, 3 d )$, $c_1 = m \alpha / ( 12 J_{\star} \sqrt{d} )$
and $C_2 = 3 L_{\ast}^2 \beta_{+}^2 / ( \ell_{\ast}^2 c_1 )$ as
in~\eqref{eq:constants-int}, define
\begin{align*}
T_\star
&:= \min\Big\{\, t \geq \max( T_1,\, 20 ) \;:\; \forall\, s \geq t, \\
&\qquad\quad
4 T_1 + r_{\lambda_0} \leq \frac{L_{\ast}^2\, \beta_{+}^2}{\ell_{\ast}^2\, c_1} \sqrt{s \Log{s}}
\ \ \text{and} \ \
C_2 \sqrt{s \Log{s}} \leq \frac{\varrho_{\ast}^2\, s}{16\, D_{\ast}^2 \ell_{\ast}^2}
\,\Big\} ,
\\
T_B &:= T_\star .
\end{align*}
The first inequality makes the pre-$T_1$ chords and the regulariser
negligible in \hyperref[fact:P1]{(P1)} below; the second is the spectral gap of \hyperref[fact:P2]{(P2)}. The
threshold is finite and explicit: writing
\[
A := \frac{256\, D_{\ast}^4 \ell_{\ast}^4\, C_2^2}{\varrho_{\ast}^4},
\qquad
B := \frac{( 4 T_1 + r_{\lambda_0} )^2\, c_1^2\, \ell_{\ast}^4}{L_{\ast}^4\, \beta_{+}^4}
= \frac{( 4 T_1 + r_{\lambda_0} )^2\, c_{\mathrm{eng}}^2\, \alpha^2\, \ell_{\ast}^4}{4\, d\, L_{\ast}^4\, \beta_{+}^4} ,
\]
the second inequality reads $s / \Log{s} \geq A$ and the first $s \Log{s}
\geq B$; both, once true, stay true, and they hold for every $s \geq
\max\{ 3 A' \Log{A'},\ B,\ 3 \}$ with $A' := \max\{ A, 3 \}$.
Hence
\begin{equation}
\label{eq:Tstar-order-int}
T_\star \;\leq\; \max\{ T_1,\ 20,\ 3 A' \Log{A'},\ B \}
\end{equation}

\emph{Dependence on the horizon.} $T_\star$ depends on $T$ only through
$\beta_{+} = \alpha + \beta$. The terms
of~\eqref{eq:Tstar-order-int} scales as $\, \beta_{+}^4 / \alpha^2$ times a
polynomial in $( d, M / m, J_{\star}^{-1} )$.
Hence, up to such polynomial factors,
\[
T_B = T_\star = \cO(\operatorname{poly}(\log T)) .
\]
\end{definition}

\subsection[Proof of~\texorpdfstring{\Cref{thm:spectral_growth}}{Theorem spectral growth}]%
  {Proof of~\Cref{thm:spectral_growth}}
\label{sec:lower-bound-int}

The engine~\eqref{eq:engine-int} is a lower bound on $x_t^{\top}
V_t^{-1} x_t$, while the crude bound~\eqref{eq:crude-int} is an
upper bound on the sum of these quantities. To go from
$\sqrt{t / \Log{t}}$ to $\sqrt{t}$ we replace the elliptic potential by
a logarithmically capped potential whose capacity does not grow with
$t$, and record how much of the engine's energy it captures. The
structure of the argument is a two-scale separation: a
\emph{doubling lemma} shows that $\omega_t^{-1}$ doubles within $O( \vartheta^2 )$
rounds of the moment it equals $\vartheta$, and a backward induction turns
this into $\omega_t^{-1} \geq c_0 \sqrt{t}$ from the initialisation
window onwards.

\subsection*{Epochs, band, and the capped potential}

Epochs are measured on the inverse width $\omega_t^{-1}$, while the
potential that records the exploration is a spectral function of
$\tilde{V}$ in the adapted coordinates. Fix an \emph{epoch} start $\tau \geq
T_\star$ with reference scale $\vartheta := \omega_\tau^{-1}$; a round $t \geq
\tau$ is \emph{active} if $\omega_t^{-1} \leq 2 \vartheta$ (active rounds form an
initial segment, by monotonicity of $\omega_{\cdot}$, and $\omega_t^{-1}
\geq \vartheta$ for all $t \geq \tau$). Set
\begin{equation}
\label{eq:mbar-int}
\bar{\vartheta} := \frac{\vartheta}{D_{\ast}^2 \ell_{\ast}^2} ,
\qquad \text{so that} \qquad
0 < \bar{\vartheta} \leq \tilde{\lambda}_{t,d} \quad \text{for all } t \geq \tau ,
\end{equation}
because $\tilde{\lambda}_{t,d} \geq \tilde{\lambda}_{\tau,d} \geq \omega_\tau^{-1} / ( D_{\ast}^2
\ell_{\ast}^2 )$ by~\eqref{eq:adapted-facts}; $\bar{\vartheta}$ is the Euclidean
shadow of the epoch scale, and it will cancel from every constant. With
$\Gamma \geq 1$ the band width of~\eqref{eq:constants-int}, define
the \emph{band}, the \emph{mid-scale set}, and the band energy
\[
\cB_t := \{\, i \geq 2 : \tilde{\lambda}_{t,i} < \Gamma \bar{\vartheta} \,\},
\qquad
\cJ_t := \{\, i \geq 2 : \tilde{\lambda}_{t,i} \geq \Gamma \bar{\vartheta} \,\},
\qquad
E_t^{\cB} := \sum_{i \in \cB_t} \frac{\tilde{x}_{t,i}^2}{\tilde{\lambda}_{t,i}} ,
\]
and, with $L := 8 \Gamma \bar{\vartheta}$ (not to be confused with the
Lipschitz constant $L_{\ast}$), the \emph{logarithmically capped
potential}
\begin{equation}
\label{eq:potential-int}
\Phi_{\bar{\vartheta}}( \mathbf{W} ) := \Tr h( \mathbf{W} ) = \sum_{i = 1}^{d} h( \lambda_i( \mathbf{W} ) ) ,
\qquad
h( \lambda ) := \bar{\vartheta} \Log{ \frac{L\, ( \lambda + \bar{\vartheta} )}{\bar{\vartheta}\, ( \lambda + L )} } ,
\qquad \mathbf{W} \succeq 0 ,
\end{equation}
where $\mathbf{W}$ is any symmetric matrix of the adapted coordinates (it
will be $\tilde{V}_t$).
The function $h$ increases from $h( 0 ) = 0$ to $h( \infty ) = \bar{\vartheta}
\Log{ L / \bar{\vartheta} } = \bar{\vartheta} \Log{ 8 \Gamma }$: it counts
$\bar{\vartheta}$-units of growth on the logarithmic scale, but only between
$\bar{\vartheta}$ and $L$, so that its total capacity is bounded independently
of the horizon.

\begin{restatable}[Capacity and per-round gain of the capped potential]{restatablelemma}{lemPotentialInt}
\phantomsection\label{lem:potential-int}
Let $\tau \geq T_\star$, $\vartheta := \omega_\tau^{-1}$, $\bar{\vartheta} = \vartheta / ( D_{\ast}^2
\ell_{\ast}^2 )$ and $L = 8 \Gamma \bar{\vartheta}$ be as above.
\begin{enumerate}[label=(\roman*), itemsep=1pt]
\item \label{it:potint-capacity}
For every $\mathbf{W} \succeq 0$, $\ 0 \leq \Phi_{\bar{\vartheta}}( \mathbf{W} ) \leq d\, \bar{\vartheta}
\Log{8 \Gamma}$.
\item \label{it:potint-gain}
For every $t \geq \tau$,
$\ \Phi_{\bar{\vartheta}}( \tilde{V}_{t+1} ) - \Phi_{\bar{\vartheta}}( \tilde{V}_t ) \geq
\frac{\bar{\vartheta}}{4}\, E_t^{\cB} \geq 0$.
\end{enumerate}
\end{restatable}

\begin{proof}[\cref{lem:potential-int}]
\emph{Step 0: a determinant representation.} Since
\[
h( \lambda ) = \bar{\vartheta} \big[ \Log{ \lambda + \bar{\vartheta} } - \Log{ \lambda + L } \big]
+ \bar{\vartheta} \Log{ L / \bar{\vartheta} } ,
\]
we have, for $\mathbf{W} \succeq 0$,
\[
\Phi_{\bar{\vartheta}}( \mathbf{W} )
= \bar{\vartheta} \Big[ \Log{ \Det{ \mathbf{W} + \bar{\vartheta} \I } }
- \Log{ \Det{ \mathbf{W} + L \I } } \Big] + d\, \bar{\vartheta} \Log{ L / \bar{\vartheta} } .
\]
\paragraph{Proof of~\ref{it:potint-capacity}.} $h$ is nondecreasing on
$[ 0, \infty )$ with $h( 0 ) = 0$ (because $\bar{\vartheta} < L$), and $h( \lambda )
\leq \bar{\vartheta} \Log{ L / \bar{\vartheta} } = \bar{\vartheta} \Log{ 8 \Gamma }$ for
every $\lambda \geq 0$; sum over the $d$ eigenvalues.

\paragraph{Proof of~\ref{it:potint-gain}.} Fix $t \geq \tau$ and
abbreviate $\tilde{V} := \tilde{V}_t$, $\tilde{x} := \tilde{x}_t$,
$\tilde{\lambda}_i := \tilde{\lambda}_{t,i}$, $\tilde{x}_i := \tilde{x}_{t,i}$,
so that $\tilde{V}_{t+1} = \tilde{V} + \tilde{x} \tilde{x}^{\top}$;
recall $\bar{\vartheta} \leq \tilde{\lambda}_d$ by~\eqref{eq:mbar-int} and
$\tilde{\lambda}_d \geq 2$ by~\eqref{eq:adapted-facts}, since $\omega_t^{-1} \geq
\Lambda_0^{\ast} \geq 2 D_{\ast}^2 \ell_{\ast}^2$ for $t \geq t_0$
(\cref{prop:burnin-int}\ref{it:bint-width}). Set
\[
a := \tilde{x}^{\top} ( \tilde{V} + \bar{\vartheta} \I )^{-1} \tilde{x} = \sum_{i = 1}^{d} \frac{\tilde{x}_i^2}{\tilde{\lambda}_i + \bar{\vartheta}} ,
\qquad
b := \tilde{x}^{\top} ( \tilde{V} + L \I )^{-1} \tilde{x} = \sum_{i = 1}^{d} \frac{\tilde{x}_i^2}{\tilde{\lambda}_i + L} .
\]
\emph{(a) Determinant lemma.} By the matrix determinant lemma applied
to $\tilde{V} + \bar{\vartheta} \I$ and to $\tilde{V} + L \I$, the representation of
Step~0 yields $\Phi_{\bar{\vartheta}}( \tilde{V} + \tilde{x} \tilde{x}^{\top} ) - \Phi_{\bar{\vartheta}}( \tilde{V} )
= \bar{\vartheta} \Log{ ( 1 + a ) / ( 1 + b ) }$, with $0 \leq b \leq a$
(because $L > \bar{\vartheta}$) and $a \leq \tilde{x}^{\top} \tilde{V}^{-1} \tilde{x} \leq 1 /
\tilde{\lambda}_d \leq 1 / 2$, since $( \tilde{V} + \bar{\vartheta} \I )^{-1} \preceq
\tilde{V}^{-1}$, $\norm{\tilde{x}}_2 \leq 1$ and $\tilde{\lambda}_d \geq 2$.

\emph{(b) Logarithm.} For $0 \leq b \leq a \leq 1 / 2$,
$\Log{ ( 1 + a ) / ( 1 + b ) } = \Log{ 1 + \frac{a - b}{1 + b} }
\geq \frac{a - b}{( 1 + b ) ( 1 + \frac{a - b}{1 + b} )}
= \frac{a - b}{1 + a} \geq \frac23 ( a - b )$, using $\log( 1 + y )
\geq y / ( 1 + y )$ for $y \geq 0$.

\emph{(c) Term-by-term difference.} Each summand of $a - b$ is
nonnegative:
\[
a - b = \sum_{i = 1}^{d} \tilde{x}_i^2 \Big( \frac{1}{\tilde{\lambda}_i + \bar{\vartheta}}
- \frac{1}{\tilde{\lambda}_i + L} \Big)
= \sum_{i = 1}^{d} \frac{( L - \bar{\vartheta} )\, \tilde{x}_i^2}
{( \tilde{\lambda}_i + \bar{\vartheta} ) ( \tilde{\lambda}_i + L )}
\;\geq\; \sum_{i \in \cB_t} \frac{( L - \bar{\vartheta} )\, \tilde{x}_i^2}
{( \tilde{\lambda}_i + \bar{\vartheta} ) ( \tilde{\lambda}_i + L )} .
\]
\emph{(d) Band coordinates.} For $i \in \cB_t$, $\bar{\vartheta} \leq \tilde{\lambda}_d
\leq \tilde{\lambda}_i < \Gamma \bar{\vartheta}$, hence $\tilde{\lambda}_i + \bar{\vartheta} \leq 2
\tilde{\lambda}_i$, $\tilde{\lambda}_i + L < ( \Gamma + 8 \Gamma ) \bar{\vartheta} = 9
\Gamma \bar{\vartheta}$ and $L - \bar{\vartheta} \geq 7 \Gamma \bar{\vartheta}$, so
\[
\frac{( L - \bar{\vartheta} )\, \tilde{x}_i^2}{( \tilde{\lambda}_i + \bar{\vartheta} ) ( \tilde{\lambda}_i + L )}
\;\geq\; \frac{7 \Gamma \bar{\vartheta}\, \tilde{x}_i^2}{2 \tilde{\lambda}_i \cdot 9 \Gamma \bar{\vartheta}}
= \frac{7}{18}\, \frac{\tilde{x}_i^2}{\tilde{\lambda}_i} .
\]
\emph{(e) Conclusion.} Combining (a)--(d), $\Phi_{\bar{\vartheta}}( \tilde{V}_{t+1} )
- \Phi_{\bar{\vartheta}}( \tilde{V}_t ) \geq \bar{\vartheta} \cdot \frac23 \cdot \frac{7}{18}
\sum_{i \in \cB_t} \tilde{x}_i^2 / \tilde{\lambda}_i = \frac{7}{27} \bar{\vartheta} E_t^{\cB}
\geq \frac{\bar{\vartheta}}{4} E_t^{\cB}$.
\end{proof}

\subsection*{Leakage decomposition}

\begin{restatable}[Leakage decomposition]{restatablelemma}{lemBandInt}
\phantomsection\label{lem:band-int}
Let $\tau \geq T_\star$ be an epoch start with $\vartheta = \omega_\tau^{-1}$, $\bar{\vartheta}
= \vartheta / ( D_{\ast}^2 \ell_{\ast}^2 )$, band $\cB_t$ and band energy
$E_t^{\cB}$ as above. Let $\max( \tau, T_\star ) \leq t \leq T$ be
active (i.e.\ $\vartheta \leq \omega_t^{-1} \leq 2 \vartheta$). Then, on $\cE$,
\begin{equation}
\label{eq:band-decomp-int}
x_t^{\top} V_t^{-1} x_t
\;\leq\; E_t^{\cB}
\;+\; \frac{24}{t}
\;+\; \frac{c_{\mathrm{eng}}^2\, \alpha^2\, \omega_t^2}{10} .
\end{equation}
\end{restatable}

\begin{proof}[\cref{lem:band-int}]
\emph{Step 0: spectral preliminaries.} We first establish, for every
$T_\star \leq t \leq T$ on $\cE$, the three facts
\begin{equation}
\label{eq:prelim-int}
\begin{gathered}
\text{(P1)}\ \ \tilde{Q}_t + r_{\lambda_0} \leq C_2 \sqrt{t \Log{t}} ,
\qquad \text{where } \tilde{Q}_t := \sum_{s = 1}^{t - 1} \norm{\tilde{x}_s - \tilde{x}^{\star}}_2^2 ,
\\[2pt]
\text{(P2)}\ \ \tilde{\lambda}_{t,1} \geq \frac{\norm{\tilde{x}^{\star}}_2^2\, t}{8}, \ \
\tilde{\lambda}_{t,1} - \tilde{\lambda}_{t,2} \geq \frac{\norm{\tilde{x}^{\star}}_2^2\, t}{16} ,
\qquad
\text{(P3)}\ \ \norm{\tilde{x}^{\star}}_2^2\, \sin^2 \tilde{\phi}_t \cdot \tilde{\lambda}_{t,d}
\leq 16\, C_2^2\, \Log{t} .
\end{gathered}
\end{equation}
Note that the chord sum $\tilde{Q}_t$ dominates the tangential energy
$\sum_{s < t} \norm{\tilde{P}^{\perp} \tilde{x}_s}_2^2$, since $\norm{\tilde{P}^{\perp} \tilde{x}_s}_2
= \norm{\tilde{P}^{\perp} ( \tilde{x}_s - \tilde{x}^{\star} )}_2 \leq \norm{\tilde{x}_s - \tilde{x}^{\star}}_2$.

\phantomsection\label{fact:P1}
\emph{(P1): a per-round chord bound, summed against the crude bound.}
For $t_0 \leq s \leq T$, on $\cE$, \cref{prop:burnin-int}\ref{it:bint-eucl}
gives $\norm{\tilde{x}_s - \tilde{x}^{\star}}_2^2 \leq L_{\ast}^2 \beta_{+}^2\,
\omega_s / \ell_{\ast}^2$. Splitting $\tilde{Q}_t$ at $T_1$, using the trivial
bound $\norm{\tilde{x}_s - \tilde{x}^{\star}}_2^2 \leq 4$ before $T_1$ (arms lie in
$\mathbb{B}_2^d(1)$), the crude bound~\eqref{eq:crude-int} on $\omega_s$
after, and $\sum_{s = T_1}^{t-1} \sqrt{\Log{s} / s} \leq 2 \sqrt{t \Log{t}}$,
\[
\tilde{Q}_t + r_{\lambda_0}
\;\leq\; 4 T_1 + r_{\lambda_0}
+ \frac{L_{\ast}^2\, \beta_{+}^2}{\ell_{\ast}^2\, c_1}
\sum_{s = T_1}^{t - 1} \sqrt{\frac{\Log{s}}{s}}
\;\leq\; 4 T_1 + r_{\lambda_0}
+ \frac{2\, L_{\ast}^2\, \beta_{+}^2}{\ell_{\ast}^2\, c_1} \sqrt{t \Log{t}}
\;\leq\; C_2 \sqrt{t \Log{t}} ,
\]
the last step by the first defining inequality of $T_\star$, since $t
\geq T_\star$.

\phantomsection\label{fact:P2}
\emph{(P2): Rayleigh quotient of $\tilde{e}^{\star}$ and the gap.} Let
$\tilde{B}_t := \tilde{\bU}^{\top} \tilde{V}_t \tilde{\bU}$ be the compression of $\tilde{V}_t$ to
$( \tilde{e}^{\star} )^{\perp}$ ($\tilde{\bU}$ an orthonormal basis of
$( \tilde{e}^{\star} )^{\perp}$); then $\Tr \tilde{B}_t = \Tr( \tilde{\bU}^{\top} \tilde{\bR}\, \tilde{\bU} )
+ \sum_{s < t} \norm{\tilde{P}^{\perp} \tilde{x}_s}_2^2 \leq r_{\lambda_0} +
\tilde{Q}_t$, and by Cauchy interlacing $\tilde{\lambda}_{t,2} \leq
\lambda_{\max}( \tilde{B}_t ) \leq \Tr \tilde{B}_t \leq \tilde{Q}_t + r_{\lambda_0}
\leq C_2 \sqrt{t \Log{t}}$. On the other hand, by~\eqref{eq:arm-sandwich-int},
$\dotprod{\tilde{e}^{\star}}{\tilde{x}_s} \geq \norm{\tilde{x}^{\star}}_2 / 2$ for every
$s \geq t_0$, so the Rayleigh quotient of the unit vector
$\tilde{e}^{\star}$ gives
\[
\tilde{\lambda}_{t,1} \;\geq\; ( \tilde{e}^{\star} )^{\top} \tilde{V}_t \tilde{e}^{\star}
\;\geq\; \sum_{s < t} \dotprod{\tilde{e}^{\star}}{\tilde{x}_s}^2
\;\geq\; \frac{\norm{\tilde{x}^{\star}}_2^2 ( t - t_0 )}{4}
\;\geq\; \frac{\norm{\tilde{x}^{\star}}_2^2\, t}{8} ,
\]
as $\tilde{\bR} \succeq 0$ and $t \geq T_1 \geq 2 t_0$. Hence $\tilde{\lambda}_{t,1} - \tilde{\lambda}_{t,2} \geq
\norm{\tilde{x}^{\star}}_2^2 t / 8 - C_2 \sqrt{t \Log{t}} \geq \norm{\tilde{x}^{\star}}_2^2
t / 16$, by the second defining inequality of $T_\star$ and
$\norm{\tilde{x}^{\star}}_2^2 \geq \varrho_{\ast}^2 / ( D_{\ast}^2 \ell_{\ast}^2 )$
\eqref{eq:adapted-facts}.

\emph{(P3): misalignment of the top eigenvector.} For the unit vector
$\tilde{e}^{\star}$, $( \tilde{e}^{\star} )^{\top} \tilde{V}_t \tilde{e}^{\star} =
\Tr \tilde{V}_t - \Tr \tilde{B}_t$ (both sides equal $( \tilde{e}^{\star}
)^{\top} \tilde{\bR}\, \tilde{e}^{\star} + \sum_{s < t}
\dotprod{\tilde{e}^{\star}}{\tilde{x}_s}^2$, as $\norm{\tilde{x}_s}_2^2 =
\dotprod{\tilde{e}^{\star}}{\tilde{x}_s}^2 + \norm{\tilde{P}^{\perp}
\tilde{x}_s}_2^2$), so that $( \tilde{e}^{\star} )^{\top} \tilde{V}_t
\tilde{e}^{\star} \geq \tilde{\lambda}_{t,1} - \tilde{Q}_t - r_{\lambda_0}$, while
expanding
$\tilde{e}^{\star}$ in the eigenbasis, $( \tilde{e}^{\star} )^{\top} \tilde{V}_t
\tilde{e}^{\star} \leq \cos^2 \tilde{\phi}_t\, \tilde{\lambda}_{t,1} + \sin^2 \tilde{\phi}_t\,
\tilde{\lambda}_{t,2}$. Comparing,
\[
\sin^2 \tilde{\phi}_t \big( \tilde{\lambda}_{t,1} - \tilde{\lambda}_{t,2} \big)
\;\leq\; \tilde{Q}_t + r_{\lambda_0} ,
\qquad \text{so} \qquad
\sin^2 \tilde{\phi}_t \;\leq\; \frac{16\, ( \tilde{Q}_t + r_{\lambda_0} )}{\norm{\tilde{x}^{\star}}_2^2\, t}
\;\leq\; \frac{16\, C_2}{\norm{\tilde{x}^{\star}}_2^2} \sqrt{\frac{\Log{t}}{t}}
\]
by (P1)--(P2). Multiplying by $\tilde{\lambda}_{t,d} \leq \tilde{\lambda}_{t,2} \leq
C_2 \sqrt{t \Log{t}}$ (here $d \geq 2$ is used) proves (P3).

\emph{Step 1: the decomposition.} Split the energy according to
$\{ 1 \} \cup \cJ_t \cup \cB_t$ (the scalar $x_t^{\top} V_t^{-1} x_t$
is invariant, and we expand it in the adapted eigenbasis):
\[
x_t^{\top} V_t^{-1} x_t
= \tilde{x}_t^{\top} \tilde{V}_t^{-1} \tilde{x}_t
= \frac{\tilde{x}_{t,1}^2}{\tilde{\lambda}_{t,1}}
+ \sum_{i \in \cJ_t} \frac{\tilde{x}_{t,i}^2}{\tilde{\lambda}_{t,i}}
+ E_t^{\cB} .
\]

\phantomsection\label{step:band-a}
\emph{(a) Top direction.} By~\eqref{eq:arm-sandwich-int} and (P2),
\[
\frac{\tilde{x}_{t,1}^2}{\tilde{\lambda}_{t,1}}
\;\leq\; \frac{\norm{\tilde{x}_t}_2^2}{\tilde{\lambda}_{t,1}}
\;\leq\; \frac{9 \norm{\tilde{x}^{\star}}_2^2 / 4}{\norm{\tilde{x}^{\star}}_2^2\, t / 8}
= \frac{18}{t} \;\leq\; \frac{24}{t} .
\]

\phantomsection\label{step:band-b}
\emph{(b) Mid-scale coordinates.} Write $\tilde{x}_{t,i} = \dotprod{\tilde{x}_t -
\tilde{x}^{\star}}{\tilde{v}_{t,i}} + \dotprod{\tilde{x}^{\star}}{\tilde{v}_{t,i}}$ for $i \in \cJ_t$
and use $( u + v )^2 \leq 2 u^2 + 2 v^2$:
\[
q_t^{\cJ} := \sum_{i \in \cJ_t} \tilde{x}_{t,i}^2
\;\leq\; 2 \norm{\tilde{x}_t - \tilde{x}^{\star}}_2^2
+ 2 \sum_{i \geq 2} \dotprod{\tilde{x}^{\star}}{\tilde{v}_{t,i}}^2
= 2 \norm{\tilde{x}_t - \tilde{x}^{\star}}_2^2 + 2 \norm{\tilde{x}^{\star}}_2^2 \sin^2 \tilde{\phi}_t ,
\]
since $\sum_{i \geq 2} \dotprod{\tilde{x}^{\star}}{\tilde{v}_{t,i}}^2 = \norm{\tilde{x}^{\star}}_2^2
- \dotprod{\tilde{x}^{\star}}{\tilde{v}_{t,1}}^2 = \norm{\tilde{x}^{\star}}_2^2 \sin^2 \tilde{\phi}_t$.
Now $\min_{i \in \cJ_t} \tilde{\lambda}_{t,i} \geq \Gamma \bar{\vartheta} =
\Gamma \vartheta / ( D_{\ast}^2 \ell_{\ast}^2 ) \geq \Gamma /
( 2 D_{\ast}^2 \ell_{\ast}^2\, \omega_t )$ by activity ($\vartheta \geq \omega_t^{-1} / 2$);
dividing,
\[
\sum_{i \in \cJ_t} \frac{\tilde{x}_{t,i}^2}{\tilde{\lambda}_{t,i}}
\;\leq\; \frac{2 D_{\ast}^2 \ell_{\ast}^2\, \omega_t\, q_t^{\cJ}}{\Gamma}
\;\leq\; \frac{4 D_{\ast}^2 \ell_{\ast}^2\, \omega_t \norm{\tilde{x}_t - \tilde{x}^{\star}}_2^2}{\Gamma}
+ \frac{4 D_{\ast}^2 \ell_{\ast}^2\, \omega_t \norm{\tilde{x}^{\star}}_2^2 \sin^2 \tilde{\phi}_t}{\Gamma} .
\]
\emph{Estimation part.} By \cref{prop:burnin-int}\ref{it:bint-eucl} and
the first fact in~\eqref{eq:Gamma-facts-int},
\[
\frac{4 D_{\ast}^2 \ell_{\ast}^2\, \omega_t \norm{\tilde{x}_t - \tilde{x}^{\star}}_2^2}{\Gamma}
\;\leq\; \frac{4 D_{\ast}^2\, L_{\ast}^2\, \beta_{+}^2\, \omega_t^2}{\Gamma}
\;\leq\; \frac{c_{\mathrm{eng}}^2\, \alpha^2\, \omega_t^2}{20} .
\]
\emph{Misalignment part.} By (P3), $\Log{t} \leq \Log{T}$,
$\tilde{\lambda}_{t,d} \geq \omega_t^{-1} / ( D_{\ast}^2 \ell_{\ast}^2 )$ and the second
fact in~\eqref{eq:Gamma-facts-int},
\[
\frac{4 D_{\ast}^2 \ell_{\ast}^2\, \omega_t \norm{\tilde{x}^{\star}}_2^2 \sin^2 \tilde{\phi}_t}{\Gamma}
\;\leq\; \frac{64\, D_{\ast}^2 \ell_{\ast}^2\, C_2^2\, \Log{T}\, \omega_t}{\Gamma\, \tilde{\lambda}_{t,d}}
\;\leq\; \frac{64\, D_{\ast}^4 \ell_{\ast}^4\, C_2^2\, \Log{T}\, \omega_t^2}{\Gamma}
\;\leq\; \frac{c_{\mathrm{eng}}^2\, \alpha^2\, \omega_t^2}{20} .
\]
\emph{Conclusion.} Adding (a) and the two parts of (b) to $E_t^{\cB}$
gives~\eqref{eq:band-decomp-int}, with $\frac{1}{20} + \frac{1}{20} =
\frac{1}{10}$.
\end{proof}

\begin{corollary}[Band energy capture]
\phantomsection\label{cor:band-int}
Under the hypotheses of \cref{lem:band-int}, combining
\eqref{eq:band-decomp-int} with the engine~\eqref{eq:engine-int},
\[
E_t^{\cB}
\;\geq\; \Big( 1 - \frac{1}{10} \Big)\, c_{\mathrm{eng}}^2\, \alpha^2\, \omega_t^2
- \frac{24}{t}
\;\geq\; \frac{9\, c_{\mathrm{eng}}^2\, \alpha^2}{40\, \vartheta^2} - \frac{24}{t} ,
\]
the last step since $\omega_t^{-1} \leq 2 \vartheta$ on active rounds.
\end{corollary}

\subsection*{Doubling}

\begin{restatable}[Doubling]{restatablelemma}{lemDoublingInt}
\phantomsection\label{lem:doubling-int}
Let $\tau \geq T_\star$ with $\vartheta := \omega_\tau^{-1}$, let $K = 40\, d \Log{8
\Gamma} / ( c_{\mathrm{eng}}^2 \alpha^2 )$ as in~\eqref{eq:constants-int},
and suppose $\tau + \lceil K \vartheta^2 \rceil \leq T$. Then, on $\cE$,
\[
\omega_{\tau + \lceil K \vartheta^2 \rceil}^{-1}
\;\geq\; \min\Big( 2 \vartheta,\ \frac{c_{\mathrm{eng}}\, \alpha \sqrt{\tau}}{32} \Big) .
\]
\end{restatable}

\begin{proof}[\cref{lem:doubling-int}]
Set $N := \lceil K \vartheta^2 \rceil$ and suppose, for contradiction, that
$\omega_{\tau + N}^{-1} < \min( 2 \vartheta, c_{\mathrm{eng}} \alpha \sqrt{\tau} / 32 )$.
Then all rounds $t \in [ \tau, \tau + N )$ are active (by monotonicity,
$\omega_t^{-1} \leq \omega_{\tau + N}^{-1} < 2 \vartheta$), and $\tau > 1024 \vartheta^2 / ( c_{\mathrm{eng}}^2
\alpha^2 )$, so that $24 / t \leq 24 / \tau < 24\, c_{\mathrm{eng}}^2 \alpha^2
/ ( 1024 \vartheta^2 ) \leq c_{\mathrm{eng}}^2 \alpha^2 / ( 40 \vartheta^2 )$ for every $t
\geq \tau$. By \cref{cor:band-int}, for every $t \in [ \tau, \tau + N )$,
\[
E_t^{\cB} \;\geq\; \frac{9\, c_{\mathrm{eng}}^2 \alpha^2}{40 \vartheta^2}
- \frac{c_{\mathrm{eng}}^2 \alpha^2}{40 \vartheta^2}
= \frac{c_{\mathrm{eng}}^2\, \alpha^2}{5\, \vartheta^2} .
\]
By \cref{lem:potential-int}\ref{it:potint-gain}, each such round
deposits at least $\frac{\bar{\vartheta}}{4} \cdot \frac{c_{\mathrm{eng}}^2 \alpha^2}{5
\vartheta^2} = \frac{\bar{\vartheta}\, c_{\mathrm{eng}}^2 \alpha^2}{20 \vartheta^2}$ into the
potential, and the potential never decreases; so
\[
\Phi_{\bar{\vartheta}}( \tilde{V}_{\tau + N} ) - \Phi_{\bar{\vartheta}}( \tilde{V}_\tau )
\;\geq\; N\, \frac{\bar{\vartheta}\, c_{\mathrm{eng}}^2 \alpha^2}{20 \vartheta^2}
\;\geq\; K \vartheta^2\, \frac{\bar{\vartheta}\, c_{\mathrm{eng}}^2 \alpha^2}{20 \vartheta^2}
= 2\, d\, \bar{\vartheta} \Log{8 \Gamma} .
\]
But by \cref{lem:potential-int}\ref{it:potint-capacity} the left side is
at most $d\, \bar{\vartheta} \Log{8 \Gamma}$: contradiction. The Euclidean
scale $\bar{\vartheta}$ appears on both sides of the budget count and cancels,
which is why $K$ depends neither on $D_{\ast}$ nor on $\ell_{\ast}$; its
factor $d$ is the capacity $d\, \bar{\vartheta} \Log{8 \Gamma}$ of the
potential, the engine contributing $\alpha^2$ only.
\end{proof}

\subsection*{Proof of the main theorem}

\begin{lemma}[Initialisation window]
\phantomsection\label{lem:init-int}
On $\cE$, $\omega_t^{-1} \geq c_0 \sqrt{t}$ for all $t \in [ T_B, \min( 9 T_B, T ) ]$.
\end{lemma}

\begin{proof}[\cref{lem:init-int}]
For $t \geq T_B = T_\star \geq T_1$, the crude bound gives $\omega_t^{-1} \geq
c_1 \sqrt{t / \Log{t}}$; and for $t \leq 9 T_\star$, $\sqrt{t / \Log{t}}
\geq \sqrt{t} / \sqrt{\Log{9 T_\star}}$, so $\omega_t^{-1} \geq c_1 \sqrt{t} /
\sqrt{\Log{9 T_\star}} \geq c_0 \sqrt{t}$ by the definition of $c_0$.
\end{proof}

\begin{restatable}[Induction]{restatableproposition}{propInductionInt}
\phantomsection\label{prop:induction-int}
On $\cE$, $\omega_t^{-1} \geq c_0 \sqrt{t}$ for all $T_B \leq t \leq T$.
\end{restatable}

\begin{proof}[\cref{prop:induction-int}]
Suppose not, and let $t^{\dagger} := \min\{ t \in [ T_B, T ] : \omega_t^{-1} <
c_0 \sqrt{t} \}$ be the first failure. By \cref{lem:init-int},
$t^{\dagger} > 9 T_B$. Put $\vartheta := \omega_{t^{\dagger}}^{-1} < c_0 \sqrt{t^{\dagger}}$
and $N := \lceil K \vartheta^2 \rceil$; then $K \vartheta^2 < K c_0^2 t^{\dagger} \leq
t^{\dagger} / 16$ since $c_0 \leq 1 / ( 4 \sqrt{K} )$, so $N \leq t^{\dagger}
/ 16 + 1 \leq t^{\dagger} / 8$ (as $t^{\dagger} \geq 20 \geq 16$). Let
$\tau := t^{\dagger} - N \geq \frac78 t^{\dagger} > 7 T_B \geq T_\star$;
by minimality of $t^{\dagger}$, $\omega_\tau^{-1} \geq c_0 \sqrt{\tau}$,
while by monotonicity $\omega_\tau^{-1} \leq \omega_{t^{\dagger}}^{-1} =
\vartheta$. Apply \cref{lem:doubling-int} at $\tau$ with reference scale $\vartheta' :=
\omega_\tau^{-1} \leq \vartheta$: since $\lceil K \vartheta'^2 \rceil \leq N$ and $\tau + \lceil K \vartheta'^2
\rceil \leq t^{\dagger} \leq T$, monotonicity gives
\[
\vartheta = \omega_{t^{\dagger}}^{-1}
\;\geq\; \omega_{\tau + \lceil K \vartheta'^2 \rceil}^{-1}
\;\geq\; \min\Big( 2 \vartheta',\ \frac{c_{\mathrm{eng}}\, \alpha \sqrt{\tau}}{32} \Big) .
\]
Two cases. If the minimum is $2 \vartheta'$, then $\vartheta \geq 2 \vartheta' \geq 2 c_0
\sqrt{\tau} \geq 2 c_0 \sqrt{7 t^{\dagger} / 8} > c_0 \sqrt{t^{\dagger}}$,
contradicting $\vartheta < c_0 \sqrt{t^{\dagger}}$. If the minimum is
$c_{\mathrm{eng}} \alpha \sqrt{\tau} / 32$, then $\vartheta \geq \frac{c_{\mathrm{eng}}
\alpha}{32} \sqrt{7 t^{\dagger} / 8} \geq c_0 \sqrt{t^{\dagger}}$,
because $c_0 \leq \frac{1}{4 \sqrt{K}} = \frac{c_{\mathrm{eng}} \alpha}{4
\sqrt{40\, d \Log{8 \Gamma}}} \leq \frac{c_{\mathrm{eng}} \alpha}{4
\sqrt{160}} < \frac{c_{\mathrm{eng}} \alpha}{32} \sqrt{\frac78}$ (using
$\Log{8 \Gamma} \geq \Log{8} \geq 2$, $d \geq 2$ and $4 \sqrt{160} >
32 \sqrt{8 / 7}$),
again contradicting the definition of $\vartheta$.
\end{proof}

With this in hand, the proof of~\cref{thm:spectral_growth} is straightforward.

\thmMinEigInt*

\begin{proof}[\cref{thm:spectral_growth}]
By \cref{prop:induction-int}, $\omega_t^{-1} \geq c_0 \sqrt{t}$ on $\cE$ for
$T_B \leq t \leq T$; since $\omega_t$ is invariant under the change of
coordinates of \cref{lem:adapted-int}, this is a statement about the
original problem, and it is the first display by
\cref{prop:width-int}\ref{it:wint-radius}. The three consequences are
\cref{prop:burnin-int}\ref{it:bint-hat}--\ref{it:bint-arm} with $\omega_t^{-1}
\geq c_0 \sqrt{t}$. The event $\cE$ has probability at least $1 -
\delta$ by \cref{prop:Yassin_conf}. For the constant, $\frac{1}{4
\sqrt{K}} = \frac{c_{\mathrm{eng}} \alpha}{\sqrt{640\, d \Log{8 \Gamma}}}$
and $\frac{c_1}{\sqrt{\Log{9 T_\star}}} = \frac{c_{\mathrm{eng}}
\alpha}{\sqrt{4\, d \Log{9 T_\star}}}$, so
\[
c_0 \;\geq\; \frac{c_{\mathrm{eng}}\, \alpha}{\sqrt{640\, d\, ( \Log{8 \Gamma} +
\Log{9 T_\star} )}}
\;\geq\; \frac{m\, \alpha}{152\, J_{\star} \sqrt{d}\, \sqrt{
\Log{8 \Gamma} + \Log{9 T_\star} }} ;
\]

where, by \cref{def:constants-int,def:ladder-int}, the factor
$\Log{8 \Gamma} + \Log{9 T_\star} = \cO\bigl( \log \log T \bigr)$
up to terms logarithmic in the problem parameters, i.e.\ it contributes
only an iterated-logarithm factor in $T$.
\end{proof}

\section{Regret of \texorpdfstring{$\alpha$}{alpha}-LinUCB}
\label{sec:regret-proof}

\subsection{Regret decomposition}
\phantomsection\label{sec:regret_decomposition}

\begin{restatable}[Per-round regret]{restatablelemma}{lemRegretDecomp}
\phantomsection\label{lem:regret-decomposition}
On the event $\cE$, for every $T_B \leq t \leq T$, the instantaneous
regret $r_t := J_{\star} - \dotprod{\theta^{\star}}{x_t}$ satisfies
\begin{align*}
    r_t \;\leq\; \frac{D_{J_{\cX}^2}( \theta^{\star}, \theta_t )}{J_{\star}} .
\end{align*}
\end{restatable}

\begin{proof}[\cref{lem:regret-decomposition}]
Fix $T_B \leq t \leq T$. By \cref{prop:global-int}\ref{it:gl-argmax},
$\nabla J_{\cX}^2( \theta_t ) = 2 J_{\cX}( \theta_t )\, x_t$ with
$\dotprod{\theta_t}{x_t} = J_{\cX}( \theta_t )$; by
\cref{prop:burnin-int}\ref{it:bint-arm}, $J_{\cX}( \theta_t ) \geq
\tfrac12 J_{\star}$, as $t \geq T_B \geq t_0$; and $\dotprod{\theta^{\star}}{x_t}
\leq J_{\star}$ since $x_t \in \cX$. Hence
\begin{align*}
    \frac{D_{J_{\cX}^2}(\theta^{\star}, \theta_t)}{J_{\star}} - r_t
    &= J_{\star} - \frac{J_{\cX}^2(\theta_t)}{J_{\star}}
        - \frac{2 J_{\cX}(\theta_t)}{J_{\star}}
        \dotprod{x_t}{\theta^{\star} - \theta_t}
        - J_{\star} + \dotprod{\theta^{\star}}{x_t} \\
    &= \frac{J_{\cX}^2(\theta_t)}{J_{\star}}
        - \Big( \frac{2 J_{\cX}(\theta_t)}{J_{\star}} - 1 \Big)
        \dotprod{\theta^{\star}}{x_t}
        && (\dotprod{\theta_t}{x_t} = J_{\cX}(\theta_t)) \\
    &\geq \frac{J_{\cX}^2(\theta_t)}{J_{\star}}
        - \Big( \frac{2 J_{\cX}(\theta_t)}{J_{\star}} - 1 \Big) J_{\star}
        && (J_{\cX}(\theta_t) \geq \tfrac12 J_{\star},\
            \dotprod{\theta^{\star}}{x_t} \leq J_{\star}) \\
    &= \frac{\big( J_{\cX}(\theta_t) - J_{\star} \big)^2}{J_{\star}}
        \;\geq\; 0 .
\end{align*}
\end{proof}

\subsection{Proof of~\cref{thm:alpha-UCB-Regret}}

\thmAlphaUCBRegret*

\begin{proof}[\cref{thm:alpha-UCB-Regret}]
We have $\cR_T = \sum_{t = 1}^{T} r_t$ with $r_t = J_{\star} -
\dotprod{\theta^{\star}}{x_t} \geq 0$. For the rounds $t < T_B$ we use the
crude bound
\[
r_t \;=\; \dotprod{\theta^{\star}}{x^{\star} - x_t}
\;\leq\; \norm{\theta^{\star}}_{\ast} \norm{x^{\star} - x_t}_{\ast}^{\circ}
\;\leq\; 2 \ell_{\ast} \norm{\theta^{\star}}_{\ast}
\;\leq\; \frac{2 \ell_{\ast} J_{\star}}{\varrho_{\ast}} ,
\]
since $\norm{x}_{\ast}^{\circ} \leq \ell_{\ast}$ on $\cX$ and
$\varrho_{\ast} \norm{\theta^{\star}}_{\ast} \leq J_{\cX}( \theta^{\star} )
= J_{\star}$ by~\eqref{eq:scalars-upperbound}. On $\cE$, for every $T_B
\leq t \leq T$,
\begin{align*}
r_t
&\leq \frac{D_{J_{\cX}^2}( \theta^{\star}, \theta_t )}{J_{\star}}
\;\leq\; \frac{M}{2 J_{\star}} \norm{\theta^{\star} - \theta_t}_{\ast}^2
& ( \text{\cref{lem:regret-decomposition} and~\eqref{eq:ajp-int}} ) \\
&\leq \frac{M}{J_{\star}} \Big( \norm{\theta^{\star} - \hat{\theta}_t}_{\ast}^2
+ \norm{\hat{\theta}_t - \theta_t}_{\ast}^2 \Big)
& ( ( a + b )^2 \leq 2 a^2 + 2 b^2 ) \\
&\leq \frac{M}{J_{\star}} \big( \alpha^2 + \beta^2 \big)\, \omega_t
& ( \text{\cref{prop:width-int}\ref{it:wint-radius}} ) ,
\end{align*}
where the last step uses $\theta^{\star} \in \cC_t( \beta )$ on $\cE$ and
$\theta_t \in \cC_t( \alpha )$ (see~\cref{prop:width-int}\ref{it:wint-radius}). Summing,
\begin{align*}
\cR_T
&\leq \frac{2 \ell_{\ast} J_{\star}}{\varrho_{\ast}}\, T_B
+ \frac{M ( \alpha^2 + \beta^2 )}{J_{\star}} \sum_{t = T_B}^{T} \omega_t \\
&\leq \frac{2 \ell_{\ast} J_{\star}}{\varrho_{\ast}}\, T_B
+ \frac{M ( \alpha^2 + \beta^2 )}{J_{\star}\, c_0} \sum_{t = T_B}^{T} \frac{1}{\sqrt{t}}
& ( \text{\cref{thm:spectral_growth}, } t \geq T_B ) \\
&\leq \frac{2 \ell_{\ast} J_{\star}}{\varrho_{\ast}}\, T_B
+ \frac{2 M ( \alpha^2 + \beta^2 )}{J_{\star}\, c_0} \sqrt{T}
& \Big( \textstyle\sum_{t \leq T} t^{-1/2} \leq 2 \sqrt{T} \Big) ,\\
&={\cO}\Big( \frac{M}{m}\, \sqrt{d}\, \frac{\alpha^2 + \beta^2}{\alpha}\
\sqrt{T} \Big).
\end{align*}
The last step uses $c_0 \gtrsim m \alpha / ( J_{\star} \sqrt{d}\,
 )$
(\cref{thm:spectral_growth}), so that the second term is at most
$\cO(\, \frac{M}{m}\, \sqrt{d}\, \frac{\alpha^2 + \beta^2}{\alpha}
\, \sqrt{T})$, and $T_B = \cO( \operatorname{poly}( \log T ) )$ (\cref{def:ladder-int}).
\end{proof}

\end{document}